\documentclass[10pt,twocolumn,letterpaper]{article}

\usepackage[pagenumbers]{cvpr} % To force page numbers, e.g. for an arXiv version

\newcommand{\whitetext}[1]{\textcolor{white}{#1}}

\usepackage{pifont}% http://ctan.org/pkg/pifont
\newcommand{\xmark}{\ding{55}}%

\usepackage{mathtools}
\usepackage{enumitem} % for [itemsep=0pt] next to {itemize} env
\usepackage{stmaryrd} % Where \llbracket and \rrbracket are defined
\usepackage[ruled,vlined]{algorithm2e}
\usepackage{appendix}     %to help with several appendices
\usepackage{stackengine} % to use \stackinset to add labels on figures

\usepackage{microtype}

\renewcommand{\paragraph}[1]{\vspace{.5em}\noindent\textbf{#1.}}

\usepackage{amsmath,amsfonts,bm}

\def\diag{{\textnormal{diag}}}

\newtheorem{remark}{Remark}[section] 

\def\eqref#1{equation~\ref{#1}}
\def\1{\bm{1}}

\def\rvx{{\mathbf{x}}}
\def\rvy{{\mathbf{y}}}

\def\rmS{{\mathbf{S}}}

\def\vm{{\bm{m}}}

\def\vx{{\bm{x}}}
\def\vy{{\bm{y}}}

\def\mA{{\bm{A}}}

\def\mF{{\bm{F}}}

\def\mH{{\bm{H}}}
\def\mI{{\bm{I}}}

\def\mM{{\bm{M}}}

\def\mS{{\bm{S}}}

\DeclareMathAlphabet{\mathsfit}{\encodingdefault}{\sfdefault}{m}{sl}
\SetMathAlphabet{\mathsfit}{bold}{\encodingdefault}{\sfdefault}{bx}{n}

\def\sC{{\mathbb{C}}}

\newcommand{\E}{\mathbb{E}}

\newcommand{\R}{\mathbb{R}}

\DeclareMathOperator*{\argmin}{arg\,min}

\definecolor{cvprblue}{rgb}{0.21,0.49,0.74}
\usepackage[pagebackref,breaklinks,colorlinks,allcolors=cvprblue]{hyperref}

\def\paperID{***} % *** Enter the Paper ID here
\def\confName{CVPR}
\def\confYear{2026}

\title{Efficient Unrolled Networks for Large-Scale 3D Inverse Problems}

\author{Romain Vo \quad Julián Tachella\\
CNRS, ENS de Lyon, Laboratoire de Physique, Lyon, France\\
{\tt\small romain.vo@ens-lyon.fr \quad julian.tachella@ens-lyon.fr}
}

\begin{document}
\maketitle
\begin{abstract}
Deep learning-based methods have revolutionized the field of imaging inverse problems, yielding state-of-the-art performance across various imaging domains. The best performing networks incorporate the imaging operator within the network architecture, typically in the form of deep unrolling. However, in large-scale problems, such as 3D imaging, most existing methods fail to incorporate the operator in the architecture due to the prohibitive amount of memory required by global forward operators, which hinder typical patching strategies. In this work, we present a domain partitioning strategy and normal operator approximations that enable the training of end-to-end reconstruction models incorporating forward operators of arbitrarily large problems into their architecture. The proposed method achieves state-of-the-art performance on 3D X-ray cone-beam tomography and 3D multi-coil accelerated MRI, while requiring only a single GPU for both training and inference.
\end{abstract}    
\section{Introduction} \label{sec:intro}

\noindent Linear inverse problems are ubiquitous in science and engineering, with applications ranging from medical imaging to astronomy and remote sensing. These problems typically involve recovering an unknown signal $\vx^* \in \R^n$ from noisy linear measurements $\vy \in \R^m$ obtained via a known linear operator $\mA \in \R^{m \times n}$:
\begin{equation} \label{eq:inverse:problem}
 \vy = \mA \vx^* + \bm{\varepsilon},
\end{equation}
where $\bm{\varepsilon}$ represents measurement noise. Such problems are often ill-posed due to the lack of observed data, necessitating the use of regularization techniques to ensure stable and meaningful solutions. 

In recent years, deep learning has emerged as a powerful tool for solving inverse problems, leveraging large datasets \cite{zbontar_fastmri_2019,leuschnerLoDoPaBCTDatasetBenchmark2021} to learn complex mappings from measurements to signals. Notable approaches include \textit{post-processing} methods and \textit{unrolled networks}, both of which are trained end-to-end and attempt to recover the Minimum Mean Squared Error estimator (MMSE) \cite{nguyenComparingPlugandPlayUnrolled2024}. The former learns a direct mapping between a low-quality reconstruction (typically the adjoint or linear pseudo-inverse reconstruction) and its ground-truth reference. While straightforward and effective, it often fails to enforce data-consistency, leading to reconstructions that may not align well with the observed measurements \cite{leuschnerLoDoPaBCTDatasetBenchmark2021,mukherjeeLearnedReconstructionMethods2023,hauptmann_convergent_2024}. Unrolled networks, on the other hand, integrate knowledge of the forward operator $\mA$ into the reconstruction process by unrolling a fixed number of iterations of an optimization algorithm and replacing specific steps with learnable components \cite{adlerSolvingIllposedInverse2017,adlerLearnedPrimaldualReconstruction2018,sriram_end--end_2020,pezzotti_adaptive_2020}. They demonstrate state-of-the-art performance across a wide range of benchmarks, combining the interpretability of traditional methods with the flexibility of deep learning.

\begin{figure}
    \centering
    \includegraphics[width=0.9\linewidth]{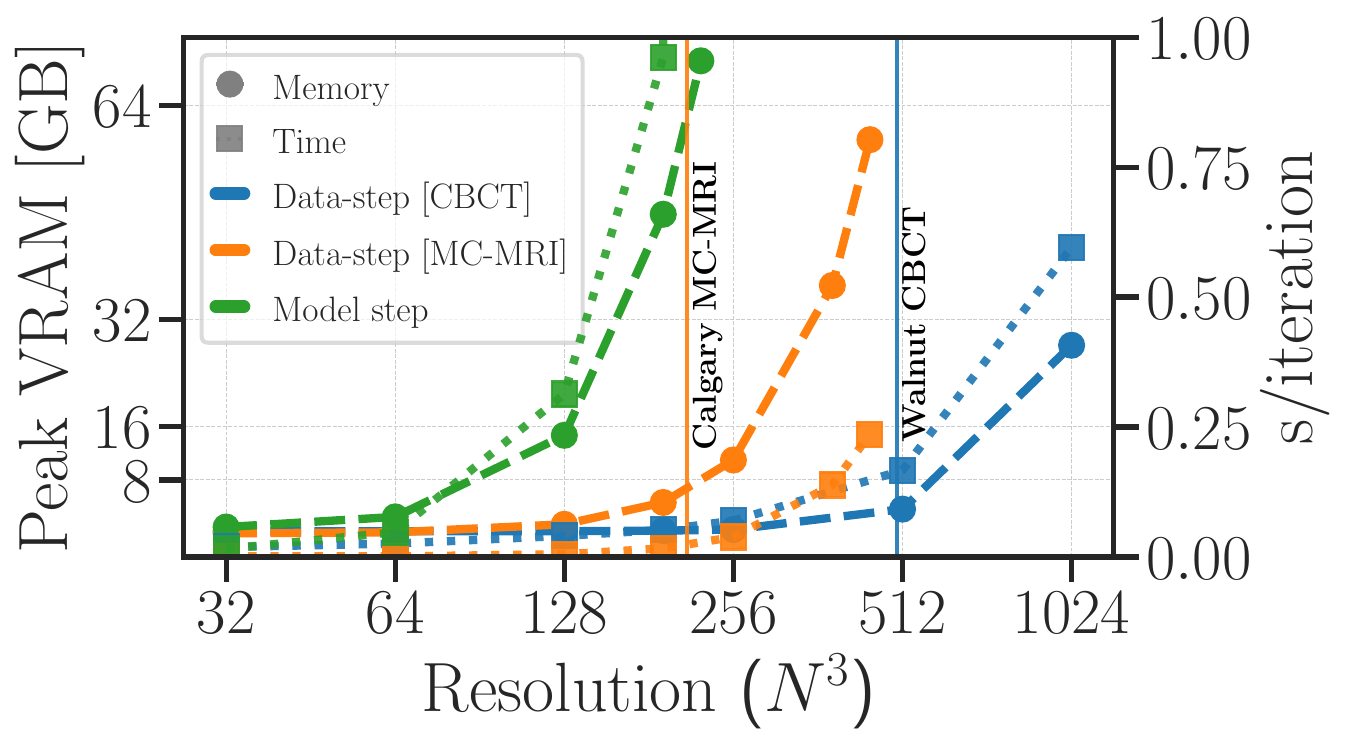}
    \caption{Peak video memory complexity (\textit{dashed lines}) and global execution times (\textit{dotted lines}) of isolated components used in unrolling. We show the cost of evaluating and back-propagating through a standard 3D data consistency step (using gradient descent) and a standard 3D network step (using a 3D DRUNet \cite{zhangPlugPlayImageRestoration2021}). We see here that the bottleneck lies in the network step, which grows rapidly with the volume size, while the data-consistency step remains manageable even at high resolutions.}
    \label{fig:intro}
\end{figure}

However, incorporating global operators $\mA$ in the architecture requires training the network step at the full resolution. While in 2D problems this is typically feasible on a single GPU, the memory requirements of a global forward pass become prohibitive when working on 3D or higher dimensional problems, as illustrated in \cref{fig:intro}.
Deep equilibrium training reduces the memory complexity of unrolled training to that of a single pass \cite{baiDeepEquilibriumModels2019,giltonDeepEquilibriumArchitectures2021,fungJFBJacobianFreeBackpropagation2021,liuOnlineDeepEquilibrium2022}. Nonetheless, it still requires evaluating the full network at each iteration, which is infeasible on a single GPU for large-scale 3D problems. 

In this work, we address this scaling challenge by introducing two complementary techniques: domain partitioning and normal operator approximation. Domain partitioning enables us to decompose a large-scale inverse problem into smaller, more manageable subproblems, allowing us to adapt the network complexity to the available resources. This approach is inspired by patch-based training methods \cite{ronnebergerUNetConvolutionalNetworks2015,zhangLearningDeepCNN2017}, but extends them to the context of unrolled networks for the more challenging case where the forward operator is not trivially decomposable into patches (\eg X-ray cone-beam CT). Normal operator approximation involves replacing $\mA^\top \mA$ by a product of diagonal and circulant matrices \cite{schmid_decomposing_2000,Vetterli_Kovačević_Goyal_2014}, enabling efficient computation of data-consistency updates via the Fast Fourier Transform (FFT). By combining these techniques, we develop a framework that allows the training and deployment of unrolled networks for arbitrarily large linear inverse problems using only one GPU. This framework can be adapted to various types of linear operators. We demonstrate the effectiveness of our approach through extensive experiments on large-scale 3D inverse problems, showcasing state-of-the-art performance while significantly reducing memory and computational requirements. Our contributions are as follows:
\begin{itemize}
    \item We propose a domain partitioning strategy of the operator $\mA$ that enables the training of unrolled networks on small patches, facilitating scaling to large problems.
    \item We introduce a normal operator approximation technique that leverages diagonal and circulant matrix products for efficient data-consistency updates. Notably, we show that the parameters of the factorization can be  recovered efficiently by gradient descent without problem-specific data.
    \item We validate our approach on large-scale 3D inverse problems, namely Multi-Coil Magnetic Resonance Imaging (MC-MRI) and Cone-Beam X-ray Computed Tomography (CBCT). We demonstrate competitive performance with reduced resource consumption, \eg handling up to $501^3$ volumes on a single GPU.
\end{itemize}
\section{Related work} \label{sec:related:work}

\paragraph{Post-processing} Also called \emph{artifact removal} or \emph{restoration} network, a straightforward approach to building a reconstruction function is to learn a one-pass mapping between a low-quality reconstruction and its ground-truth reference \cite{hanDeepResidualLearning2016,lee_deep_2018}. State-of-the-art post-processing techniques rely on widespread architectures to perform the reconstruction, \eg UNet-based networks \cite{ronnebergerUNetConvolutionalNetworks2015} and residual convolutional networks \cite{zhangPlugPlayImageRestoration2021}. The most recent developments focus on transformer-based architectures \cite{dosovitskiyImageWorth16x162021,Zamir_2022_CVPR}, which provide better performance in the high-data regime. While simple in design, post-processing scales well to large problems by leveraging patch-based training \cite{zhangLearningDeepCNN2017}. However, it does not leverage the knowledge of the forward operator, which can lead to reconstructions that are not data-consistent and is not well-suited for problems where the acquisition conditions, represented by $\mA$, may vary \cite{leuschnerQuantitativeComparisonDeep2021,terris_reconstruct_2025}.

\paragraph{Learned priors} Instead of handcrafting a regularization term, data-driven approaches \cite{lunzAdversarialRegularizersInverse2018,mukherjeeLearnedReconstructionMethods2023} use deep neural networks to learn the prior from data. In particular, Plug-and-Play (PnP) approaches \cite{venkatakrishnanPlugPlayPriorsModel2013,sunOnlinePlugandPlayAlgorithm2019} leverage off-the-shelf denoisers to replace the \textit{prior} step in iterative optimization algorithms. More recently, denoising diffusion models \cite{hoDenoisingDiffusionProbabilistic2020} have also been used as priors in inverse problems, leveraging their strong generative capabilities \cite{songSolvingInverseProblems2022,chungSolving3DInverse2023,zhuDenoisingDiffusionModels2023}. Due to their reliance on pre-trained networks, these methods can scale relatively well to large-scale problems, \ie by trading off speed for memory \cite{voPlugPlayLearnedProximal2025}. Nonetheless, they are prone to instabilities \cite{pesquetLearningMaximallyMonotone2021,cohenRegularizationDenoisingFixedPoint2021}, and typically obtain worse MSE than unrolled networks trained in an end-to-end fashion. Moreover, in the case of 3D problems, pre-trained denoisers are often limited to 2D networks, which do not leverage the 3D structure of the data. 

\paragraph{Unrolling} 
The core idea of unrolling is to transform a fixed number of optimization iterations into a network architecture, with the prior step parameterized as a learnable component \cite{adlerSolvingIllposedInverse2017,aggarwal_modl_2019}.
As opposed to PnP, an unrolled network is trained end-to-end for the specific inverse problem, which generally leads to better performance \cite{leuschnerLoDoPaBCTDatasetBenchmark2021,WANG2021102579}. However, unrolled methods are limited in practice to small-scale problems due to the high memory consumption required for storing all intermediate activations during training. This is especially true for 3D problems, where the memory consumption of the network step explodes and largely dominates that of the data-consistency step (\cref{fig:intro}). To alleviate this issue, some works have proposed to use reversible networks \cite{sanderMomentumResidualNeural2021,rudzusikaInvertibleLearnedPrimalDual2021}, sketching \cite{tangAcceleratingDeepUnrolling2022}, or checkpointing \cite{kellman_memory-efficient_2020,hauptmannModelBasedLearningAccelerated2020}. Nonetheless, these approaches have yet to be sufficient to train unrolled networks on real-world 3D problems ($\approx 512^3$ voxels reconstructions) with global operators, \ie operators acting on the entire volume.

\paragraph{Implicit neural representations (INR)} Popularized by recent advances in computer graphics \cite{mildenhallNeRFRepresentingScenes2020,mullerInstantNeuralGraphics2022} and also known as \emph{neural fields}, INRs consist in representing a signal as a continuous function parameterized by a neural network \cite{tancikFourierFeaturesLet2020,xieNeuralFieldsVisual2022}. They generally require fewer parameters than traditional discrete representations, making them appealing for large-scale 3D problems \cite{sunCoILCoordinateBasedInternal2021,shenNeRPImplicitNeural2022,zhaNAFNeuralAttenuation2022,voPlugPlayLearnedProximal2025}. INRs are usually optimized per-sample using gradient-descent based algorithms, \ie a separate network representation is trained for each test sample \cite{xieNeuralFieldsVisual2022}, which limits their applicability in large-scale inverse problems. 
They also tend to yield subpar results compared to standard feed-forward networks, as they do not leverage prior knowledge from data. Indeed, they fail to thoroughly address the scalability issues in large-scale problems, as efficiently learning a prior over INRs requires training a network in weight space, which remains a significant challenge \cite{dupontDataFunctaYour2022}

\paragraph{Fast linear operator approximations} Matrix factorization techniques approximate a linear operator by a product of structured matrices \cite{zheng_efficient_2023}, \eg sparse, low-rank, or circulant matrices. This topic has been widely studied to reduce the computational burden of large-scale linear operators across various fields, including inverse problems \cite{elad_sparse_2010,bolte_proximal_2014}. An example of interest is the factorization of the Discrete Fourier Transform matrix by the Fast Fourier Transform (FFT) algorithm, which reduces the evaluation to a product of sparse matrices \cite{le_spurious_2023}. Closely related to the method we propose, Schmid \etal \cite{schmid_decomposing_2000} show that any square linear operator can be factored into a finite product of diagonal and circulant matrices. We build upon these results to efficiently approximate the normal operator $\mA^\top \mA$ of large-scale inverse problems as the product of two factors, enabling fast data-consistency updates during unrolled training.

\begin{remark}
    This work is still relevant in the context of matrix-free operators, which is the case for the inverse problems implemented in our experiments, \ie we compute $\mA \vx$ and $\mA^\top \vy$ for any $\vx$ and $\vy$ without explicitly forming the matrices. 
\end{remark}
\section{Background} \label{sec:background}

\paragraph{Variational approach} 
When an inverse problem is ill-posed, \ie when $m \ll n$ or $\mA$ is ill-conditioned, a common approach to recover $\vx^{*}$ is to solve a regularized optimization problem of the form

\begin{equation} \label{eq:variational:problem}
    \hat{\vx} \in \argmin_{\vx \in \R^{n}} d(\mA\vx, \vy) + \lambda g(\vx), \quad \lambda > 0,
\end{equation}

\noindent where $d$ is a data consistency term and $g$ a suitable regularization term. The data-fidelity term is typically chosen as the $\ell_2$ norm, \ie $d(\mA\vx, \vy) = \frac{1}{2}\|\mA\vx - \vy\|_2^2$, which corresponds to the negative log-likelihood of the measurements under the Gaussian noise assumption. The regularization term encodes prior knowledge on the unknown signal, such as sparsity in a given basis \cite{mallatWaveletTourSignal1999} or smoothness \cite{getreuerRudinOsherFatemiTotalVariation2012}.
\noindent 

\paragraph{Post-processing} Let us denote by $\operatorname{D}_\phi$ a deep neural network with parameters $\phi$. Post-processing methods aim to learn a direct mapping from a fast, low-quality reconstruction, \eg $\mA^\top \vy$ or $\mA^\dagger \vy$, to the ground-truth signal $\vx^{*}$. When dealing with large volumes, \eg $512^3$ voxels, a popular approach is to leverage patch-based processing to mitigate the cost of training:

\begin{equation} \label{eq:postp}
    \begin{aligned}
        &\mathcal{L}_{\text{POST}}(\phi) = \E_{\rmS}\E_{\rvx^*,\rvy }~ \| \operatorname{D}_\phi(\rmS \rvx_0) - \rmS \rvx^{*} \|_2^2, \\
    \end{aligned}
\end{equation}

\noindent where $\vx_0 = \mA^\top \vy \text{ or } \mA^\dagger \vy$ can be pre-computed before training, and $\rmS \in \R^{p \times n}$ is the random linear operator that extracts patches in $\R^p$ during training.

\paragraph{Unrolled methods} 

Deep unrolled or deep unfolded optimization \cite{adlerSolvingIllposedInverse2017,sriram_end--end_2020,dingLowDoseCTDeep2020} can be seen as a form of data-driven regularization. The general idea is to replace the hand-crafted regularization term $g$ by a learnable prior, parameterized by a deep neural network and trained end-to-end as part of an iterative procedure. It produces state-of-the-art results with fewer iterations than PnP methods \cite{kiss_benchmarking_2025}.

Proximal Gradient Descent (PGD) \cite{combettesProximalSplittingMethods2010} is a popular choice for unrolling, where the proximal step is replaced by a learned operator

\begin{equation} \label{eq:unrolled}
 \vx_{k+1} = \operatorname{D}_{\phi}\Big(\vx_{k}- \eta \nabla_{\vx_{k}} d(\mA\vx_{k}, \vy) \Big), \quad \eta > 0.
\end{equation}

\noindent The reconstruction function $\operatorname{R}_\phi: \R^{m} \times \R^{m \times n} \to \R^{n}$ is defined as the output of $K$ iterations of the procedure and is trained to minimize the following loss:
\begin{equation} \label{eq:unrolled:loss}
    \begin{aligned}
        &\mathcal{L}_{\text{UNR}}(\phi) = \E_{\rvx^{*},\rvy}~ \| \operatorname{R}_\phi(\rvy, \mA) - \rvx^{*} \|_2^2 \\
        &\text{with } \operatorname{R}_\phi(\rvy, \mA) = \rvx_{K}(\phi), ~ \rvx_0 = \mA^\top \rvy \text{ or } \mA^\dagger \rvy. \\
    \end{aligned}
\end{equation}

\noindent While post-processing networks (\ref{eq:postp}) can be easily scaled to large 3D volume by training on smaller patches (e.g. $64^3$ voxels), this simple idea does not apply to unrolled networks (\ref{eq:unrolled}), which require training on full volumes.

More specifically, gradient computation during post-processing training does not involve the normal operator $\mA^\top \mA$:
\begin{equation}
    \nabla_{\phi} \mathcal{L}_{\text{POST}} = {\frac{\partial \mathcal{L}_{\text{POST}}}{\partial \hat{\vx}}}^\top \frac{\partial \operatorname{D}_\phi}{ \partial \phi}\bigg|_{\vx_0}.
\end{equation}

\noindent On the other hand, the gradient $\nabla_\phi\mathcal{L}_{\text{UNR}}$ accumulates contributions from all iterations, especially the data-step, which requires the full volume to be evaluated:
\begin{equation}
    \begin{aligned}
    &\frac{\partial \mathcal{L}_{\text{UNR}}}{\partial \vx_{k+1}} = \frac{\partial \mathcal{L}_{\text{UNR}}}{\partial \vx_K} \prod_{j=k+1}^{K-1} \frac{\partial \vx_{j+1}}{\partial \vx_{j}}, \\
    &\text{with } \frac{\partial \vx_{j+1}}{\partial \vx_{j}} = \frac{\partial \operatorname{D}_\phi}{\partial \vx_j} (\bm{I} - \eta \mA^\top\mA). \\
    \end{aligned}
\end{equation}

\noindent \textbf{Block-separable inverse problems} \quad Consider a non-overlapping decomposition of the signal $\vx = [\vx_0^\top, \ldots, \vx_{B}^\top]^\top, ~\vx_{b} \in \R^{n_b}$. If $\mA$ admits a block-diagonal structure then the following can be evaluated efficiently:

\begin{equation}
    \begin{aligned}
    &\mA_b \vx_b = \vy_b + \bm{\varepsilon}_{b}, \quad \mA_b \in \R^{m \times n_b}, \quad \forall b \in \llbracket 1, B \rrbracket \\
    &\mA \vx = [(\mA_0 \vx_0)^\top, \ldots, (\mA_B \vx_B)^\top]^\top.
    \end{aligned}
\end{equation}

\noindent When dealing with large-scale problems, such a decomposition allows breaking down the reconstruction task into $B$ smaller problems that can be solved independently. For deep learning applications \cite{sunBlockCoordinateRegularization2019}, a particularly interesting decomposition is one where each subspace $\R^{n_b}$, preferably with $n_b \ll n$, corresponds to a rectangular or cuboid patch of the original volume. 

\begin{equation} \label{eq:coord:unrolled}
    \hat{\vx}_b = \operatorname{R}_{\phi}(\vy_b, \mA_b) ~ \forall b \in \llbracket 1, B \rrbracket, \quad \hat{\vx} = \sum_{b=1}^{B} \mS_b^\top \hat{\vx}_b,
\end{equation}

\noindent where $\mS_b \in \R^{n_b \times n}$ is the selection operator that extracts the $b$-th patch from the full volume. However, such a decomposition does not often exist in real-world scenarios, especially in the case of 3D problems where the forward operator often couples all the voxels together. In \cref{sec:method}, we propose two strategies to alleviate this issue, independently of the existence of a coordinate-friendly decomposition.
\section{Method} \label{sec:method}

In this section, we present a novel approach to scale unrolled networks to arbitrarily large linear inverse problems. We first introduce a \textbf{domain partitioning strategy} that allows us to reduce the size of the problem at train-time, which in turn permits the use of arbitrarily small networks that fit with memory constraints. Second, we introduce an \textbf{approximation} of the normal operator $\mA^\top \mA$ as a \textbf{product of diagonal and circulant matrices} that allows to compute sub-domains in a memory efficient manner. Relying on efficient implementations of the FFT, this approximation allows a significant speed-up of the data-consistency update. A circulant matrix evaluation can be performed exactly as a diagonal product in Fourier.  We first present each technique independently, then discuss how they can be combined to solve large-scale 3D inverse problems.

\subsection{Domain partitioning} \label{subsec:domain:partitioning}

Consider the decomposition of $\R^{n}$ into 2 orthogonal subspaces

\begin{equation} \label{eq:decomposition}
    \R^{n} = \R^{p} \oplus \R^{q} \quad \text{with } q = n -p.
\end{equation}

\noindent We define the matrices $\mS \in \R^{p \times n}$ and $\mS_\perp \in \R^{q \times n}$ which extracts a vector in $\R^{p}$, respectively in $\R^{q}$, from $ \R^{n}$. % by padding it with zeros.

\noindent Using this decomposition, we assume that when solving the problem in (\ref{eq:inverse:problem}), we already have part of the solution, \ie instead of seeking $\vx^{*} \in \R^{n}$, we want to recover the unknown $\vx_{\text{patch}} \in \R^{p}$ such that

\begin{equation} \label{eq:x:decomposition}
    \vx^{*} = \mS^\top \vx_{\text{patch}} + \mS_{\perp}^\top \vx_{\text{context}},
\end{equation}

\noindent where $\vx_{\text{context}} \in \R^{q}$ is known. Typically, we choose $\vx_{\text{patch}}$ as a rectangular or cuboid patch. This allows us to rewrite the linear system in (\ref{eq:inverse:problem}) as

\begin{equation} \label{eq:ip:decomposition}
    \tilde{\vy} = \tilde{\mA} \vx_{\text{patch}}, 
\end{equation}

\noindent where $\tilde{\mA} = \mA \mS^\top$ and $\tilde{\vy} = \vy - \mA\mS_{\perp}^\top \vx_{\text{context}}$. 

\begin{figure}[h]
    \centering
    \includegraphics[width=\linewidth]{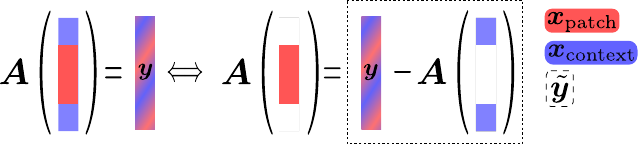}
    \caption{Domain partitioning strategy: in case of a forward operator $\mA$ that mixes the signal $\vx$ in a non-trivial manner, we can still decompose the full domain $\R^{n}$ into two orthogonal subspaces $\R^{p}$ and $\R^{q}$. Then by linearity, we solve for the unknown smaller patch $\vx_{\text{patch}} \in \R^{p}$ (\textit{red}) given the context $\vx_{\text{context}} \in \R^{q}$ (\textit{blue}).}
    \label{fig:domain:partitioning}
\end{figure}

\noindent We use the decomposition in (\ref{eq:ip:decomposition}) to reduce the large-scale problem into arbitrarily small ones. In the context of supervised learning, we choose $\vx_{\text{context}} = \mS_{\perp}\vx^{*}$.
Akin to patch-based training, we vary the position of the subspace $\R^p$  at random and minimize the following loss:

\begin{equation} \label{eq:patch:training}
    \begin{aligned}
        &\mathcal{L}_{\text{PART}}(\phi) =\E_{\rmS}\E_{\rvx^*, \rvy}~ \| \operatorname{R}_{\phi}(\tilde{\rvy}, \tilde{\mA}) - \rmS \rvx^{*} \|_2^2, \\
        &\text{with } \tilde{\mA} = \mA \mS^\top, \tilde{\vy} = \vy - \mA\mS_{\perp}^\top \mS_{\perp} \vx^{*},
    \end{aligned}
\end{equation}

\noindent where $\mS \vx^{*} \in \R^{p}$ is the ground-truth patch corresponding to the subspace $\R^{p}$.

\paragraph{Test-time without ground-truth context} We deploy the network $\operatorname{R}_\phi$, trained with \textit{domain partitioning}, in a two-step procedure: 
\begin{itemize}[labelindent=0.5em]
    \item[\textbf{1.}] At test-time, we can mitigate the complexity of evaluating the prior step $\operatorname{D}_\phi$ on the full volume by evaluating it sequentially along patches and merging the processed patches. Thus, we obtain an estimation $\tilde{\vx}$ of the ground-truth signal by solving the full problem in (\ref{eq:inverse:problem}). We use a standard unrolled scheme, denoted by $\tilde{\vx}= \operatorname{R}_\phi(\vy, \mA)$, where the data-step is computed on the whole volume, \ie without partitioning, and the prior step $\operatorname{D}_\phi$ is performed patch-by-patch.
    \item[\textbf{2.}] We use the estimation $\tilde{\vx}$ to build a context within our \textit{domain-partitioned} framework, \ie $\vx_{\text{context}} = \mS_{\perp}\tilde{\vx}$, and independently solve each subproblem using the same network $\operatorname{R}_\phi$.
\end{itemize} 

\noindent For the CBCT experiments, we observe that this two-step refining procedure consistently improves the reconstruction quality compared to using only the initial estimation $\tilde{\vx}$. For the MC-MRI experiments, we observe that the first estimation $\tilde{\vx}$ is already of high-quality, thus the second refinement step brings negligible improvements. A summary of the test-time algorithm is provided in \cref{alg:test:patch:unrolling}, and we give more details in \cref{appendix:test:details}.

\SetKwInput{KwInit}{init}
\SetKwFunction{patches}{patches}
\SetKwFunction{append}{append}
\SetKwFunction{aggregate}{aggregate}

\DontPrintSemicolon 
\begin{algorithm}[hbt!] 
   \DontPrintSemicolon
   \caption{Test-time domain partitioned inference} \label{alg:test:patch:unrolling}
   \KwInit{$\vy \in \mathbb{R}^m$, $\mA \in \R^{m \times n}$, $\vx_0 = \mA^\dagger \vy \in \mathbb{R}^n$, $k=0, K > 0$, step size $\eta > 0$, $\operatorname{R}_\phi$.}

    \vspace*{0.25em}
   \emph{First unrolling procedure to get $\tilde{\vx} \approx \vx^{*}$ } \\

    $\tilde{\vx} \leftarrow \operatorname{R}_\phi(\vy, \mA)$ \\
    \vspace*{0.25em}
    
    \emph{Domain partitioned computation} \\
    $X_{\mathrm{patches}} = [~]$ \\
    \For{$\mS$ in \patches{}}{
        \KwInit{$\vx_{\text{context}} = \mS_{\perp}\tilde{\vx}$,~ $\tilde{\mA} = \mA \mS^\top$, 
        $\tilde{\vy} = \vy - \mA\mS_{\perp}^\top \mS_{\perp} \tilde{\vx}$~~ (\ref{eq:ip:decomposition})}
        $\hat{\vx}_{\text{patch}} \leftarrow \operatorname{R}_{\phi}(\tilde{\vy}, \tilde{\mA})$ \\
        \append{$X_{\mathrm{patches}}$, $\hat{\vx}_{\text{patch}}$} \\
    }
    $\hat{\vx} \leftarrow$ \aggregate{$X_{\mathrm{patches}}$}
\end{algorithm}

\subsection{Normal operator approximation} \label{subsec:normal:operator:approximation}

We previously introduced the domain partitioning decomposition where $\tilde{\vy} = \tilde{\mA} \vx_{\text{patch}}$ with $\tilde{\mA} = \mA \mS^\top$. Note here that solving (\ref{eq:ip:decomposition}) still requires the global forward operator and its adjoint, \ie through the evaluation of $\tilde{\mA}^\top \tilde{\mA} = \mS \mA^\top \mA \mS^\top$. In this section, we introduce an efficient approximation of the normal operator $\mA^\top \mA$ that significantly reduces the computational burden of naively evaluating $\tilde{\mA}^\top \tilde{\mA}$.

Let us recall the data-consistency update typically computed in the form of a gradient descent step:

\begin{equation} \label{eq:gradient:step}
    \centering
    \begin{aligned}
        h(\vx) & = \vx - \eta \nabla_{\vx} d(\mA\vx, \vy) \\
               & = \vx - \eta (\mA^\top \mA) \vx + \mA^\top \vy.
    \end{aligned}
\end{equation}

\noindent Discarding the constant term $\mA^\top \vy$ which can be pre-computed, we see that a computing (\ref{eq:gradient:step}) only requires the evaluation of the normal operator $\mA^\top \mA \vx$.

\begin{remark}
    Focusing on the evaluation of the normal operator $\mA^\top \mA$ is not restrictive. In the case of the data consistency step being a proximal step, \ie $h(x) = \mathrm{prox}_{\eta d(\mA \cdot, \vy)}(\vx)$, $\mA^\top \mA$ is again the main component in most popular solvers, \eg conjugate gradient.
\end{remark}

\paragraph{Translation-equivariant operators} Assuming that $\mA$ is translation-equivariant, then $\mA^\top \mA$ is a convolutional operator \cite{Vetterli_Kovačević_Goyal_2014}. In this case, we can leverage the convolution theorem to efficiently compute the normal operator evaluation in the Fourier domain. More precisely, we can write
\begin{equation} \label{eq:fft:normal:operator}
    \mA^\top \mA \vx = \mF^{-1} \diag(\bm{\lambda}) \mF \vx,
\end{equation}

\noindent where $\mF$ and $\mF^{-1}$ are the Fourier and inverse Fourier transforms, respectively, and $\bm{\lambda} \in \sC^{n}$ is the frequency response of the convolution kernel associated with $\mA^\top \mA$.

\paragraph{Spatial modulation} We can generalize the factorization to a larger class of non-translation equivariant operators, \eg inpainting, by modulating the output of the convolution by a diagonal operation in the spatial domain. More precisely, we factorize the normal operator as
\begin{equation} \label{eq:approx:normal:operator}
    \mA^\top \mA = \mH = \diag(\vm) \mF^{-1} \diag(\bm{\lambda}) \mF,
\end{equation}

\noindent where $\vm \in \R^{n}$ is homogeneous to a spatial sensitivity map, or mask. 

Following Schmid \etal \cite{schmid_decomposing_2000}, we could increase the expressivity of the approximation by adding more diagonal-circulant factors, \ie $\mH = \prod_{i=1}^N \diag(\vm_i) \mF^{-1} \diag(\bm{\lambda}_i) \mF$. However, in practice we observe that a single factorization ($N=1$) is sufficient to obtain good reconstruction results while keeping the computational cost low.

\begin{remark}
    Note that in (\ref{eq:approx:normal:operator}) we approximate the symmetric operator $\mA^\top \mA$ with a non-symmetric factorization $\mH$. In practice, we observe that imposing symmetry constraints during the fitting procedure does not improve the quality of the approximation nor the final reconstruction results. At test-time, using either $\mH$ or its symmetrized version $\frac{1}{2}(\mH + \mH^\top)$ yields similar results, thus we keep the simpler version $\mH$ for efficiency.
\end{remark}

\paragraph{Examples} 
For CT, the Fourier slice theorem states that each row of the forward operator corresponds to sampling a \textit{radial} line in the Fourier domain \cite{fesslerAnalyticalTomographicImage2021}, which is exactly written as a diagonal operation in frequency space. The spatial modulation $\diag(\vm)$ can then be interpreted as a resampling map which compensates the \textit{cartesian} sampling done by the FFT, rather than a \textit{radial} computation. For other modalities, such as deconvolution problems, $\mA$ is exactly a convolutional operator and $\vm$ is simply an all-one vector. For inpainting problems, $\mA$ is a diagonal binary masking operator and $\mA^\top \mA = \mA$, which can be exactly represented by our proposed form in (\ref{eq:approx:normal:operator}) with $\bm{\lambda}$ being an all-one vector.

\begin{remark} \label{remark:mcmri:approximation}
    For multi-coil MRI, the normal operator writes $\mA^\top \mA = \sum_{c=1}^{C} \mS_c^\top \mF^{-1} \mM^\top \mM \mF \mS_c$, where $\mS_c$ is the sensitivity map of coil $c$ and $\mM$ is the undersampling mask in frequency space. In this case, the normal operator cannot be exactly represented by our proposed form in (\ref{eq:approx:normal:operator}). Indeed, the approximation breaks down to finding a single-coil equivalent of a multi-coil forward model. In practice, we observe that by slightly changing the factorization to $\mH = \diag(\vm)^H \mF^{-1} \diag(\bm{\lambda}) \mF \diag(\vm)$, we can obtain a good approximation of the multi-coil normal operator.
\end{remark}

\paragraph{Fitting the approximation} We fit the parameters $\vm$ and $\bm{\lambda}$ by gradient descent on a set of random vectors $\{\vx_i\}_{i=1}^N$. More precisely, we minimize the following loss:

\begin{equation} \label{eq:fitting:loss}
    \mathcal{L}(\vm, \bm{\lambda}) = \E_{\rvx \sim \mathcal{N}(\bm{0}, \mI)} \| \mA^\top \mA \rvx - \mH(\vm, \bm{\lambda}) \rvx \|_2^2.
\end{equation}

\noindent Using properties of standard Gaussian random vectors, note that (\ref{eq:fitting:loss}) is equal to the Frobenius norm of the residual, \ie

\begin{equation} \label{eq:frobenius:loss}
    \begin{aligned}
        \mathcal{L}(\vm, \bm{\lambda}) &= \E_{\rvx \sim \mathcal{N}(\bm{0}, \mI)} \| \mA^\top \mA \rvx - \mH(\vm, \bm{\lambda}) \rvx \|_2^2 \\
        &= \| \mA^\top \mA - \mH(\vm, \bm{\lambda}) \|_F^2
    \end{aligned}
\end{equation}

\noindent This means that we can fit the parameters $(\vm, \bm{\lambda})$ of the factorization without any problem-specific data, which is particularly useful when the forward operator $\mA$ is known but a limited amount of training data is available.

\subsection{Efficient approximation on partitioned domain}

\noindent When using the domain partitioning strategy presented in \cref{subsec:domain:partitioning}, we need to efficiently compute the normal operator evaluation on any patch $\vx_{\text{patch}} \in \R^{p}$, \ie $\tilde{\mA}^\top \tilde{\mA} \vx_{\text{patch}} = \mS \mA^\top \mA \mS^\top \vx_{\text{patch}}$. Using the approximation in (\ref{eq:approx:normal:operator}), we can write
\begin{equation} \label{eq:approx:normal:operator:patch}
    \tilde{\mA}^\top \tilde{\mA} \vx_{\text{patch}} \approx \mS \diag(\vm) \mF^{-1} \diag(\bm{\lambda}) \mF \mS^\top \vx_{\text{patch}}.
\end{equation}

\noindent Let us decompose (\ref{eq:approx:normal:operator:patch}) into finer steps: \textbf{(i)} $\mS^\top \vx_{\text{patch}}$ is a zero-padding operation, that materializes in memory a vector of size $n$, from a patch of size $p \ll n$, \textbf{(ii)} $\mF^{-1} \diag(\bm{\lambda}) \mF$ is equivalent to a convolution with a kernel of size $n$, \textbf{(iii)} $\mS\diag(\vm)$ is a diagonal operator of size $p$.

Here, note that both steps \textbf{(i)} and \textbf{(ii)} are highly inefficient as they require going back to the full signal volume. As the input of the convolution is a zero-padded cuboid, we can restrict the convolution kernel to a smaller size $k = 2p \ll n$ and maintain exact computation. More precisely, we can write
\begin{equation}
    \centering
    \begin{aligned}
        &~\mS \diag(\vm) \mF^{-1} \diag(\bm{\lambda}) \mF \mS^\top \vx_{\text{patch}} \\
        = &~\diag(\mS\vm) \mF_{k}^{-1} \diag(\bm{\lambda}_{k}) \mF_{k} \vx_{\text{patch}},
    \end{aligned}
\end{equation}

\noindent where $\mF_{k}$ and $\mF_{k}^{-1}$ are the Fourier and inverse Fourier transforms restricted to size $k$ (with zero-padding or cropping from $\R^p$ to $\R^k$), respectively, and $\bm{\lambda}_{k} \in \sC^{k}$ is the frequency response of the convolution kernel restricted to size $k$. 

\begin{figure*}[tb] \centering %\setkeys{Gin}{width=0.24\textwidth}

    \includegraphics[width=0.125\textwidth]{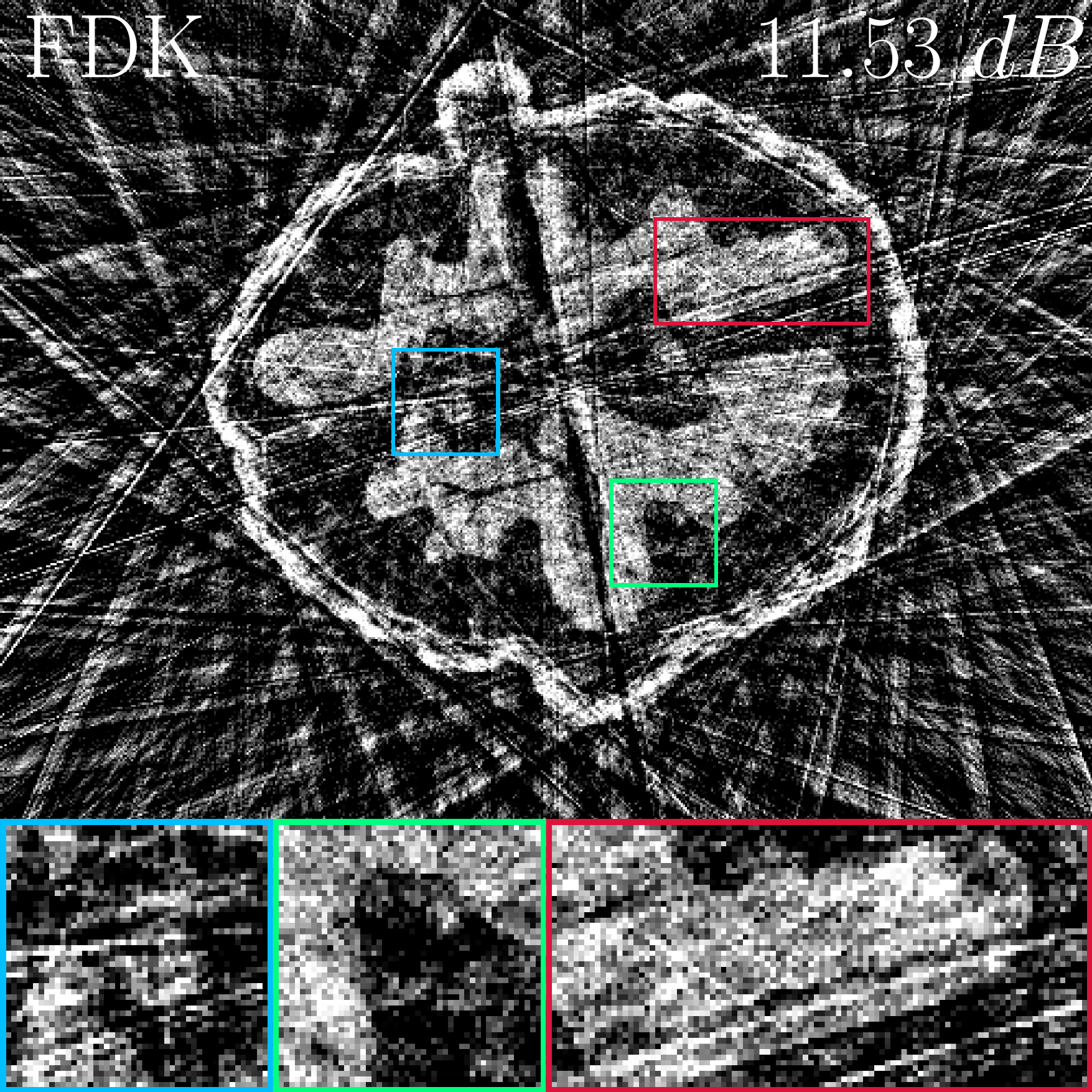}%
    \includegraphics[width=0.125\textwidth]{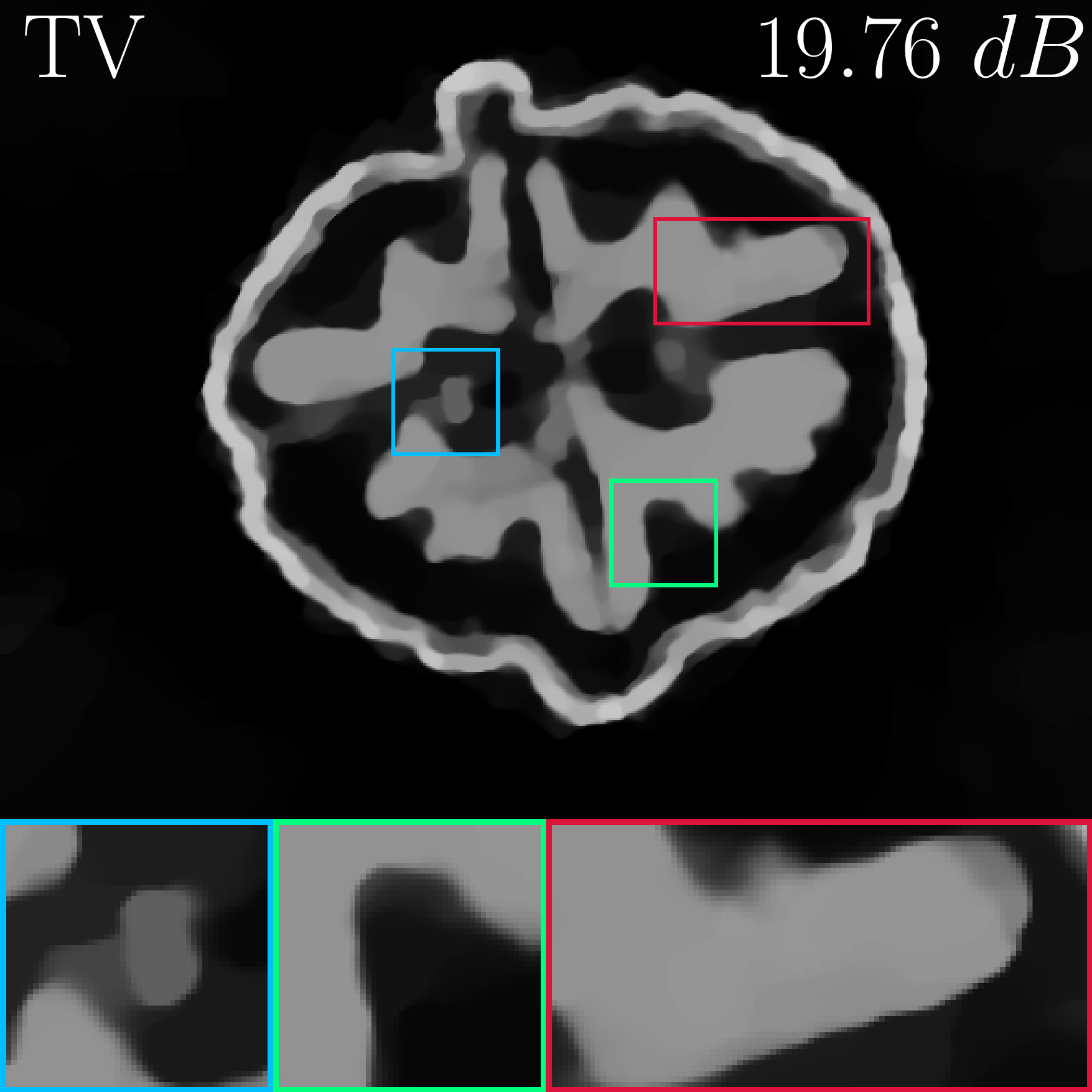}%
    \includegraphics[width=0.125\textwidth]{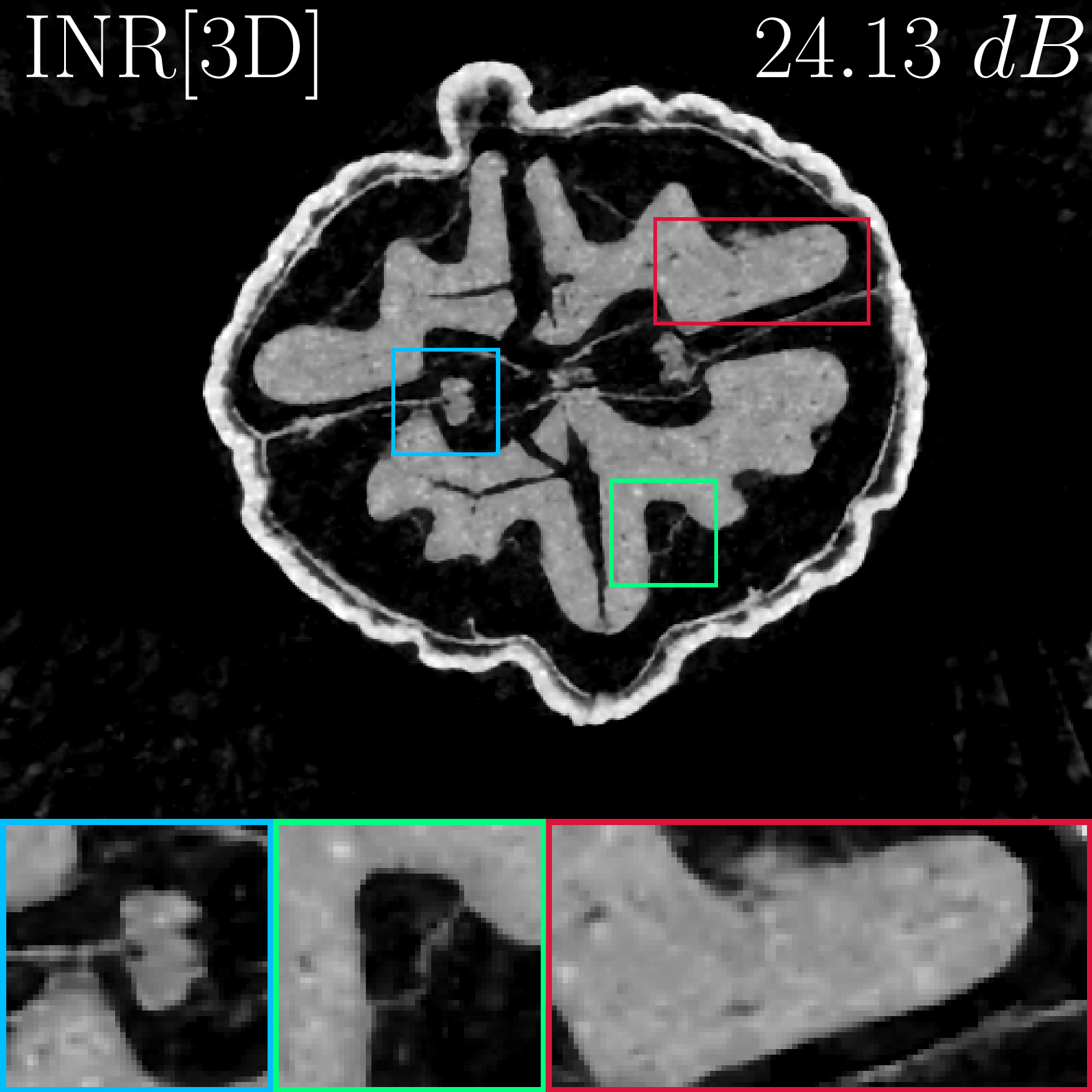}%
    \includegraphics[width=0.125\textwidth]{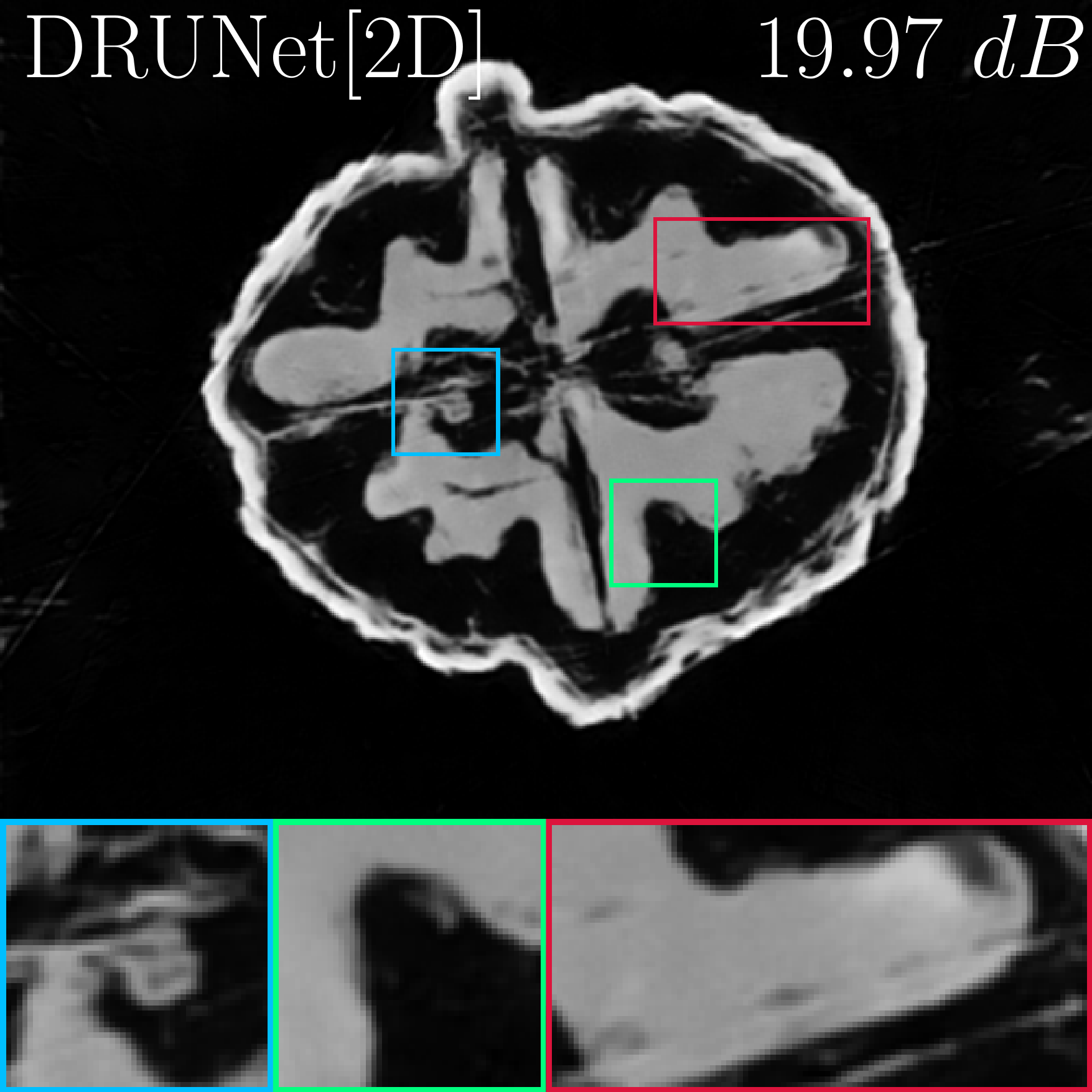}%
    \includegraphics[width=0.125\textwidth]{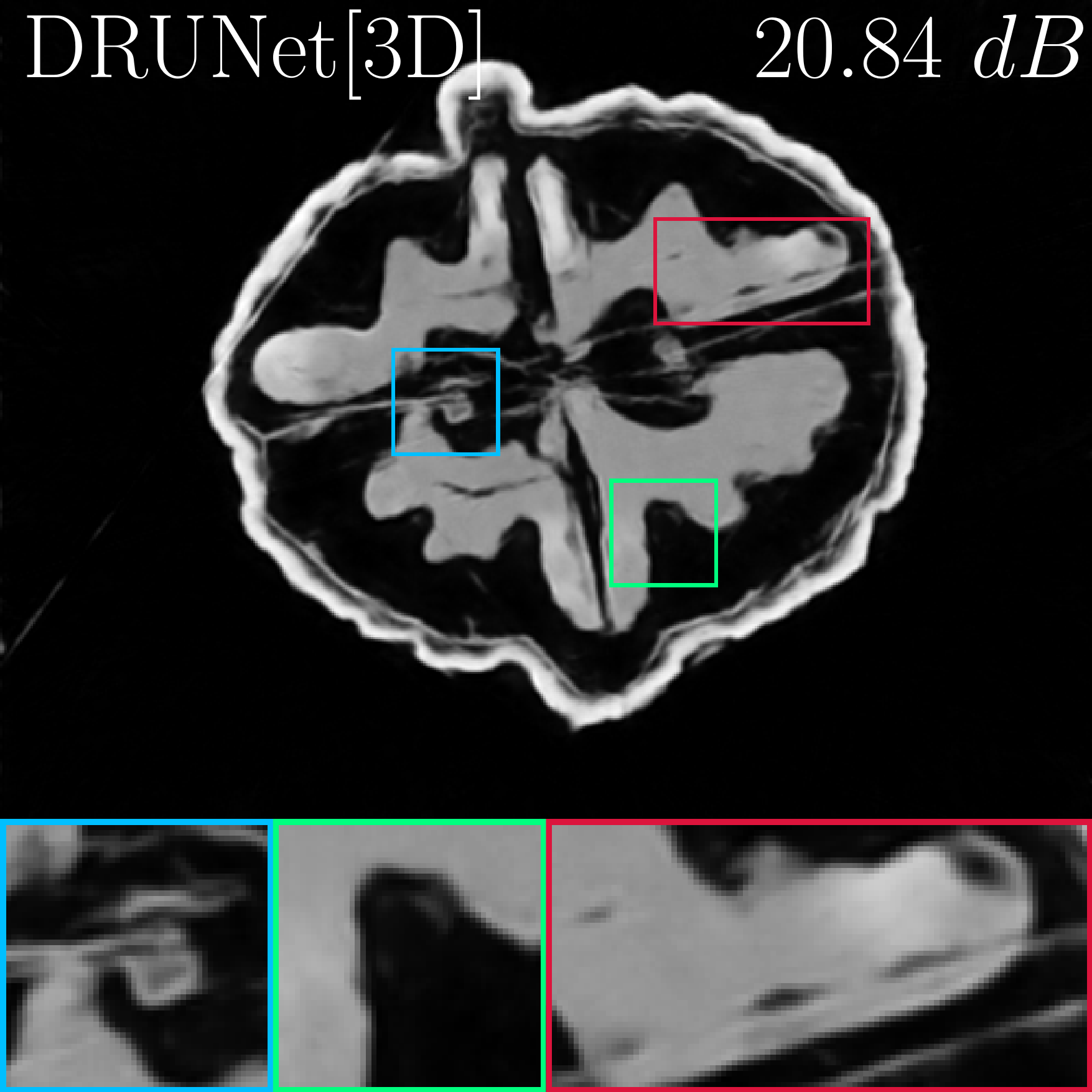}%
    \includegraphics[width=0.125\textwidth]{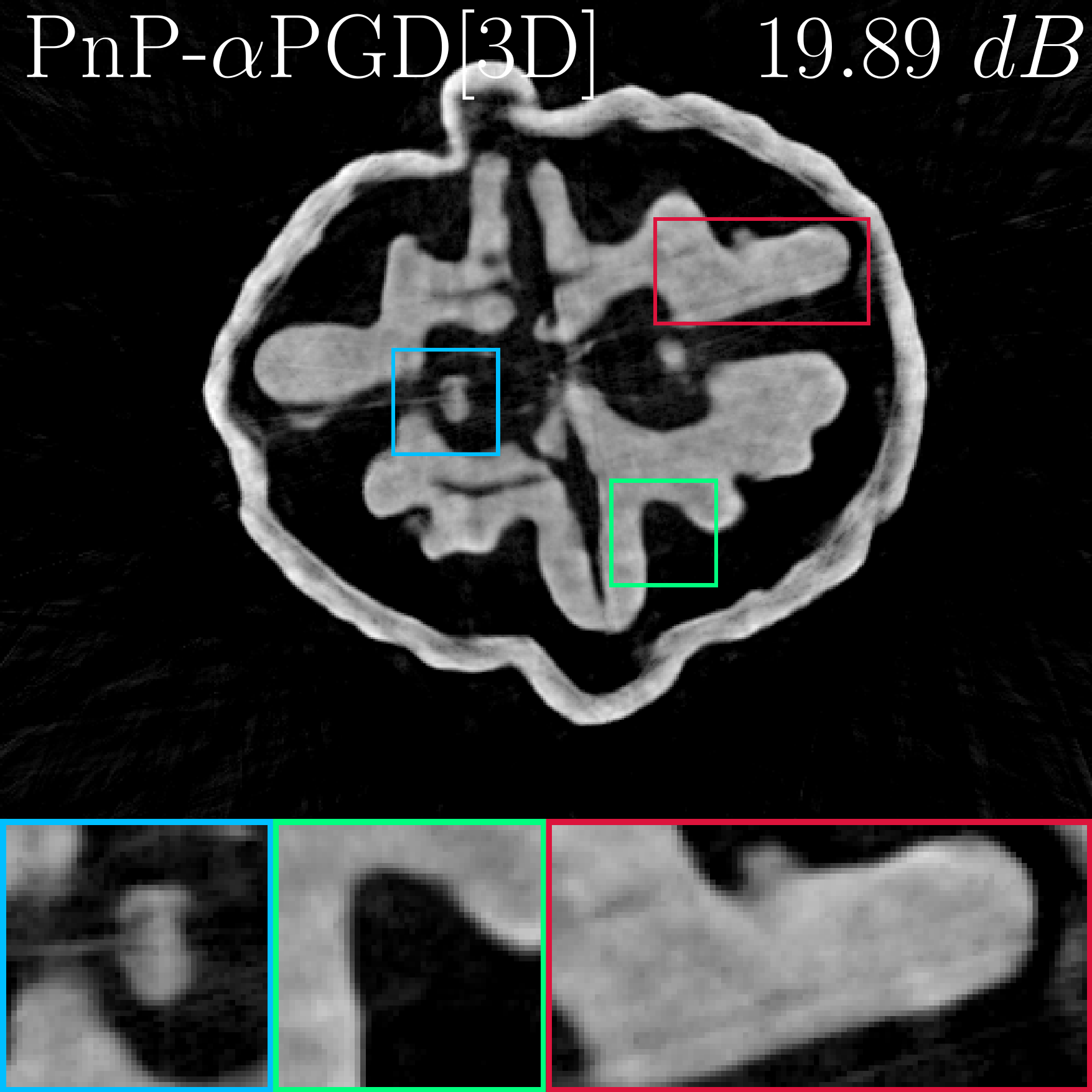}%
    \includegraphics[width=0.125\textwidth]{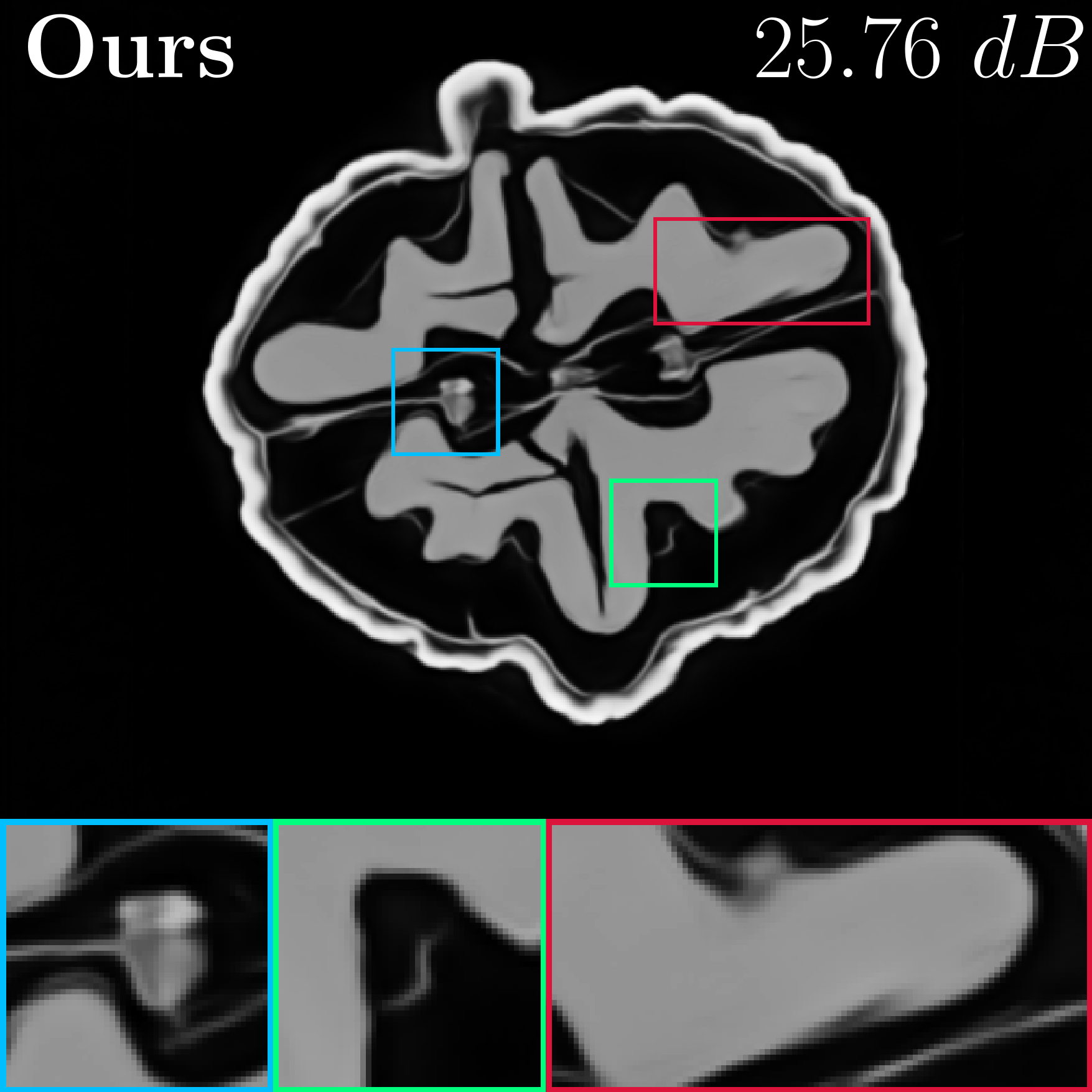}%
    \includegraphics[width=0.125\textwidth]{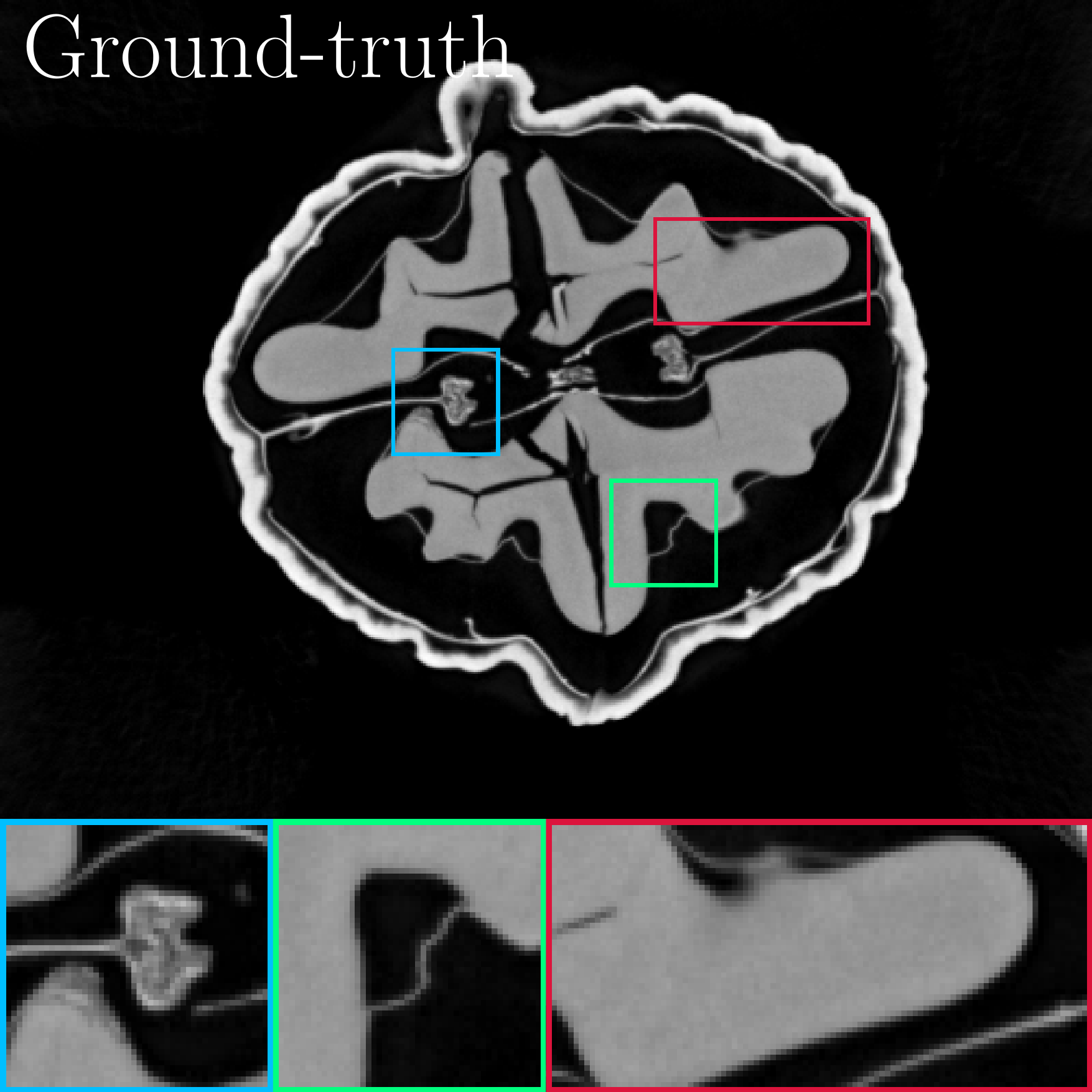}%

    \includegraphics[width=0.125\textwidth]{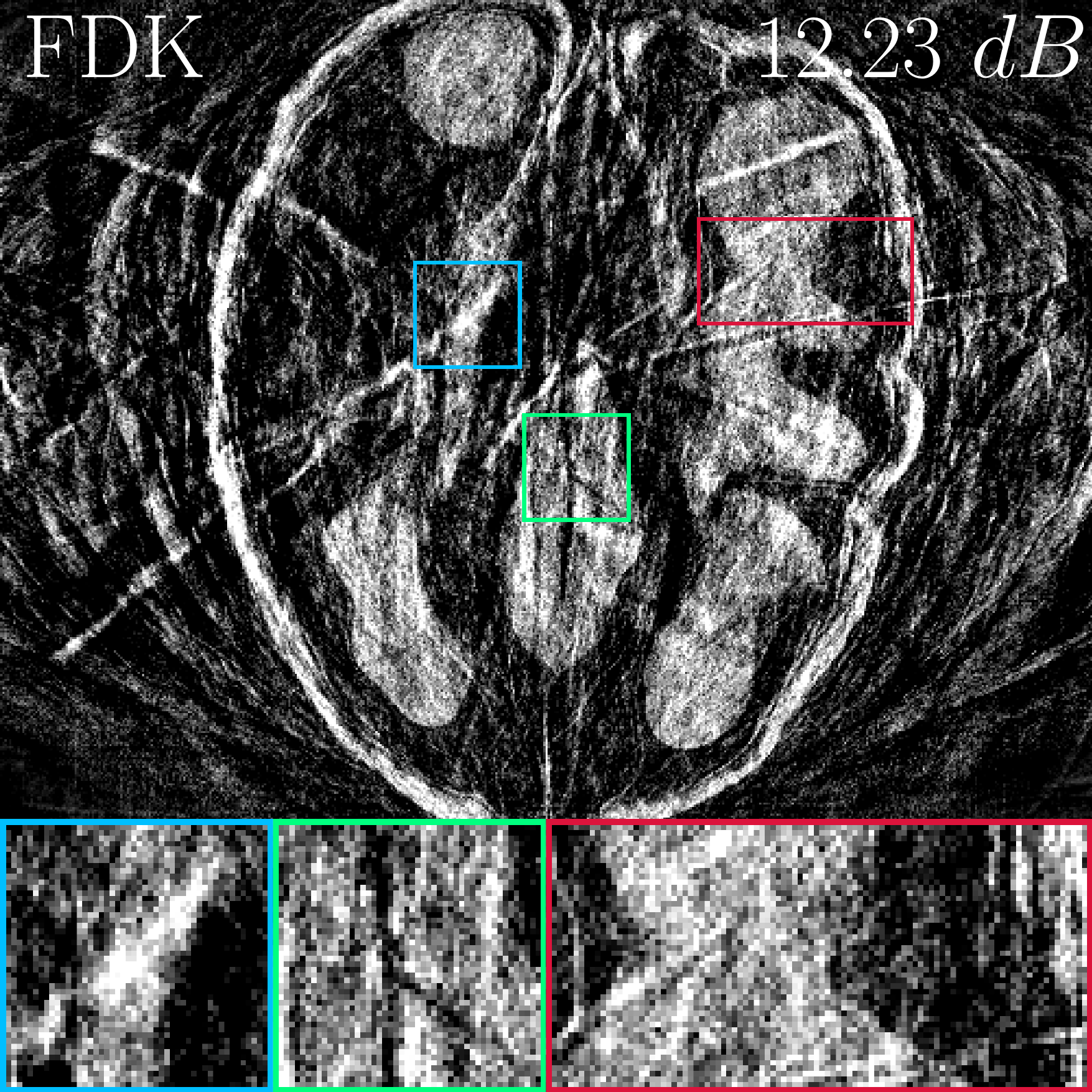}%
    \includegraphics[width=0.125\textwidth]{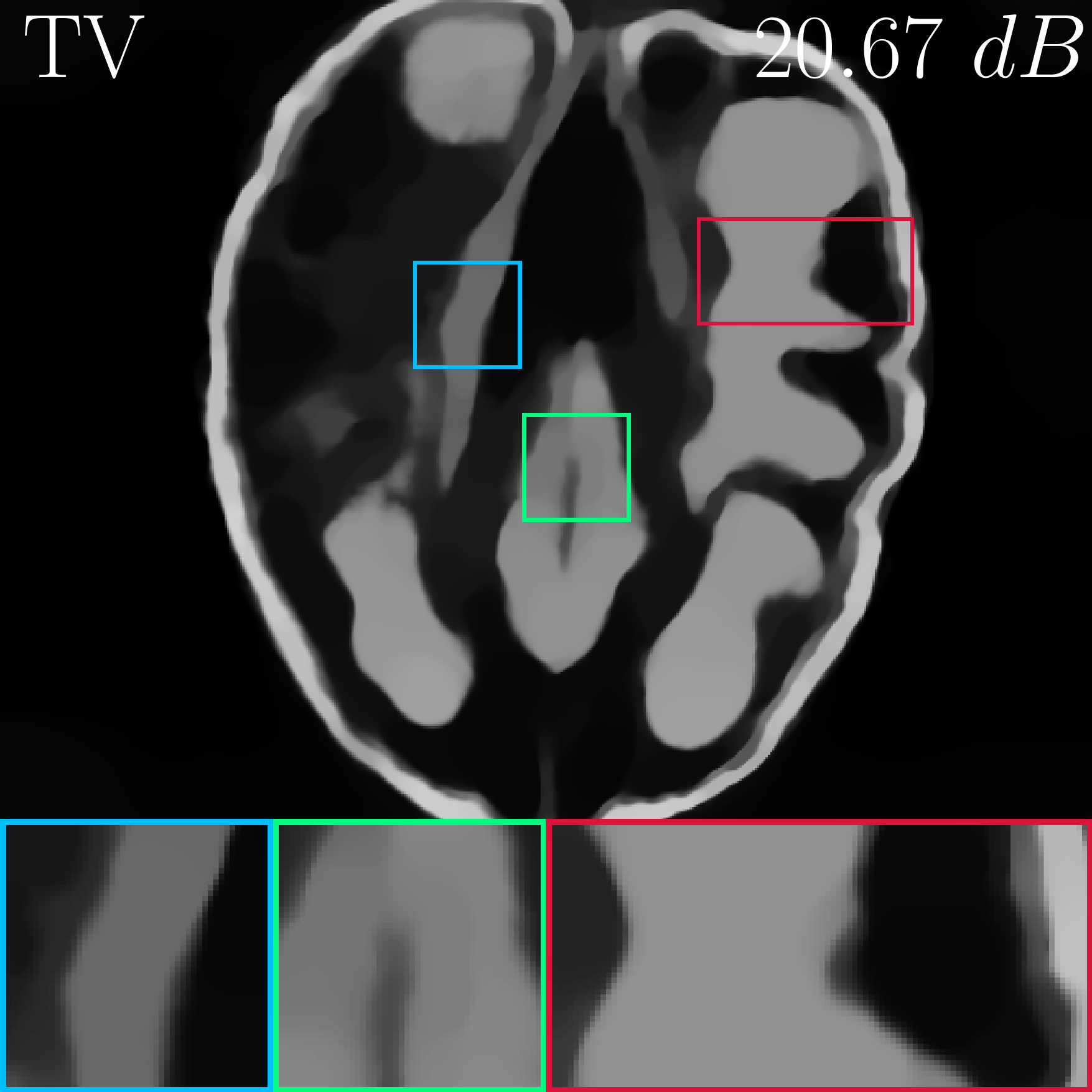}%
    \includegraphics[width=0.125\textwidth]{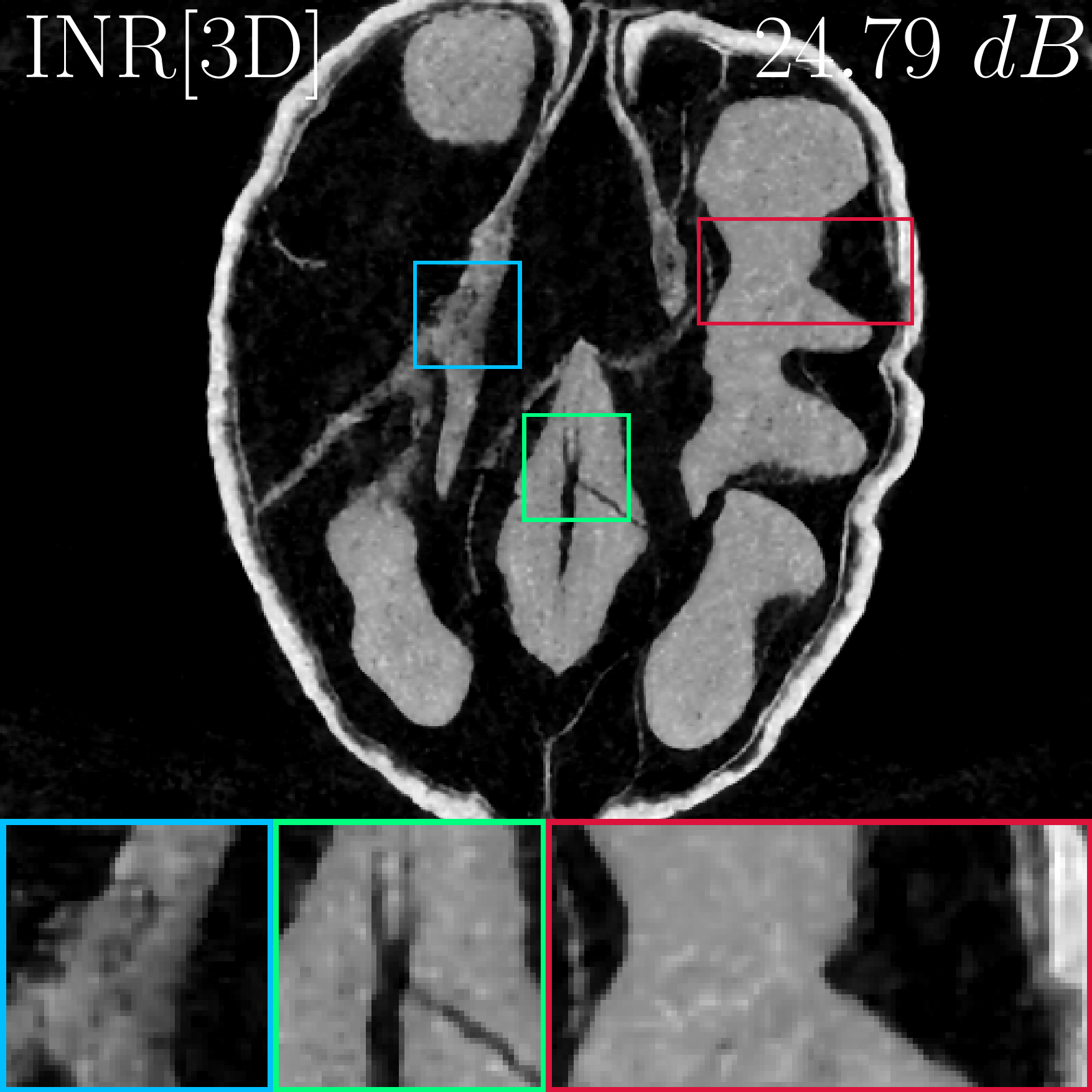}%
    \includegraphics[width=0.125\textwidth]{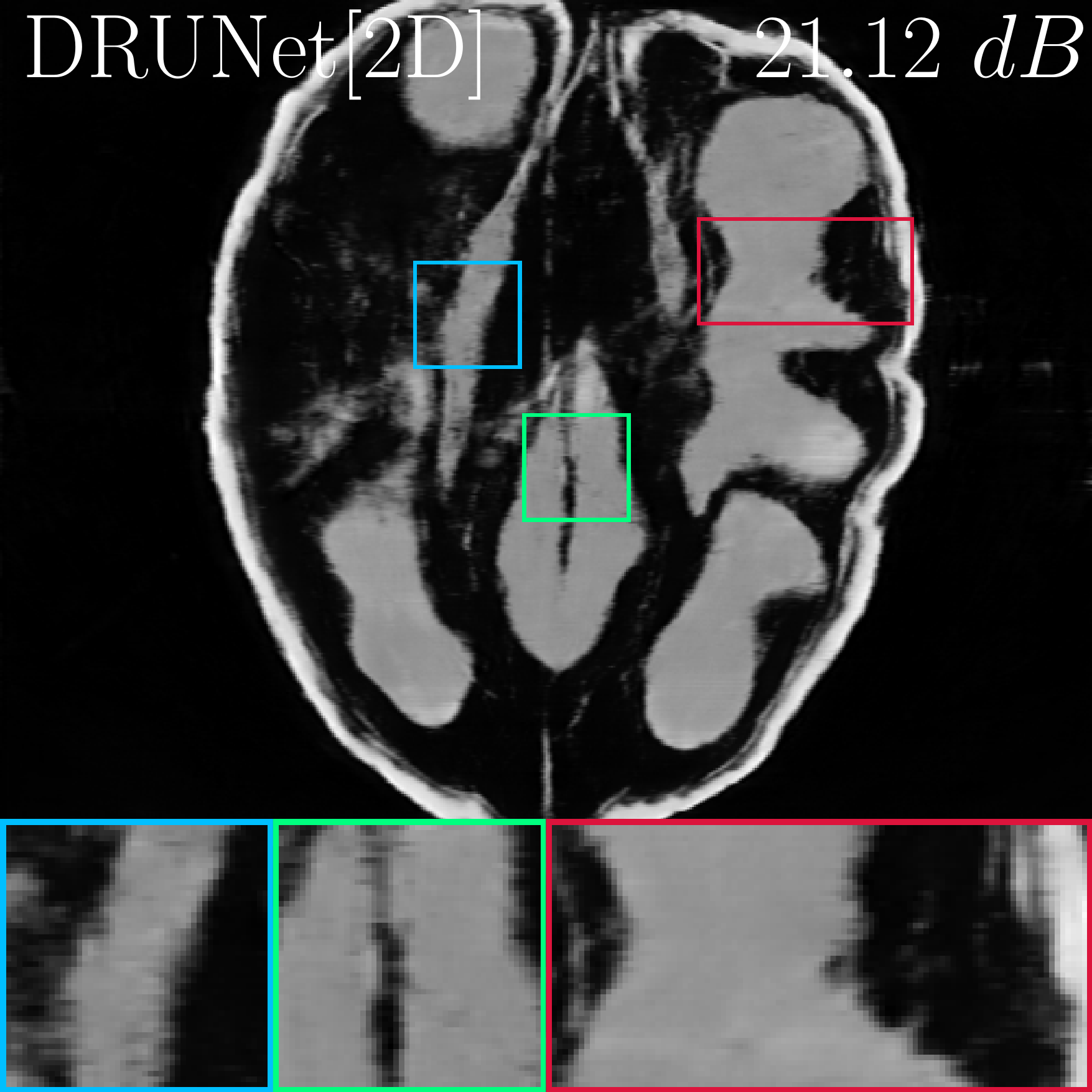}%
    \includegraphics[width=0.125\textwidth]{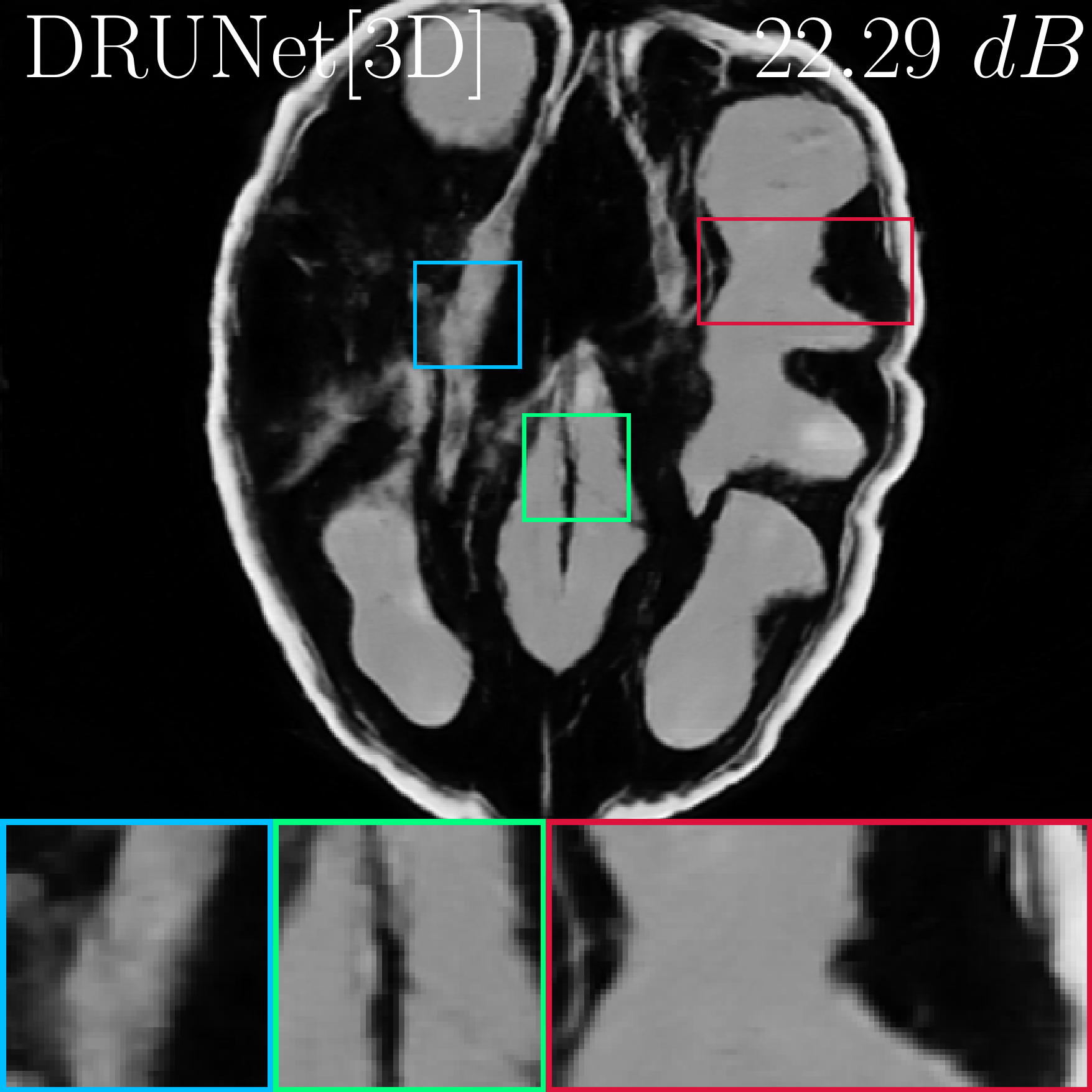}%
    \includegraphics[width=0.125\textwidth]{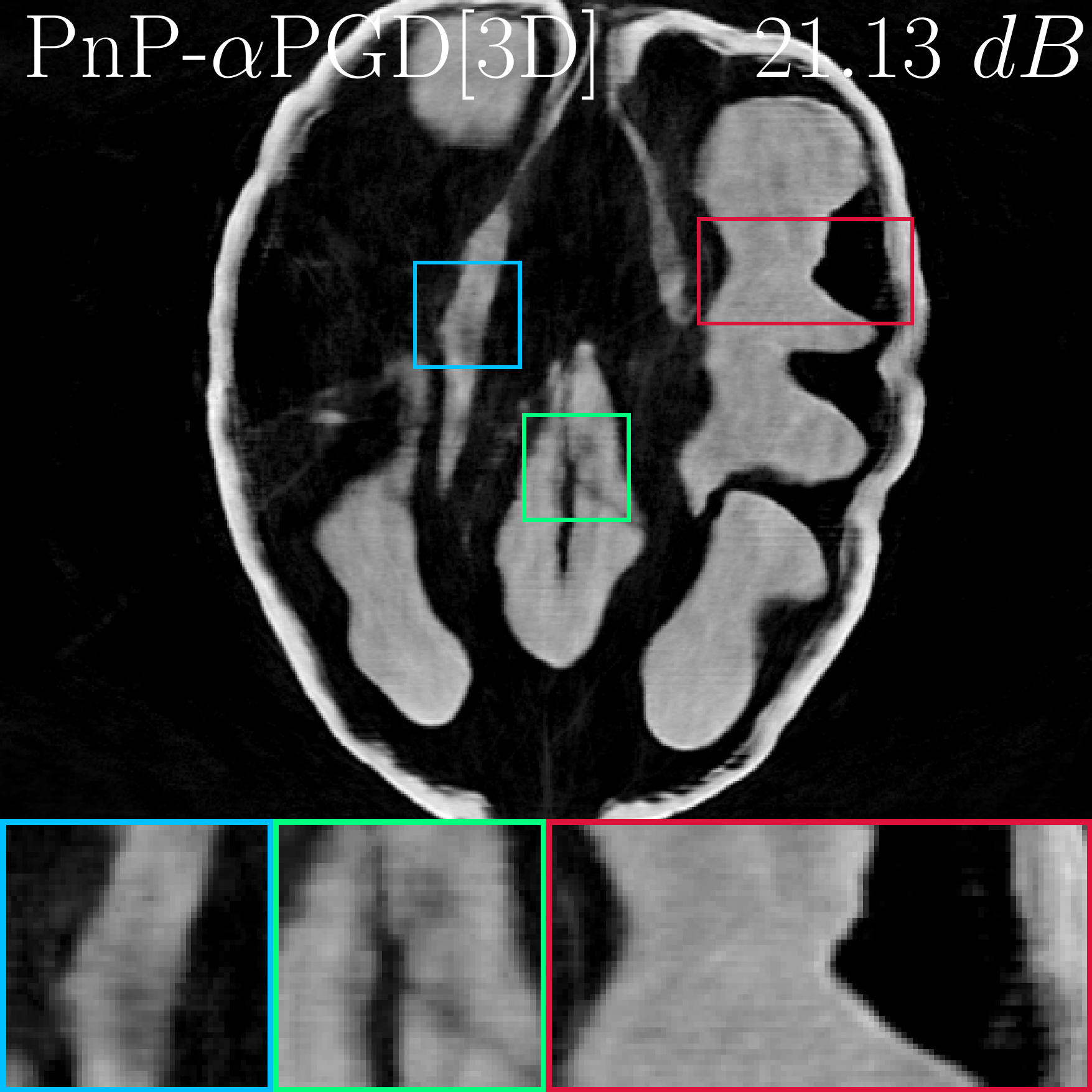}%
    \includegraphics[width=0.125\textwidth]{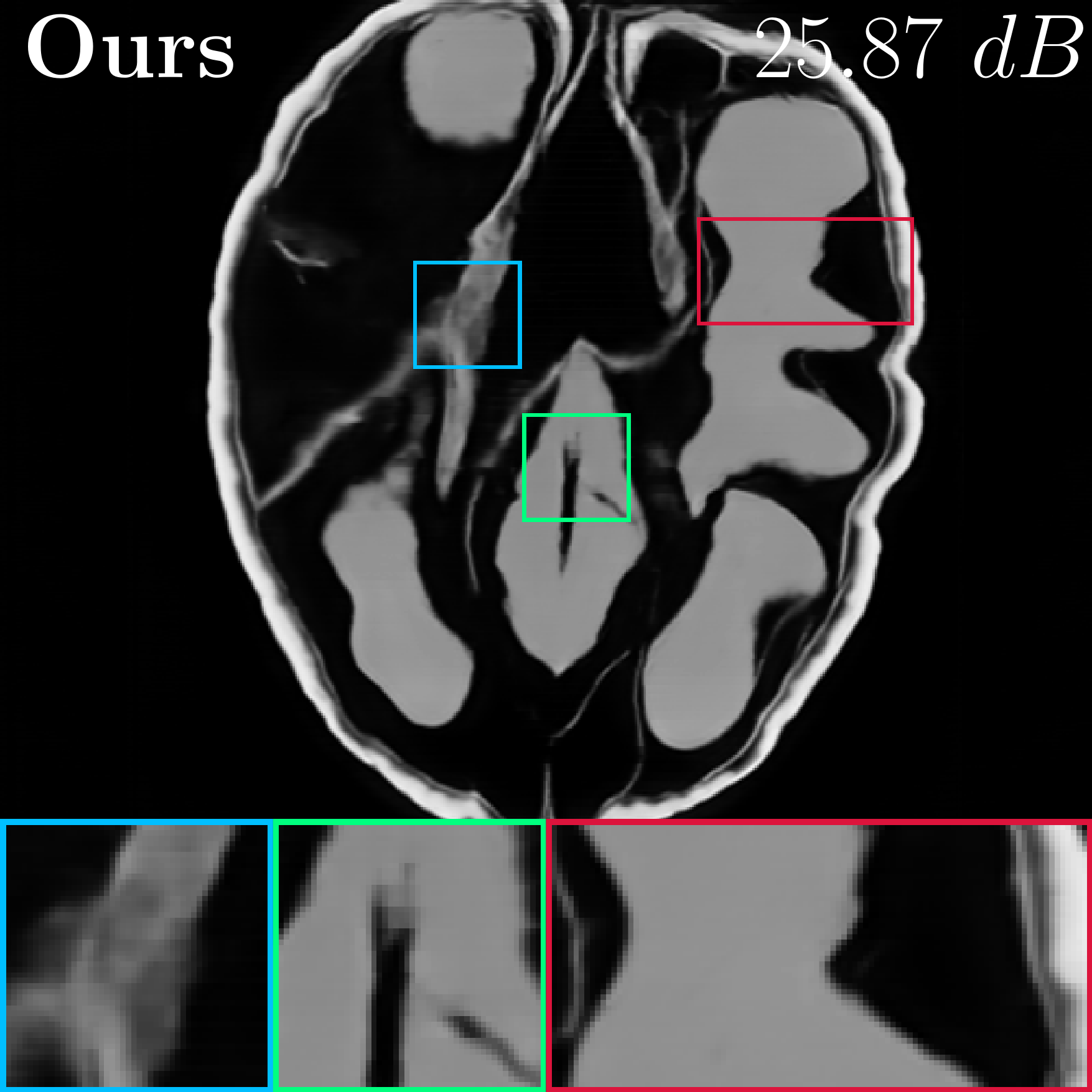}%
    \includegraphics[width=0.125\textwidth]{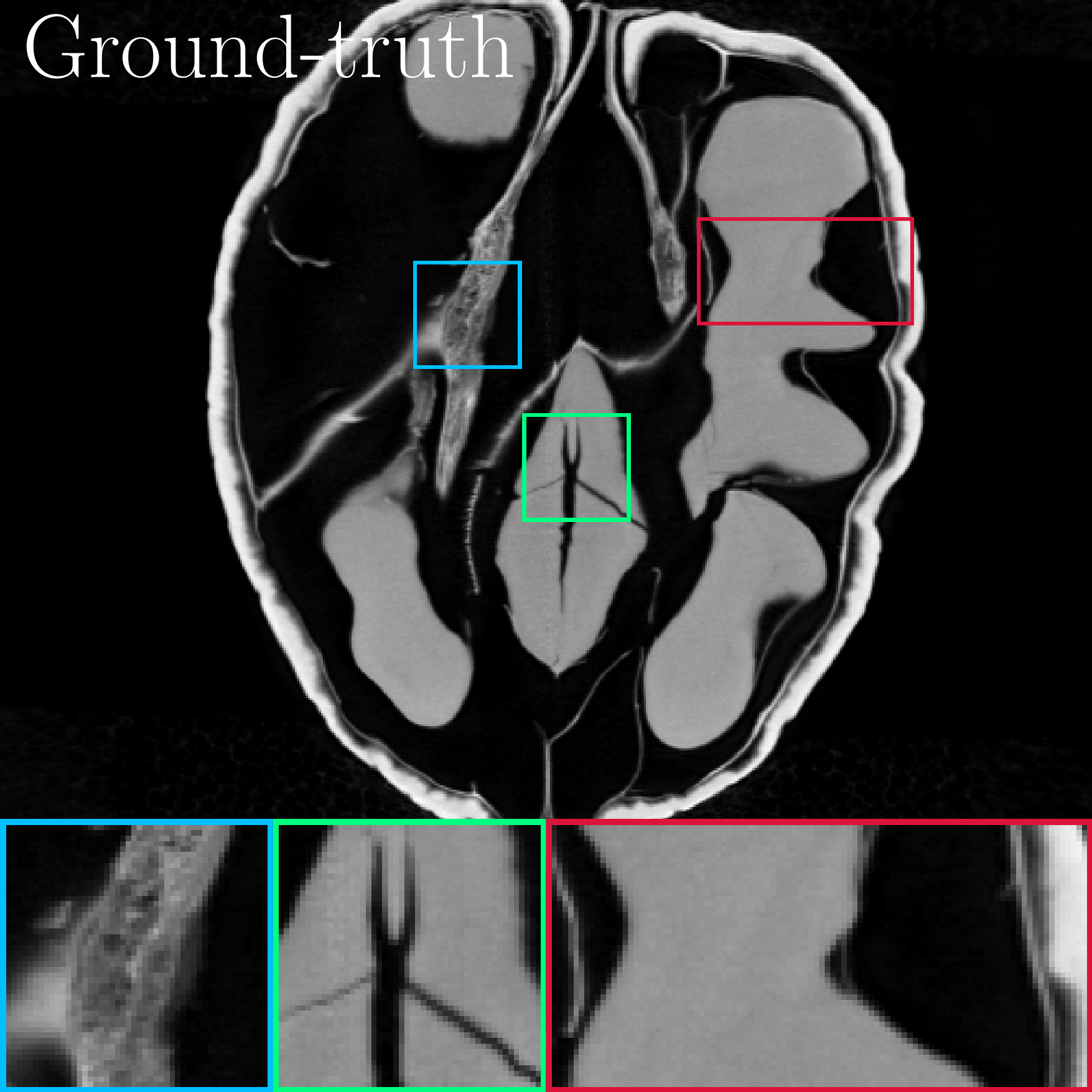}%

    \caption{Illustrations of sparse view reconstructions with [30/1200] projections on the Walnut-CBCT \cite{dersarkissianConebeamXrayComputed2019} dataset using the methods compared in Tab.~\ref{tab:walnut:quantitative}. \textit{First row} axial slices, \textit{second row} vertical slices from the same sample. PSNR is computed per slice.}
    \label{fig:walnut:imagettes}
 \end{figure*}

\section{Experiments} \label{sec:experiments}

We conduct experiments on two large-scale 3D inverse problem modalities: X-ray Cone-Beam Computed Tomography (CBCT) and Multi-Coil Magnetic Resonance Imaging (MC-MRI). First, we evaluate the performance of our \textbf{domain partitioning} approach, which allows us to train unrolled networks on small patches while deploying them on the whole problem at test-time. Second, we assess the benefits of our \textbf{normal operator approximation} technique, which enables efficient data-consistency updates during unrolled training.

\noindent \textbf{Baselines~~} As non-learned methods, we consider the Feldkamp-Davis-Kress (FDK) algorithm \cite{feldkampPracticalConebeamAlgorithm1984} for CBCT and the zero-filled root-sum-of-squares (RSS) reconstruction for MC-MRI. We also compare against Total Variation (TV) minimization \cite{getreuerRudinOsherFatemiTotalVariation2012} as a standard variational method. For each dataset, we also compare against several deep learning-based baselines, representative of the prominent families of methods for inverse problems and also scalable to 3D (see \cref{sec:related:work}): (i) post-processing networks \cite{hanDeepResidualLearning2016,lee_deep_2018}, which learn a one-pass mapping from a low-quality reconstruction to the ground-truth; (ii) PnP-$\alpha$PGD \cite{venkatakrishnanPlugPlayPriorsModel2013,hurault_relaxed_2023,voPlugPlayLearnedProximal2025} and DPIR \cite{zhangPlugPlayImageRestoration2021,terris_reconstruct_2025}, (iii) unrolled networks \cite{adlerLearnedPrimaldualReconstruction2018,dingLowDoseCTDeep2020,aggarwal_modl_2019,sriram_end--end_2020}, and (iv) implicit neural representations with instantNGP \cite{mullerInstantNeuralGraphics2022}. We use tied-weights unrolled networks, with $K = 5$ iterations on the MC-MRI dataset, and $K=3$ iterations on the CBCT dataset. Notably in the CBCT experiments, \cref{tab:walnut:quantitative}, the unrolled network with our strategy is trained with a batch of size $1$, while all other learned methods use a batch size of $4$, we accumulate gradients over $4$ steps to match the effective batch size.

For all learned experiments, we fix the backbone architecture to a DRUNet \cite{zhangPlugPlayImageRestoration2021}, a residual UNet-like convolutional network with 36.2M parameters, widely used in inverse problems. When applicable, we report results for both 2D and 3D versions of the methods to highlight the benefits of leveraging the 3D structure of the data, rather than processing it slice-by-slice. The 3D counterpart of the DRUNet has approximately 96.5M parameters. The different methods are implemented in \texttt{pytorch} \cite{paszkePyTorchImperativeStyle2019} and trained on a single H100 GPU with 80GB. We use the \texttt{astra-toolbox} \cite{vanaarleASTRAToolboxPlatform2015,vanaarleFastFlexibleXray2016} and \texttt{deepinverse} \cite{tachella_deepinverse_2025} libraries for CBCT and MC-MRI operators. Additional implementation details are provided in \cref{appendix:training:details}.
\subsection{Cone-Beam Computed Tomography}

\setlength{\tabcolsep}{2.5pt}
\begin{table}[h]
   \caption{Reconstruction performances on the \textbf{Walnut-CBCT} dataset. We do not report results for the standard unrolled approach as training it is infeasible on a single GPU and yields \textit{out of memory} errors (OOM). When applicable we report the peak GPU memory (VRAM) usage in gigabytes (GB) as well as the training speed in seconds per step (s/step). \colorbox[rgb]{0.7,1.0,0.7}{\textit{Best}} and \colorbox[rgb]{0.7,0.7,1.0}{\textit{second-best}} results highlighted.} \label{tab:walnut:quantitative}
   \centering{%
   \resizebox{1.\columnwidth}{!}{%
   \begin{tabular}{cc|ccc|ccc|c|c}
   \toprule
   \multicolumn{2}{c|}{\text{Walnut-CBCT}} & \multicolumn{3}{c|}{\textbf{SSIM} $\uparrow$} & \multicolumn{3}{c|}{\textbf{PSNR} $\uparrow$} & \textbf{VRAM $\downarrow$} & \textbf{train - s/step $\downarrow$ } \\
   \cmidrule{1-10} 

   \multicolumn{2}{l|}{Views $(\cdot / 1200)$} & 30 & 50 & 100 & 30 & 50 & 100 & \multicolumn{2}{c}{\text{30, 50, 100}} \\
   \cmidrule{1-10} 
   
   \multicolumn{2}{l|}{\text{FDK}} 
   & 0.197 & 0.263 & 0.375
   & 18.53 & 21.16 & 24.74 
   & N/A & N/A \\

   \multicolumn{2}{l|}{\text{TV}}
   & 0.799 & 0.850 & 0.893
   & 27.88 & 29.72 & 31.63 
   & N/A & N/A \\

   \cmidrule(lr){1-2} \cmidrule(lr){3-10} 

   \multicolumn{2}{l|}{\text{INR} [3D]} 
   & 0.805 & 0.862 & 0.913
   & 29.97 & 32.18 & 33.74
   & N/A & N/A \\

   \multicolumn{2}{l|}{\text{PnP-$\alpha$PGD} [2D]} 
   & 0.805 & 0.875 & 0.889
   & 28.74 & 32.15 & 33.97
   & 11.94 & 0.07 \\

   \multicolumn{2}{l|}{\text{PnP-$\alpha$PGD} [3D]} 
   & 0.803 & 0.868 & 0.884
   & 28.63 & 31.69 & 33.72
   & 67.50 & 1.39 \\

   \multicolumn{2}{l|}{\text{DPIR} [RAM][2D]} 
   & 0.774 & 0.815 & 0.826
   & 28.33 & 30.19 & 31.17 
   & N/A & N/A \\

   \cmidrule(lr){1-2} \cmidrule(lr){3-10} 

   \multicolumn{2}{l|}{\text{DRUNet}[2D]} 
   & 0.820 & 0.866 & 0.865
   & 28.30 & 31.14 & 33.66
   & 11.94 & 0.07 \\

   \multicolumn{2}{l|}{\text{DRUNet}[3D]} 
   & \colorbox[rgb]{0.7,0.7,1.0}{0.857} & 0.905 & 0.931 
   & \colorbox[rgb]{0.7,0.7,1.0}{29.47} & 32.49 & 35.22
   & 67.50 & 1.39 \\

   \cmidrule(lr){1-2} \cmidrule(lr){3-10} 

    \multicolumn{2}{l|}{Unrolled[3D]} 
   & \xmark & \xmark & \xmark
   & \xmark & \xmark & \xmark 
   & OOM & \xmark \\

   \cmidrule(lr){1-2} \cmidrule(lr){3-10} 
   
   \multicolumn{2}{l|}{Unrolled[2D] - \textbf{ours}}
   & 0.855 & \colorbox[rgb]{0.7,0.7,1.0}{0.911} & \colorbox[rgb]{0.7,0.7,1.0}{0.942}
   & 29.37 & \colorbox[rgb]{0.7,0.7,1.0}{32.52} & \colorbox[rgb]{0.7,0.7,1.0}{35.79}
   & 15.17 & 2.56 \\
   
 \multicolumn{2}{l|}{Unrolled[3D] - \textbf{ours}} 
   & \colorbox[rgb]{0.7,1.0,0.7}{0.877} & \colorbox[rgb]{0.7,1.0,0.7}{0.926} & \colorbox[rgb]{0.7,1.0,0.7}{0.947}
   & \colorbox[rgb]{0.7,1.0,0.7}{31.17} & \colorbox[rgb]{0.7,1.0,0.7}{34.21} & \colorbox[rgb]{0.7,1.0,0.7}{37.07}
   & 44.70 & 1.20 $\times 4$ \\

   \bottomrule%
   \end{tabular}%
    }%
   }
\end{table}

CBCT is a typical example of a 3D inverse problem that does not admit a coordinate-friendly partitioning of the forward operator, thus making it impossible to train unrolled networks on patches without our proposed domain partitioning strategy. 

We use real-world measurements from the \textbf{Walnut-CBCT} dataset \cite{dersarkissianConebeamXrayComputed2019}, which contains high-resolution CBCT scans of several walnuts. The detector has a size of $972 \times 768$ and the reconstructed volumes have a size of $501^3$ voxels. We simulate sparse-view CBCT by sub-sampling the projections to 30, 50, and 100 views out of the original 1200. During training, we use patches of size $384^2$ when using a 2D DRUNet and $8 \times 384^2$ with a 3D DRUNet. For each experiment, we train simultaneously on the three sub-sampling configurations.

In \cref{tab:walnut:quantitative}, we see that our domain partitioning strategy combined with our normal operator approximation technique allows us to train an unrolled 3D DRUNet, which achieves the best reconstruction performance among all compared methods. In \cref{tab:ablation:quantitative,} we also demonstrate the individual benefits of our normal operator approximation technique, which significantly reduces the training time while boosting the performance.

\subsection{Multi-Coil Magnetic Resonance Imaging}

\setlength{\tabcolsep}{2.5pt}
\begin{table}[h!]
   \caption{Reconstruction performances on the \textbf{Calgary-Campinas} dataset. PSNR and SSIM are measured on amplitude images. \colorbox[rgb]{0.7,1.0,0.7}{\textit{Best}} and \colorbox[rgb]{0.7,0.7,1.0}{\textit{second-best}} results highlighted.} \label{tab:calgary:quantitative}
   \centering{%
    \resizebox{0.9\columnwidth}{!}{%
   \begin{tabular}{cc|cc|cc|c|c}
   \toprule
   \multicolumn{2}{c|}{\text{Calgary-Campinas}} & \multicolumn{2}{c|}{\textbf{SSIM} $\uparrow$} & \multicolumn{2}{c|}{\textbf{PSNR} $\uparrow$} & \textbf{VRAM $\downarrow$} & \textbf{train - s/step $\downarrow$ } \\

   \cmidrule{1-8} 

   \multicolumn{2}{l|}{Acceleration rate} & R=5 & R=10 & R=5 & R=10 & \multicolumn{2}{c}{\text{R = 5 \& 10}} \\

   \cmidrule{1-8} 

   \multicolumn{2}{l|}{\text{RSS}} 
   & 0.473 & 0.334
   & 24.40 & 21.92
   & N/A & N/A \\

   \multicolumn{2}{l|}{\text{TV}}
   & 0.790 & 0.726 
   & 32.42 & 30.23
   & N/A & N/A \\

   \cmidrule(lr){1-2} \cmidrule(lr){3-8} 

   \multicolumn{2}{l|}{\text{INR} [3D]} 
   & 0.748 & 0.666 
   & 30.23 & 28.06
   & N/A & N/A \\

   \multicolumn{2}{l|}{\text{PnP-$\alpha$PGD} [2D] } 
   & 0.639 & 0.661 
   & 27.08 & 28.70
   & 10.04 & 0.080 \\

   \multicolumn{2}{l|}{\text{PnP-$\alpha$PGD} [3D]} 
   & 0.649 & 0.648
   & 27.27 & 28.47
   & 18.54 & 0.610\\

   \multicolumn{2}{l|}{\text{DPIR} [RAM][2D]} 
   & 0.612 & 0.582
   & 28.24 & 27.35 
   & N/A & N/A \\

   \cmidrule(lr){1-2} \cmidrule(lr){3-8} 

   \multicolumn{2}{l|}{\text{DRUNet}[2D]} 
   & 0.913 & 0.873
   & 33.87 & 31.31 
   & 10.04 & 0.080 \\

   \multicolumn{2}{l|}{\text{DRUNet}[3D]} 
   & 0.930 & 0.900
   & 35.02 & 32.67
   & 17.85 & 0.610\\

   \cmidrule(lr){1-2} \cmidrule(lr){3-8} 

   \multicolumn{2}{l|}{\text{Unrolled}[2D]} 
   & 0.941 & 0.905
   & 36.43 & 33.20
   & 15.04 & 0.190 \\

   \multicolumn{2}{l|}{\text{Unrolled}[3D]} 
   & \colorbox[rgb]{0.7,1.0,0.7}{0.952} & \colorbox[rgb]{0.7,1.0,0.7}{0.926}
   & \colorbox[rgb]{0.7,1.0,0.7}{37.74} & \colorbox[rgb]{0.7,1.0,0.7}{34.72}
   & 75.93 & 2.16 \\

   \cmidrule(lr){1-2} \cmidrule(lr){3-8} 

   \multicolumn{2}{l|}{Unrolled[2D] - \textbf{ours}} 
   & 0.942 & 0.906
   & 36.61 & 33.26
   & 11.66 & 0.590 \\

   \multicolumn{2}{l|}{Unrolled[3D] - \textbf{ours}} 
   & \colorbox[rgb]{0.7,0.7,1.0}{0.948} & \colorbox[rgb]{0.7,0.7,1.0}{0.919} 
   & \colorbox[rgb]{0.7,0.7,1.0}{37.36} & \colorbox[rgb]{0.7,0.7,1.0}{34.25}
   & 37.02 & 1.10 \\

   \bottomrule%
   \end{tabular}%
     }%
   }
\end{table}

\noindent For the MC-MRI modality, we use real-world measurements from the \textbf{Calgary-Campinas} dataset \cite{souza_open_2018}, which contains high-resolution multi-coil MRI scans of brains. The fully-sampled k-space data has a size of $256 \times 218 \times 170$ per coil, with 12 coils in total. We simulate accelerated MC-MRI by retrospectively undersampling the k-space with acceleration rates of $5$ and $10$ using a Poisson-disc sampling pattern \cite{souza_open_2018}. For each experiment, we train simultaneously on both acceleration configurations.

As opposed to the CBCT experiments, the underlying Fourier operator in the \textbf{Calgary-Campinas} dataset uses a Cartesian sampling pattern, which admits a coordinate-friendly partitioning along the depth dimension. This makes it possible to solve the 3D problem by solving a set of independent, smaller problems. Therefore, we also train an unrolled network on plain 2D slices or small 3D volumes truncated in the depth dimension. Using our domain partitioning strategy, we show that we can further reduce the memory footprint by partitioning the data in all three dimensions. We see in \cref{tab:calgary:quantitative} that our approach achieves similar reconstruction performance as the unrolled 3D network, while using significantly less memory during training. When using patch-training, we either use patches of size $128^2$ for 2D networks or $8 \times 128^2$ for 3D networks. We provide more details on the influence of the patch size in \cref{appendix:additional:results}.

\begin{figure*}[tb] \centering %\setkeys{Gin}{width=0.24\textwidth}

    \includegraphics[width=0.125\textwidth]{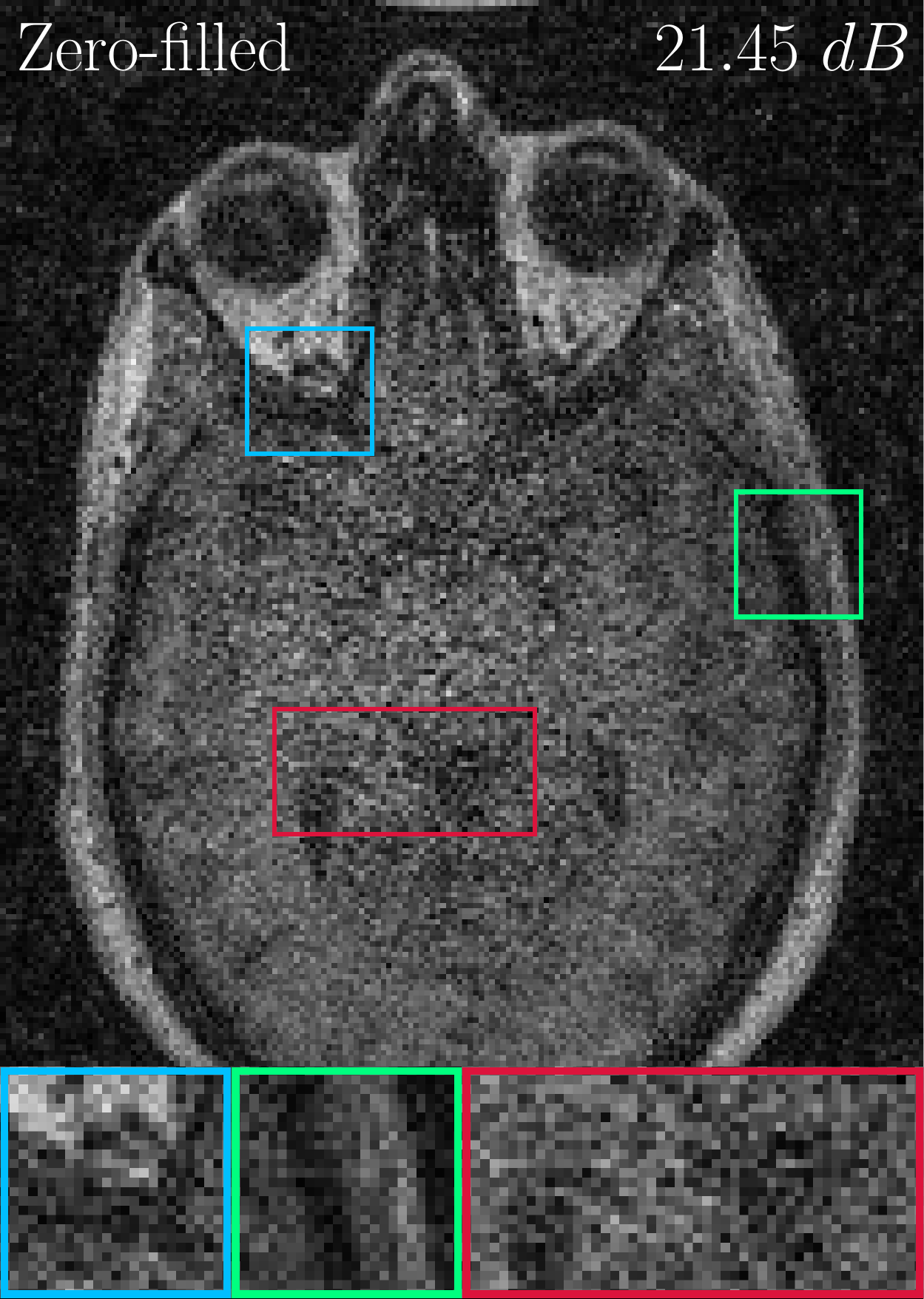}%
    \includegraphics[width=0.125\textwidth]{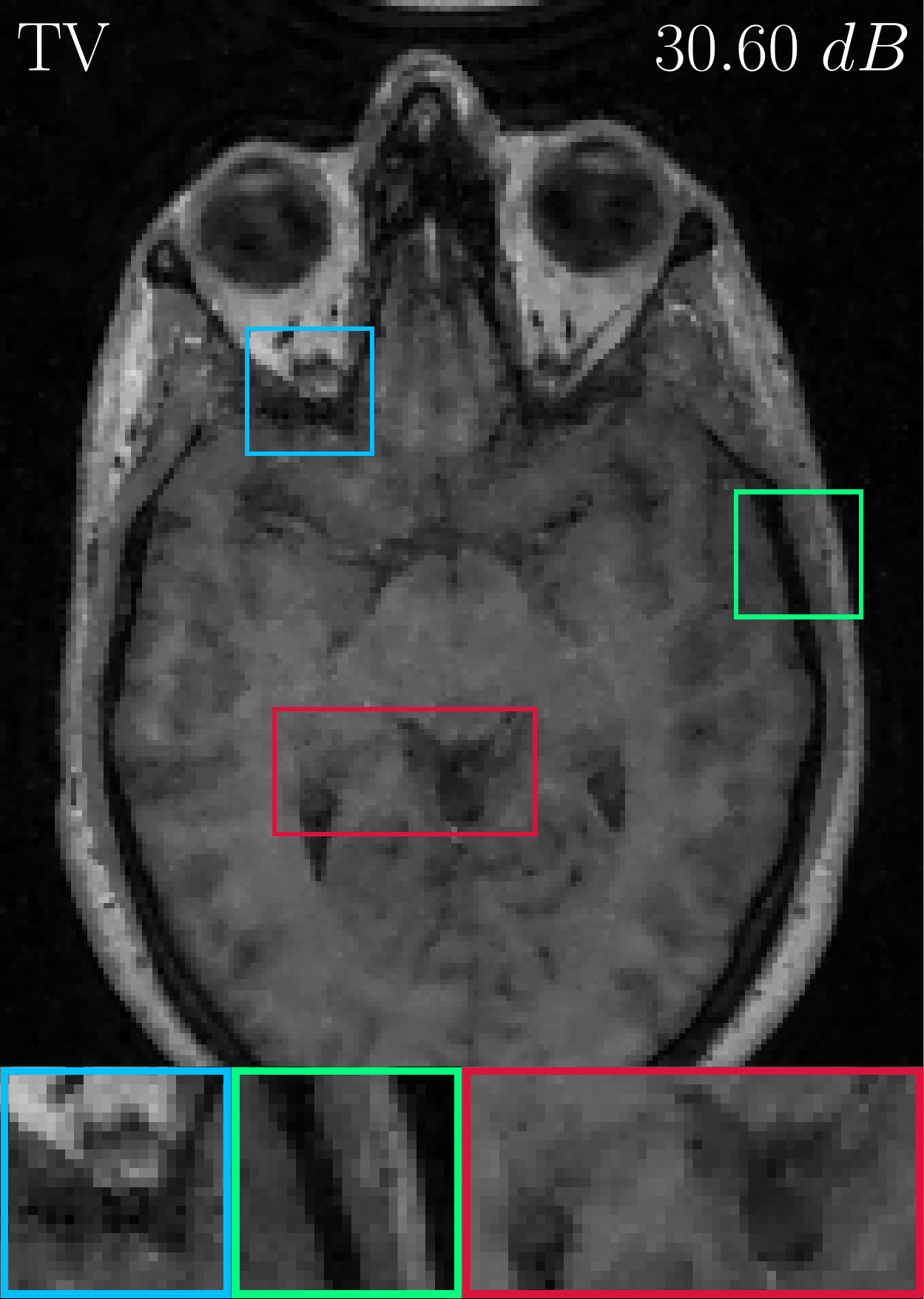}%
    \includegraphics[width=0.125\textwidth]{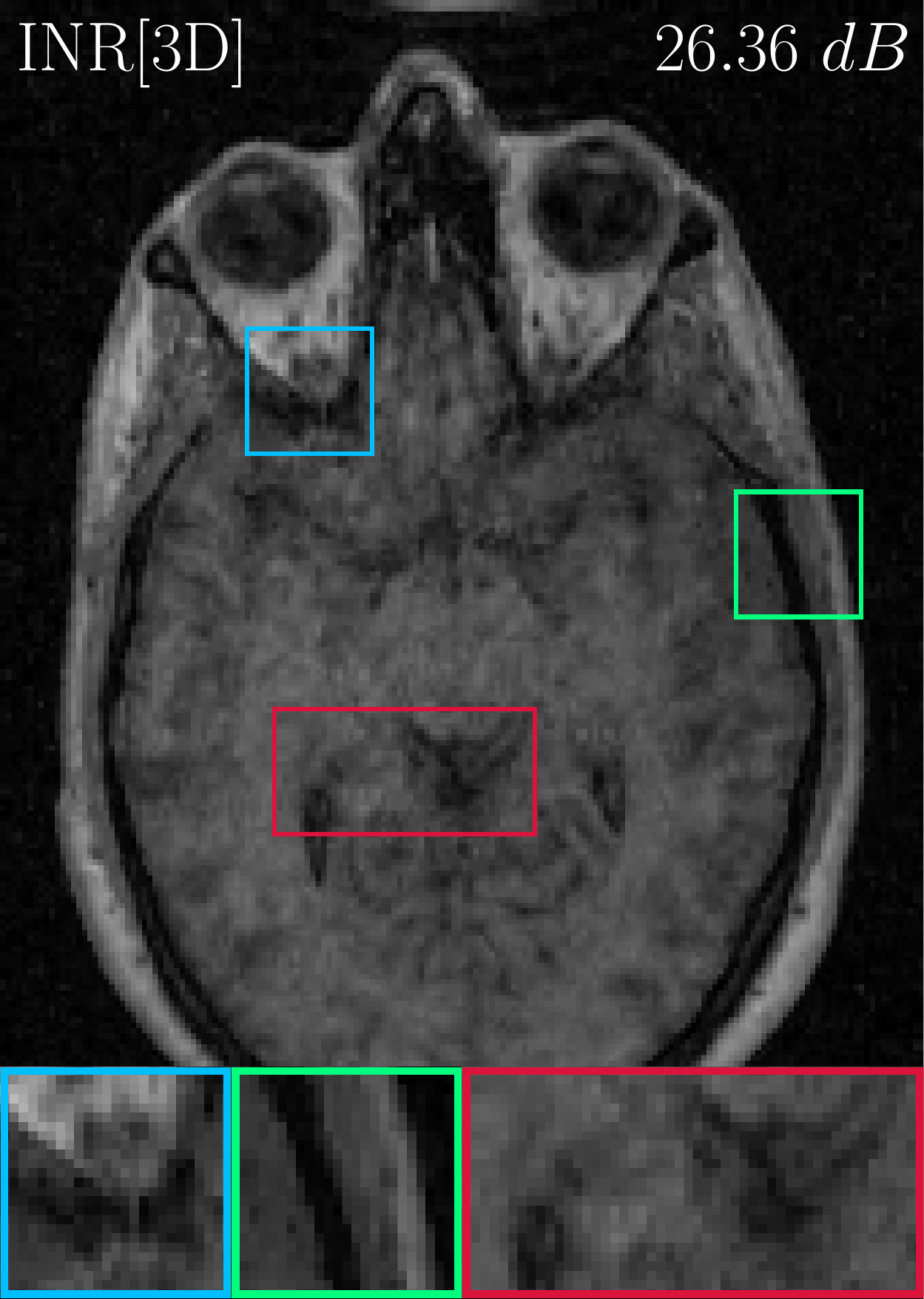}%
    \includegraphics[width=0.125\textwidth]{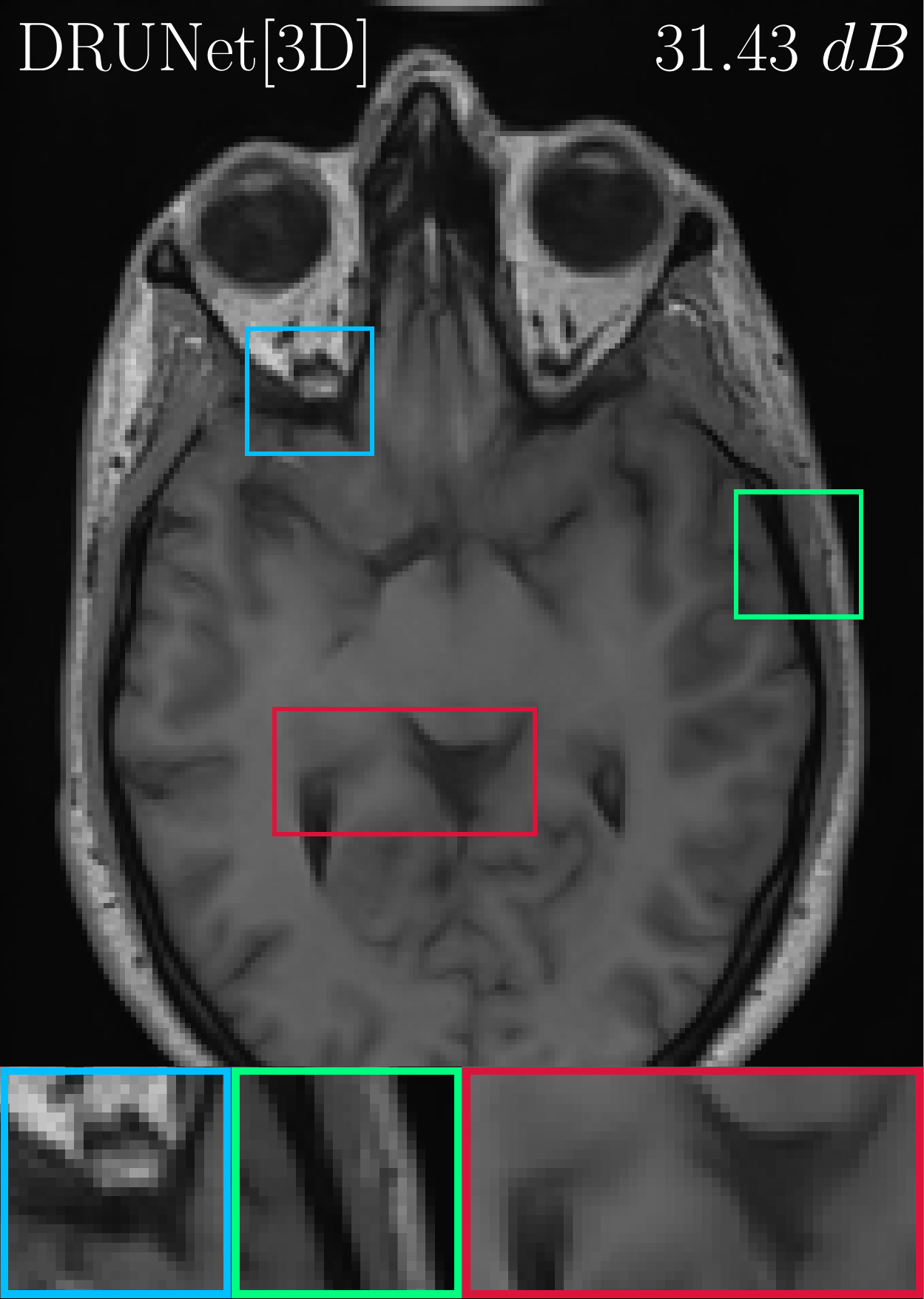}%
    \includegraphics[width=0.125\textwidth]{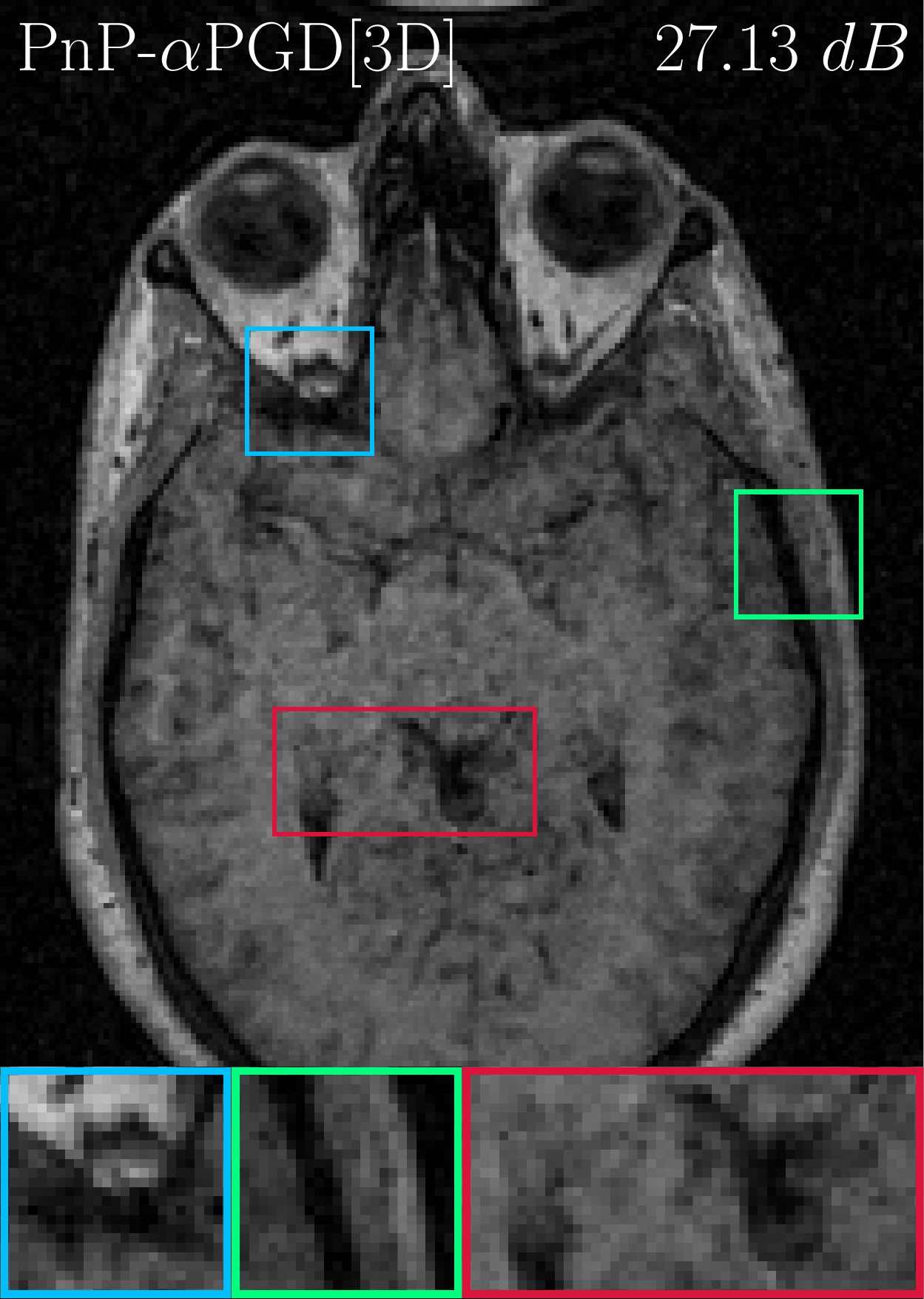}%
    \includegraphics[width=0.125\textwidth]{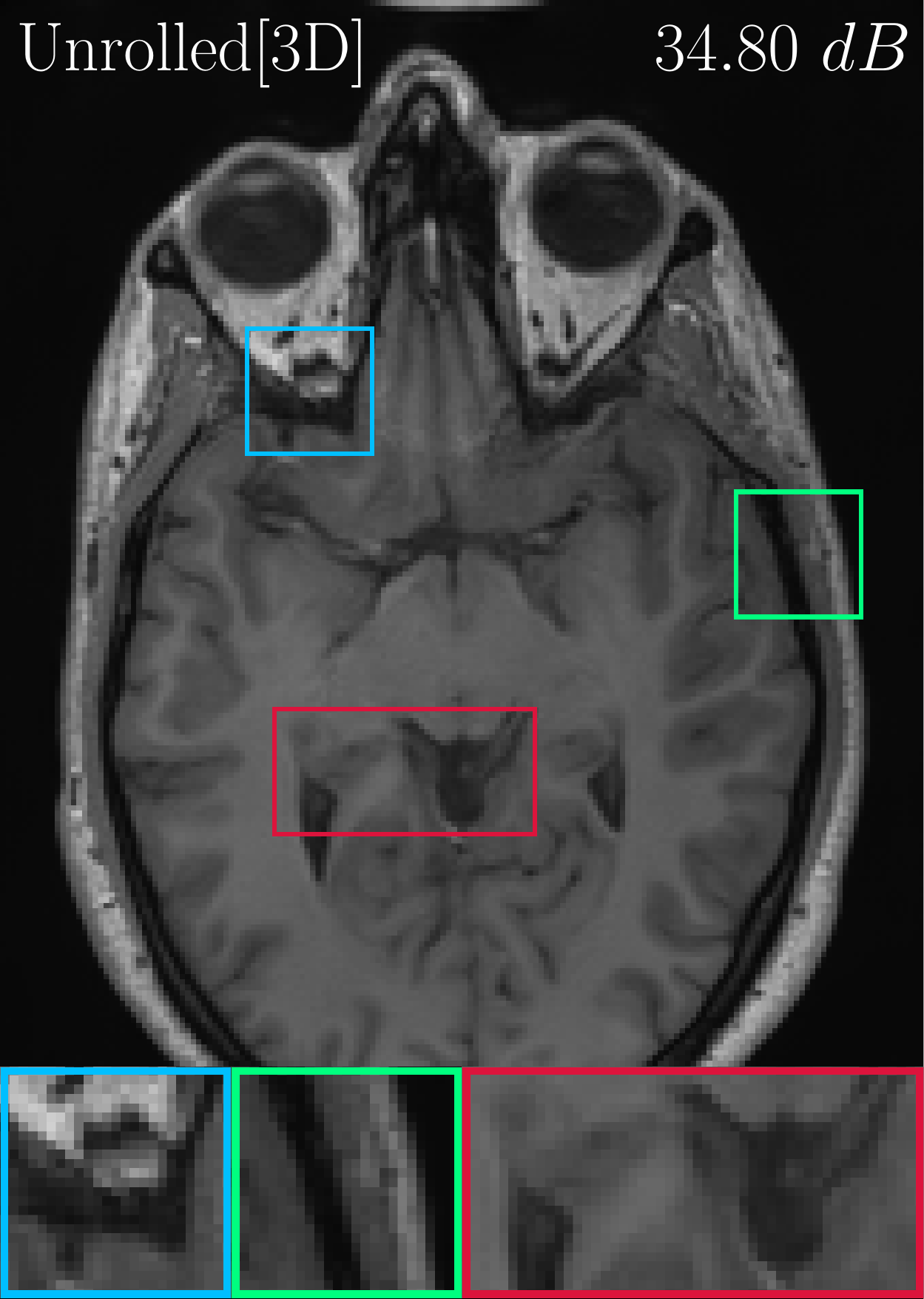}%
    \includegraphics[width=0.125\textwidth]{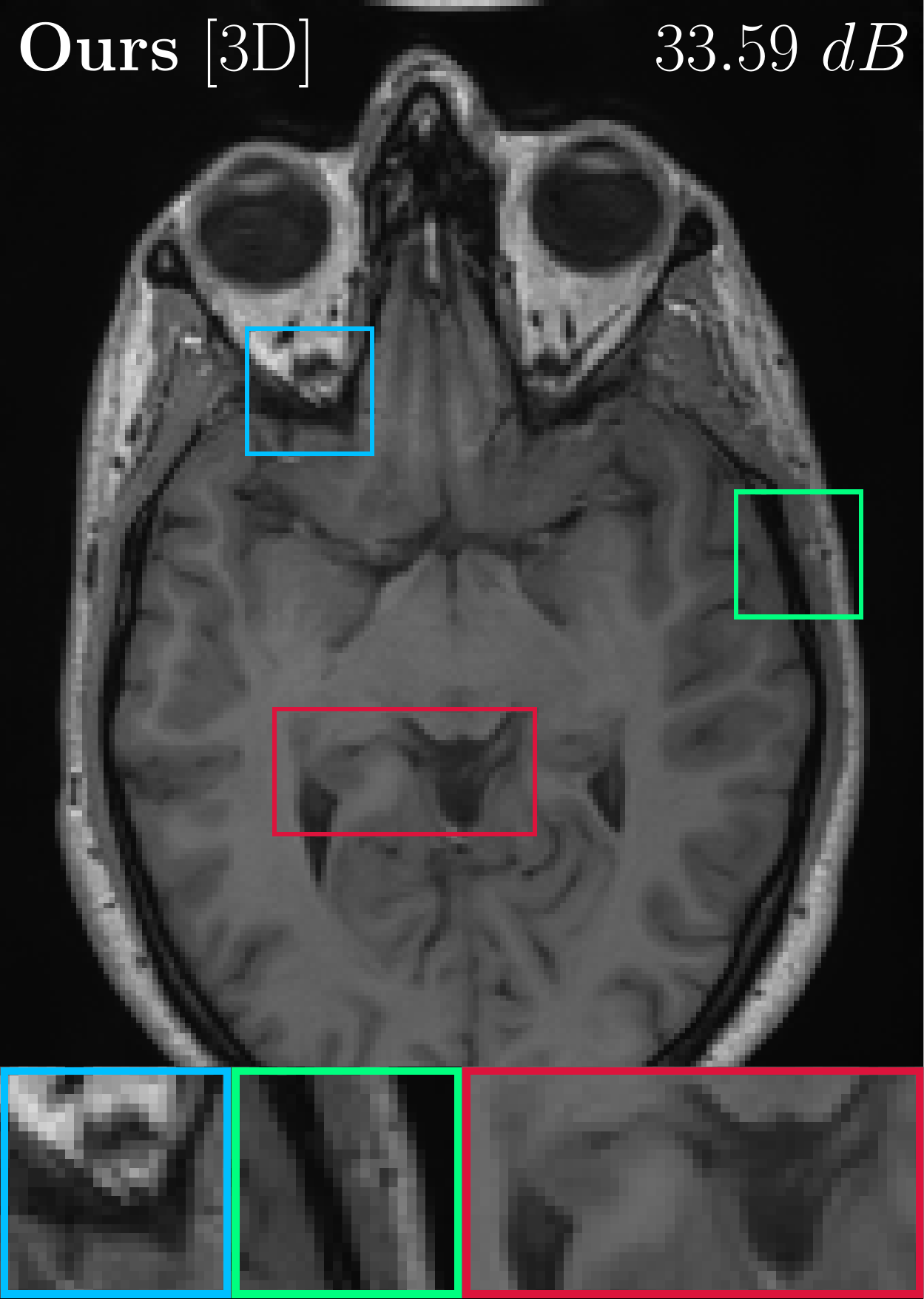}%
    \includegraphics[width=0.125\textwidth]{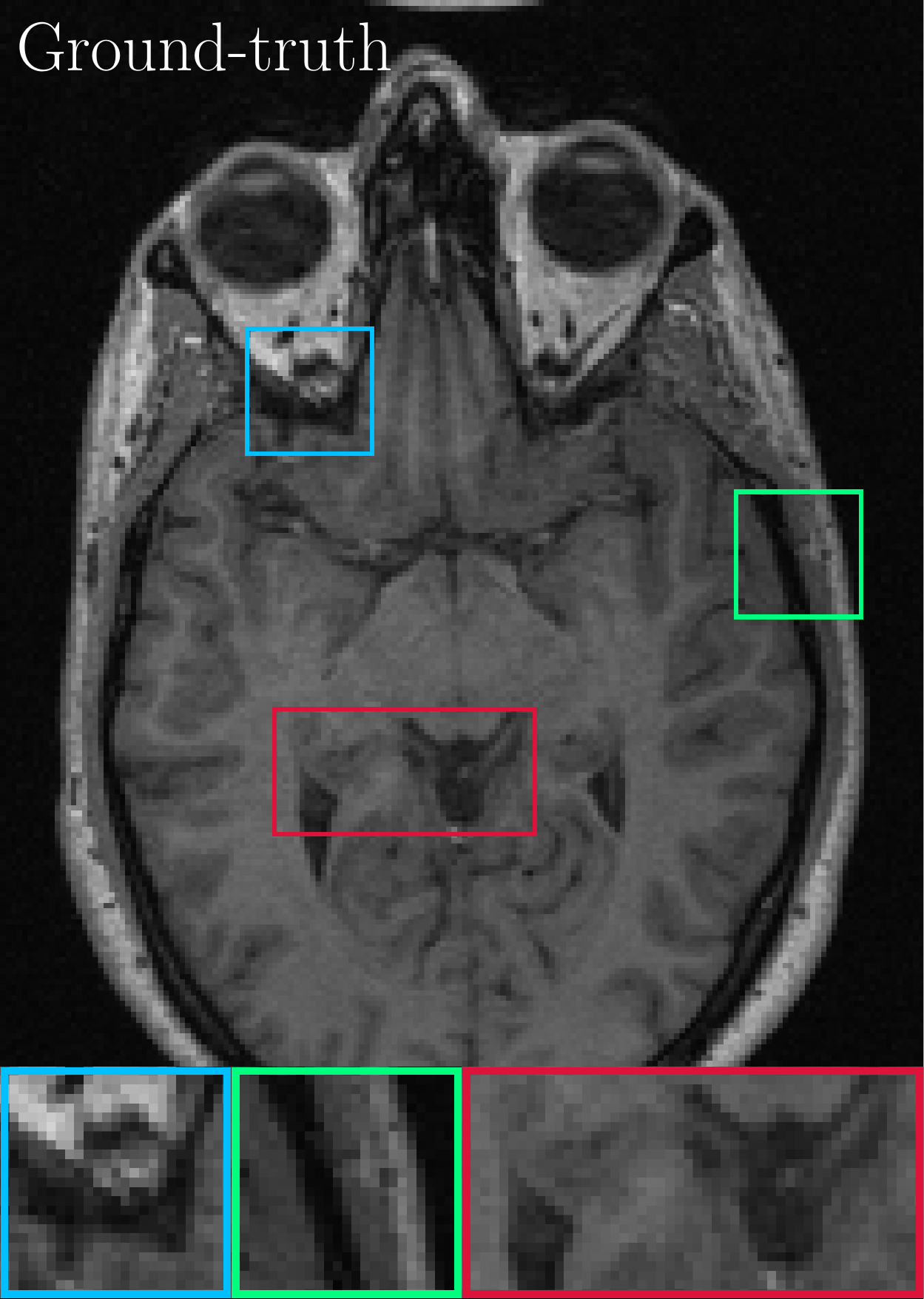}%

    \includegraphics[width=0.125\textwidth]{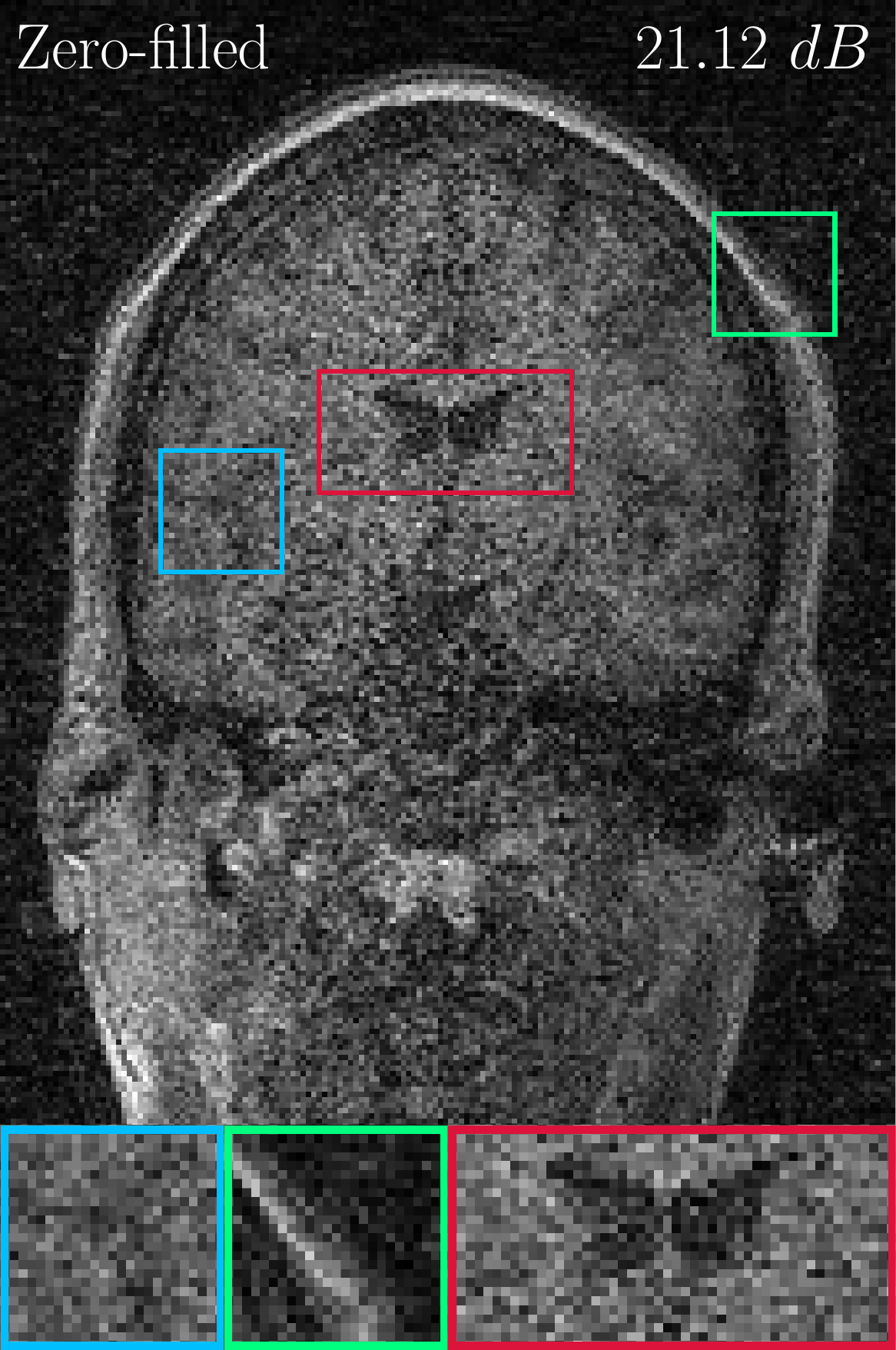}%
    \includegraphics[width=0.125\textwidth]{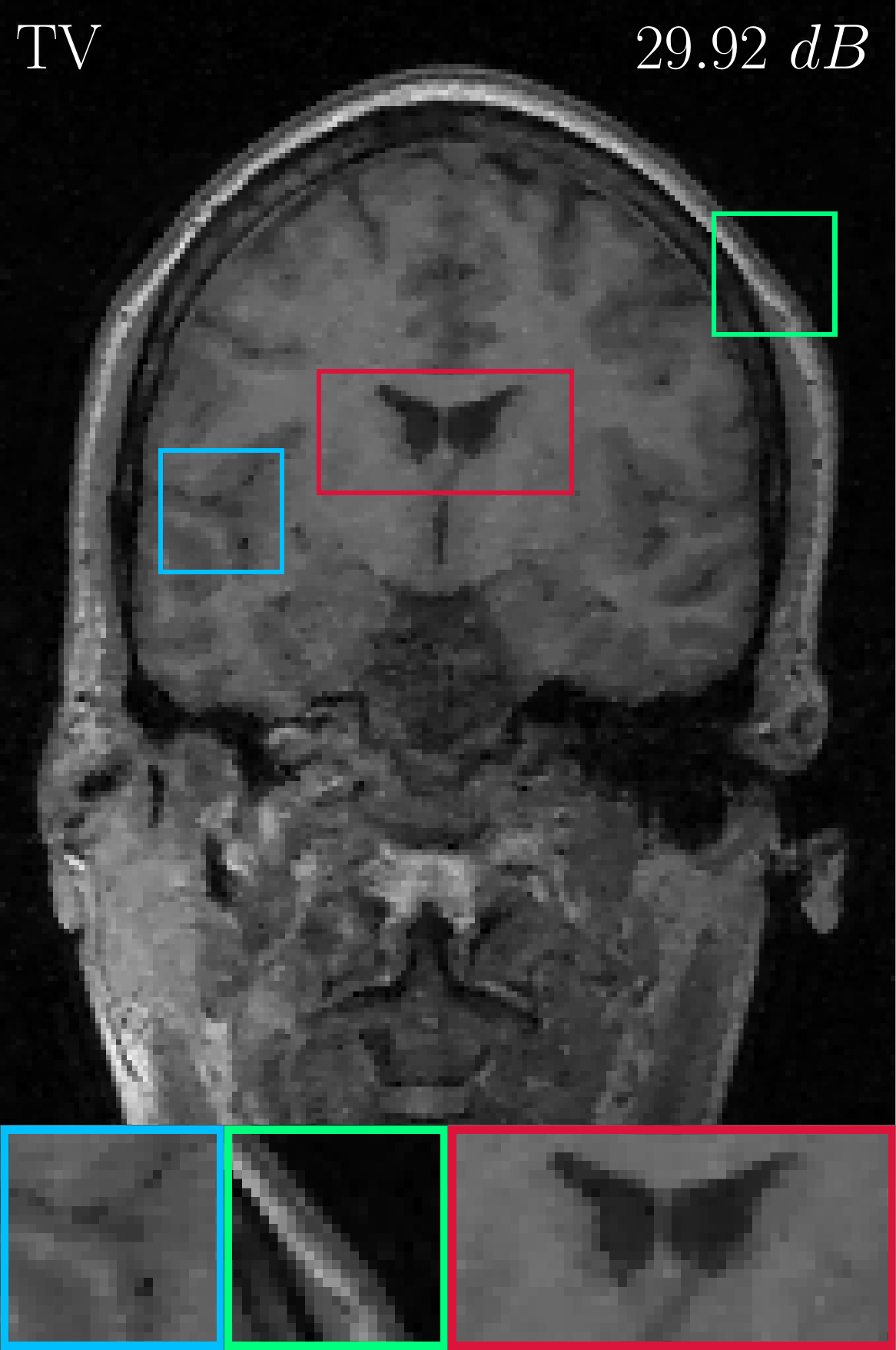}%
    \includegraphics[width=0.125\textwidth]{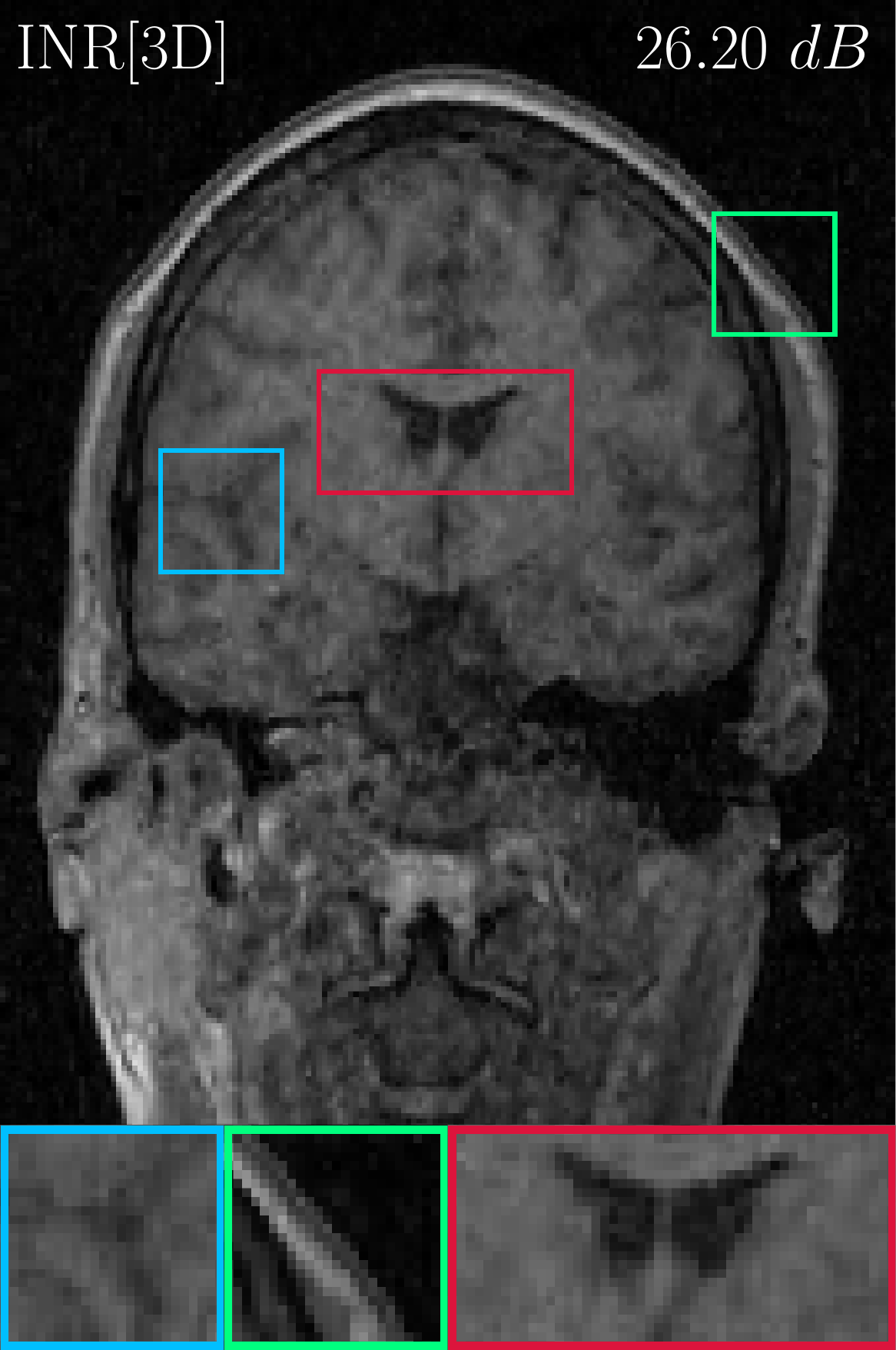}%
    \includegraphics[width=0.125\textwidth]{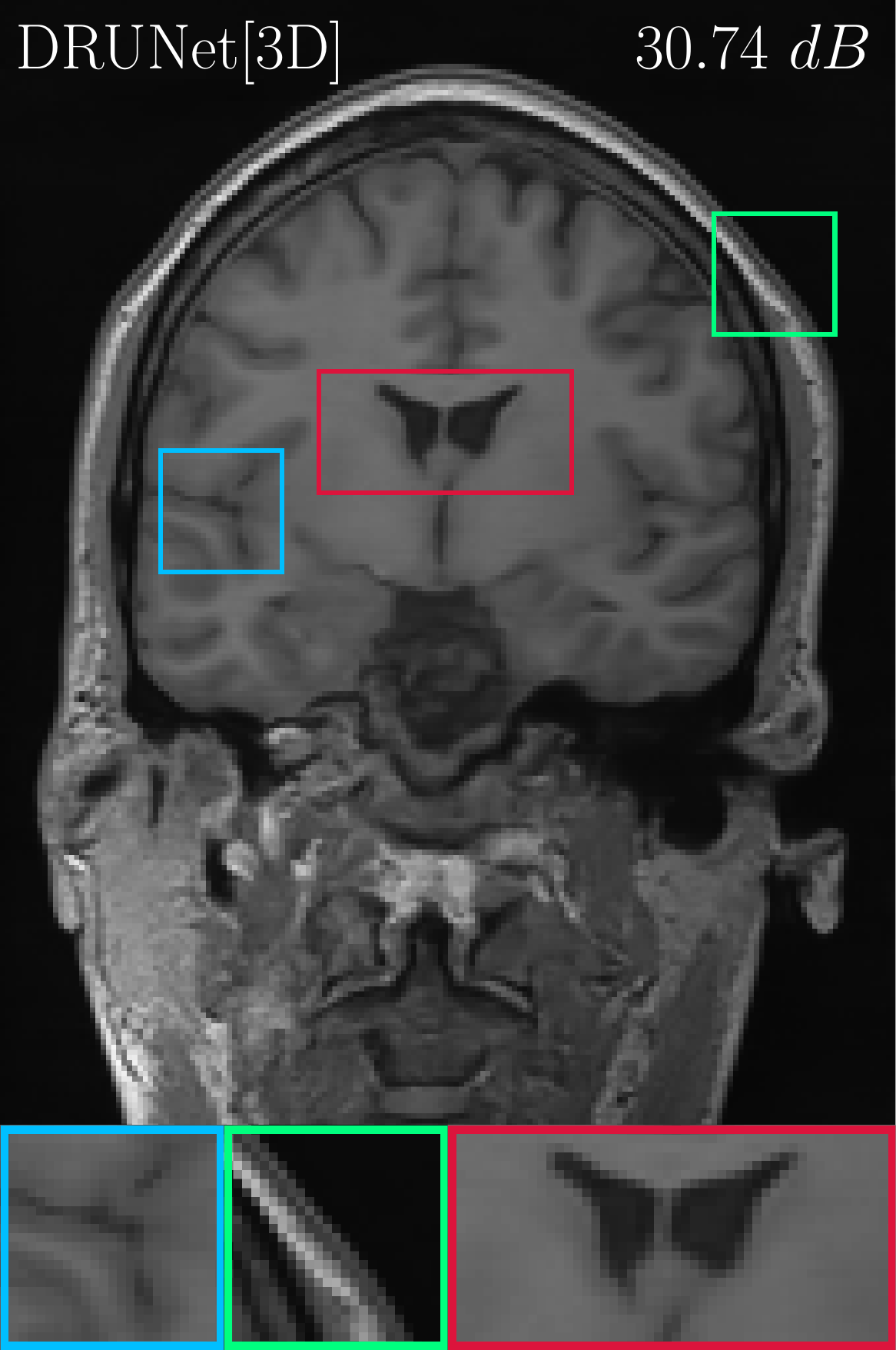}%
    \includegraphics[width=0.125\textwidth]{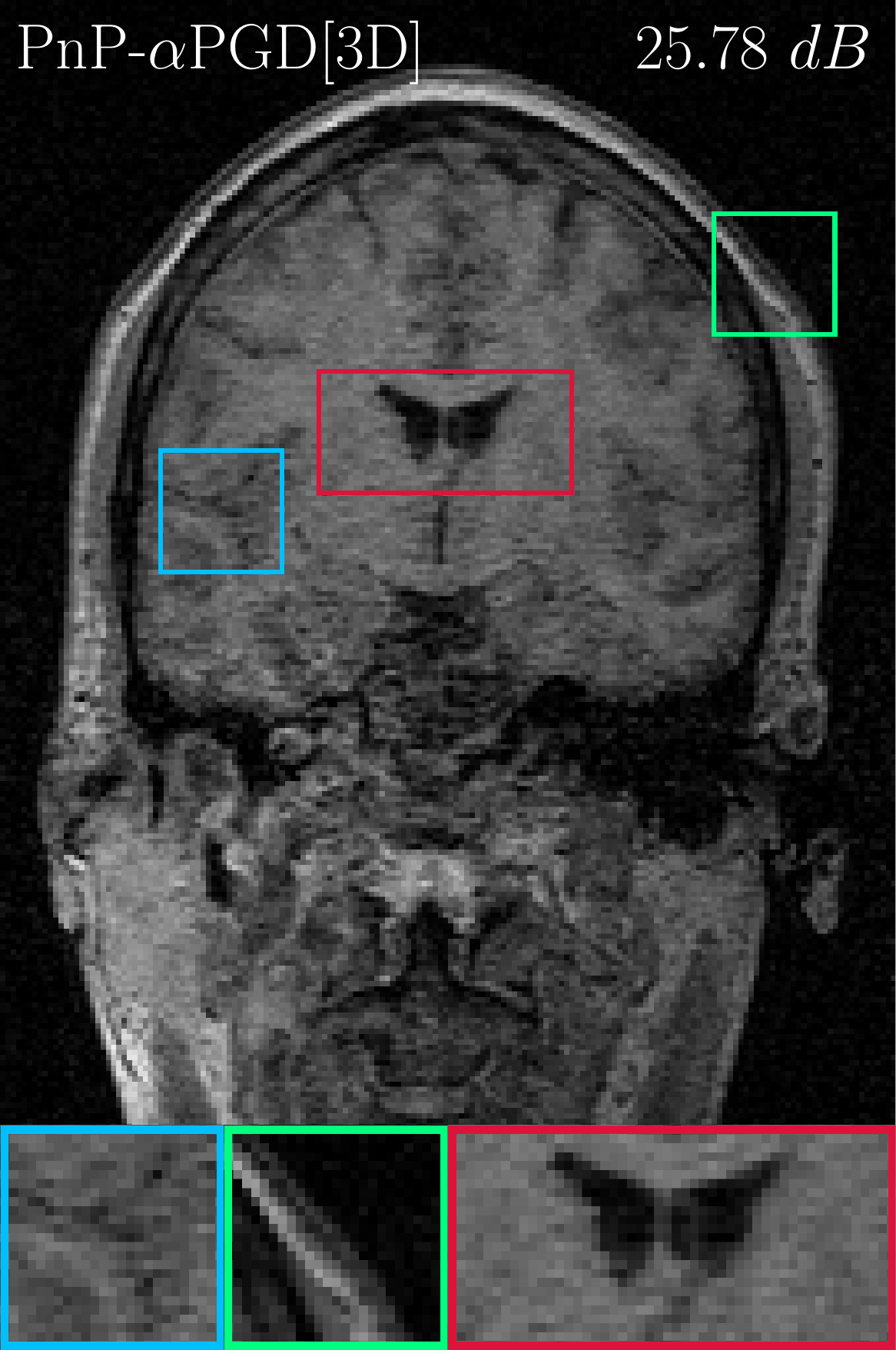}%
    \includegraphics[width=0.125\textwidth]{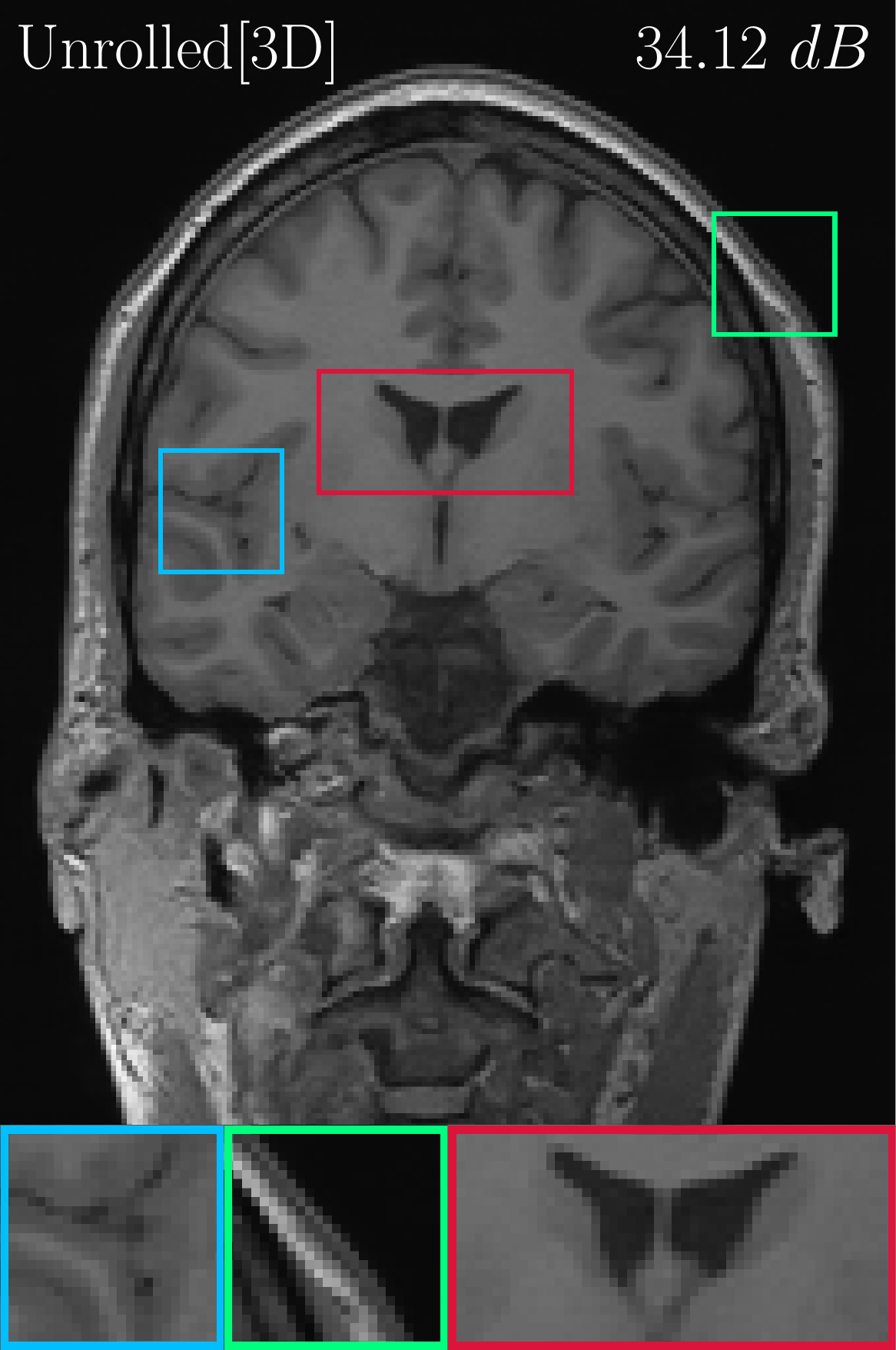}%
    \includegraphics[width=0.125\textwidth]{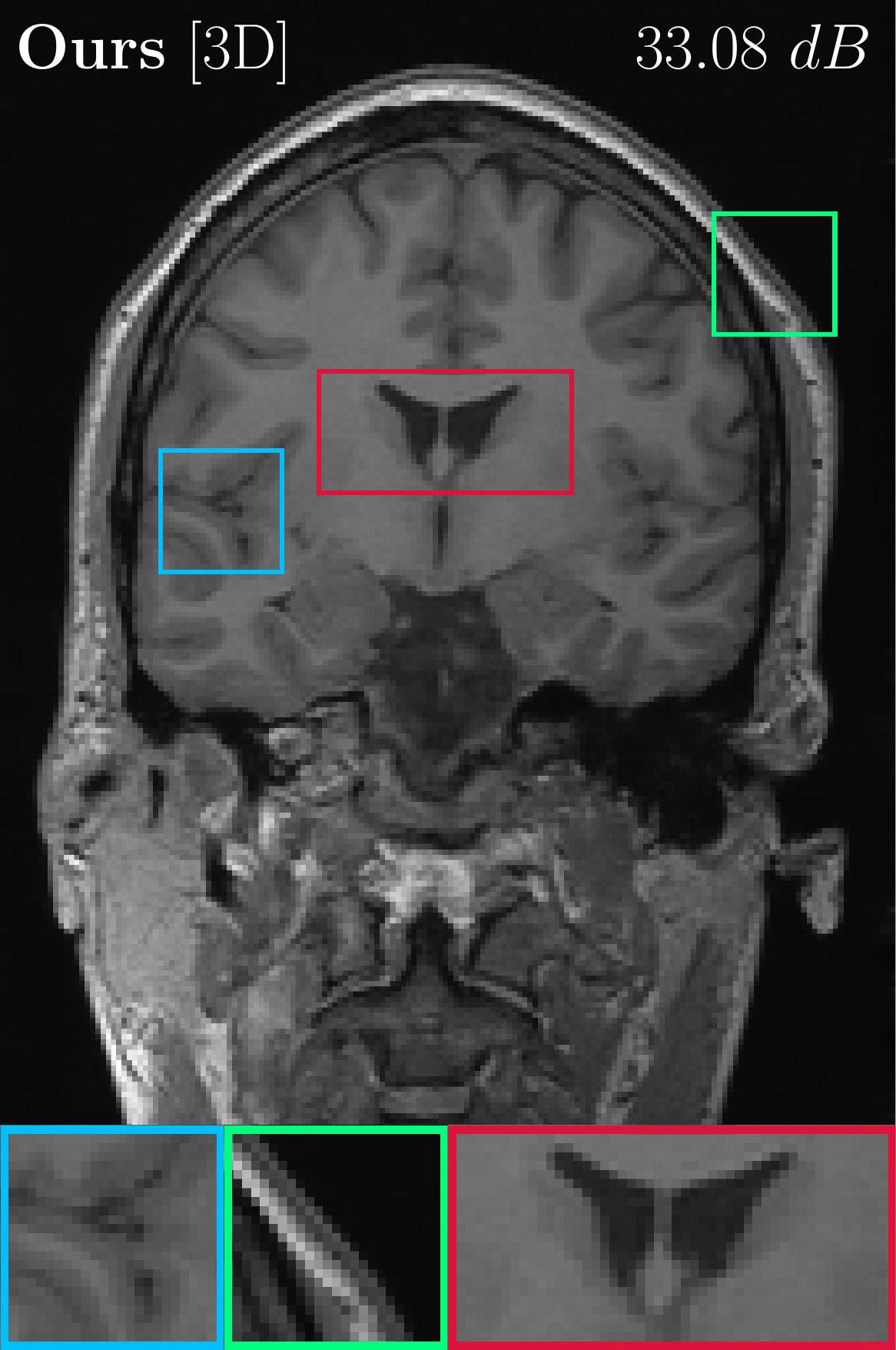}%
    \includegraphics[width=0.125\textwidth]{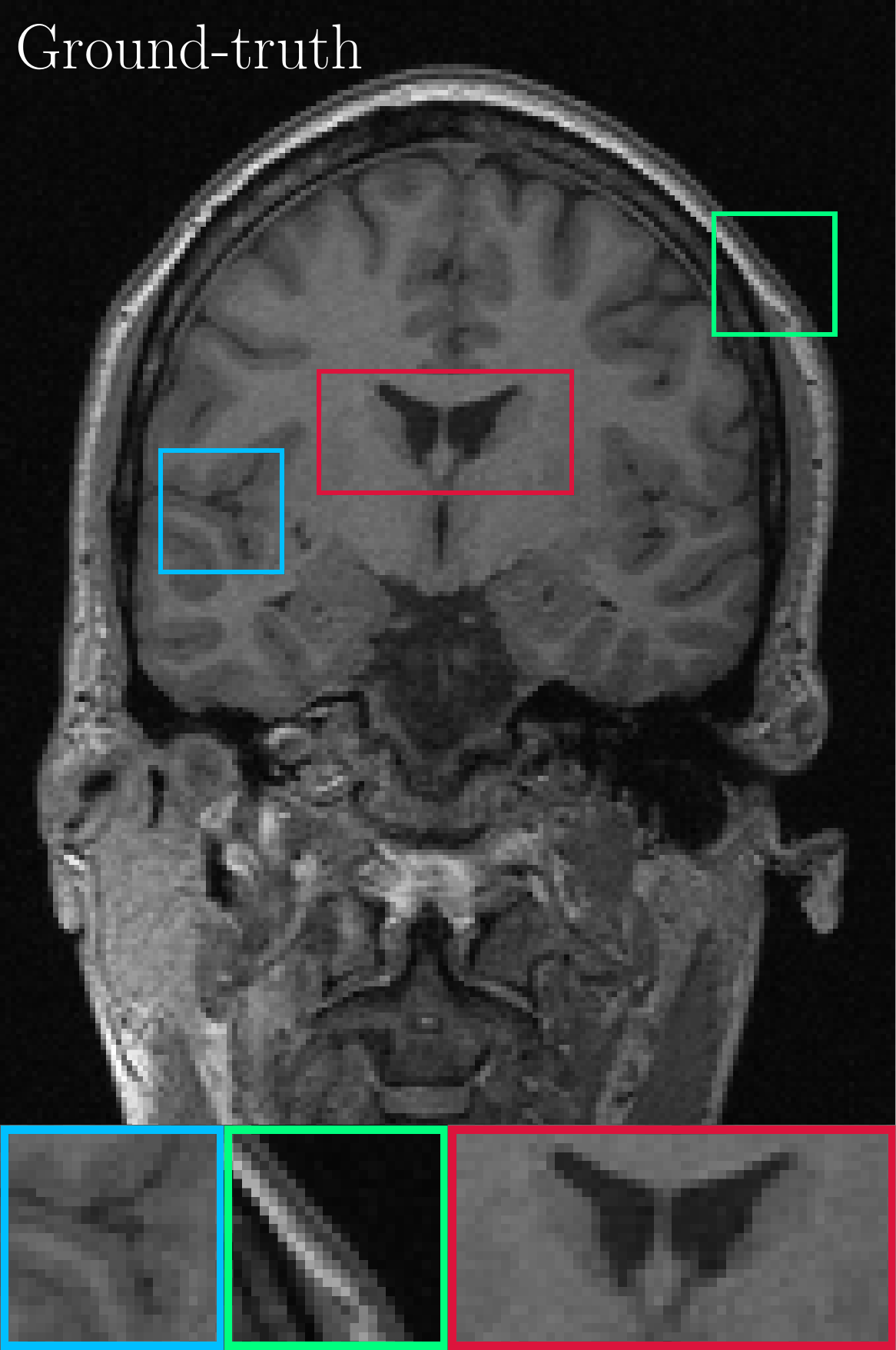}%

  \caption{Illustrations of MC-MRI reconstructions with acceleration rate of $5$ on the \textbf{Calgary-Campinas} dataset \cite{souza_open_2018} for the methods compared in \cref{tab:calgary:quantitative}. \textit{First row:} axial slice, \textit{second row:} coronal slice from the same sample. PSNR is computed per slice.}
  \label{fig:calgary:imagettes}
\end{figure*}

\subsection{Ablation study} \label{sec:ablation}

We conduct an ablation study (\cref{tab:ablation:quantitative}) to assess the individual contributions of our proposed domain partitioning strategy and normal operator approximation technique.

\noindent \textbf{Walnut-CBCT} \quad The size of the CBCT problem makes it impossible to evaluate the normal operator $\mA^\top \mA$ directly, thus we report no ablation for this setting. As reported in \cref{tab:ablation:quantitative}, the domain partitioning strategy enables the training of unrolled networks and yields state-of-the-art performance. When combined with the normal operator approximation, we observe that both the performance and the training speed are further improved, with a near $30\%$ reduction in training time.

\noindent \textbf{Calgary-Campinas MC-MRI} \quad In \cref{tab:ablation:quantitative}, we see that the domain partitioning strategy allows us to divide the memory footprint by a factor of $2$ while maintaining similar reconstruction performance (-0.38 dB in PSNR). Naturally, it also reduces the training time per step, as smaller patches require fewer computations.

As opposed to CBCT, the normal operator in MC-MRI already relies on efficient FFT computations. Indeed, we observe that the original data-step in MC-MRI only accounts for ~5\% of the forward step, thus limiting the potential speed-up. As explained in \cref{remark:mcmri:approximation}, we observe a drop in performance (\cref{tab:ablation:quantitative}) when using the normal operator approximation alone, although it still remains second best among compared methods. We expect the approximation to bring more significant benefits in Non-Cartesian MRI settings \cite{ramzi_ncpdnet_2022}, where the normal operator is much more expensive to compute.

\setlength{\tabcolsep}{2.5pt}
\begin{table}[h!]
   \caption{Ablation study on the \textbf{Calgary-Campinas} dataset and the \textbf{Walnut-CBCT} dataset. For each line we report the PSNR averaged on the different subsampling configurations, as well as the peak video memory usage in GB and training speed. \colorbox[rgb]{0.7,1.0,0.7}{\textit{Best}} and \colorbox[rgb]{0.7,0.7,1.0}{\textit{second-best}} results highlighted.} \label{tab:ablation:quantitative}
   \centering{%
   \resizebox{\columnwidth}{!}{%
   \begin{tabular}{cc|ccc|ccc|}
%    \cmidrule{1-8} 
   \multicolumn{2}{c|}{} & \multicolumn{3}{c|}{\textbf{Calgary MC-MRI}} & \multicolumn{3}{c|}{\textbf{Walnut-CBCT}}  \\

   \cmidrule(lr){1-8} 
   \multicolumn{2}{c|}{\textbf{Configuration}} & \textbf{PSNR} $\uparrow$ & \textbf{VRAM} $\downarrow$ & \textbf{s/step} $\downarrow$ & \textbf{PSNR} $\uparrow$ & \textbf{VRAM} $\downarrow$ & \textbf{s/step} $\downarrow$ \\

   \cmidrule(lr){1-2} \cmidrule(lr){3-8} 

   \multicolumn{2}{l|}{Unrolled~[3D]} 
   & \colorbox[rgb]{0.7,1.0,0.7}{36.23} & 75.93 & 2.16 
   & \xmark & OOM & \xmark \\

   \multicolumn{2}{l|}{\quad $\rightarrow$ w. $\mA^\top\mA$ \textit{approximation}}
   & 35.12 & 74.48 & 2.15
   & \xmark & OOM & \xmark \\

   \multicolumn{2}{l|}{\quad $\rightarrow$ w. \textit{partitioning}}
   & \colorbox[rgb]{0.7,0.7,1.0}{35.85} & 37.02 & 1.10
   & \colorbox[rgb]{0.7,0.7,1.0}{34.11} & 44.70 & 1.65 $\times 4$ \\

    % walnut 31.19, 34.07, 37.07
   \multicolumn{2}{l|}{\quad $\rightarrow$ w. \textit{part.} + $\mA^\top\mA$ \textit{approx.}}
   & 35.09 & 37.04 & 1.09 
   & \colorbox[rgb]{0.7,1.0,0.7}{34.15} & 44.70 & 1.21 $\times 4$ \\   

   \bottomrule%
   \end{tabular}%
     }%
   }
\end{table}

\section{Limitations} \label{sec:limitations}

In this work, we focus on inverse problems with Gaussian noise models, where the data-consistency step involves evaluating the normal operator $\mA^\top\mA$ to minimize a least-squares objective. While Poisson noise model is more appropriate for certain modalities such as low-dose CT, and does not rely on evaluating the normal operator, previous works have shown that unrolled networks designed for Gaussian noise models demonstrate state-of-the-art performance even under non-Gaussian noise conditions \cite{adlerLearnedPrimaldualReconstruction2018,gnanasambandam_secrets_2024,terris_reconstruct_2025}. Therefore, our proposed techniques remain relevant for a wide range of inverse problems beyond those strictly adhering to Gaussian noise assumptions.

\section{Conclusion} \label{sec:conclusion}

In this work, we address the challenge of scaling unrolled networks for large-scale inverse problems by introducing two key techniques: domain partitioning and normal operator approximation. The domain partitioning strategy decomposes large-scale inverse problems into smaller subproblems, yielding lower memory complexity at train-time. This approach extends patch-based training methods to unrolled architectures, ensuring scalability without compromising performance. Our proposed normal operator approximation completes our previous technique and replaces its evaluation with a product of diagonal and circulant matrices, enabling efficient data-consistency updates via the FFT. This reduces computational overhead, making it feasible to train and deploy unrolled networks for large-scale problems. We validate our approach through extensive experiments on 3D CBCT and 3D multi-coil accelerated MRI, demonstrating state-of-the-art performance while significantly reducing resource requirements. Notably, our method handles volumes as large as \(501^3\) on a single GPU, highlighting its potential for practical deployment in resource-constrained environments.

\section*{Acknowledgements}

R.~Vo and J.~Tachella acknowledge support by the French National Research Agency (Agence Nationale de la Recherche) grant UNLIP (ANR-23-CE23-0013). This project was provided with computing HPC and storage resources by GENCI at IDRIS thanks to the grant 2025-AD011014958R1 on the supercomputer Jean Zay's V100 and H100 partitions.

{
    \small
    \bibliographystyle{ieeenat_fullname}
    \bibliography{main}

@String(PAMI  = {IEEE Trans. Pattern Anal. Mach. Intell.})

@String(CVPR  = {IEEE Conf. Comput. Vis. Pattern Recog.})

@String(ECCV  = {European Conference on Computer Vision})

@String(NIPS = {Adv. Neural Inform. Process. Syst.})

@String(ICML  = {Int. Conf. Mach. Learn.})

@String(AAAI  = {AAAI})

@String(TOG   = {ACM Transactions on Graphics})

@String(TCI   = {IEEE Transactions on Computational Imaging})

@String(SPM	  = {IEEE Signal Processing Magazine})

@String(JOIS	= {SIAM Journal on Imaging Science})

@String(TMI   = {IEEE Transactions on Medical Imaging})

@String(MICCAI= {Medical Image Computing and Computer Assisted Intervention})

@article{adlerSolvingIllposedInverse2017,
    author = {Adler, Jonas and {\"O}ktem, Ozan},
    title = {Solving ill-posed inverse problems using iterative deep neural networks},
    year = {2017},
    journal = {Inverse Problems},
}

@article{adlerLearnedPrimaldualReconstruction2018,
    author = {Adler, Jonas and {\"O}ktem, Ozan},
    title = {Learned Primal-dual Reconstruction},
    year = {2018},
    journal = {IEEE Transactions on Medical Imaging},
}

@inproceedings{baiDeepEquilibriumModels2019,
  title = {Deep {{Equilibrium Models}}},
  booktitle = {Advances in {{Neural Information Processing Systems}}},
  author = {Bai, Shaojie and Kolter, J. Zico and Koltun, Vladlen},
  year= {2019}
}

@article{beckFastIterativeShrinkageThresholding2009,
  title = {A {{Fast Iterative Shrinkage-Thresholding Algorithm}} for {{Linear Inverse Problems}}},
  author = {Beck, Amir and Teboulle, Marc},
  year = {2009},
  journal = {SIAM Journal on Imaging Sciences},
}

@inproceedings{chungSolving3DInverse2023,
  title = {Solving {{3D Inverse Problems Using Pre-Trained 2D Diffusion Models}}},
  author = {Chung, Hyungjin and Ryu, Dohoon and McCann, Michael T. and Klasky, Marc L. and Ye, Jong Chul},
  year = {2023},
  booktitle = CVPR
}

@article{cohenRegularizationDenoisingFixedPoint2021,
  title = {Regularization by {{Denoising}} via {{Fixed-Point Projection}} ({{RED-PRO}})},
  author = {Cohen, Regev and Elad, Michael and Milanfar, Peyman},
  year = {2021},
  journal = {SIAM Journal on Imaging Sciences}
}

@inbook{combettesProximalSplittingMethods2010,
  author = {Combettes, Patrick L.
  and Pesquet, Jean-Christophe},
  title = {Proximal Splitting Methods in Signal Processing},
  booktitle = {Fixed-Point Algorithms for Inverse Problems in Science and Engineering},
  year= {2011},
  publisher = {Springer New York},
  pages = {185--212}
}

@book{fesslerAnalyticalTomographicImage2021,
  title = {Analytical {{Tomographic Image Reconstruction Methods}}},
  booktitle = {Image Reconstruction: {{Algorithms}} and Analysis},
  author = {Fessler, Jeffrey A.},
  year = {2021},
  url = {https://web.eecs.umich.edu/~fessler/book/c-tomo-prop.pdf}
}

@article{dersarkissianConebeamXrayComputed2019,
  title = {A Cone-Beam {{X-ray}} Computed Tomography Data Collection Designed for Machine Learning},
  author = {Der Sarkissian, Henri and Lucka, Felix and {van Eijnatten}, Maureen and Colacicco, Giulia and Coban, Sophia Bethany and Batenburg, Kees Joost},
  year = {2019},
  journal = {Scientific Data}
}

@article{dingLowDoseCTDeep2020,
  title = {Low-{{Dose CT}} with {{Deep Learning Regularization}} via {{Proximal Forward Backward Splitting}}},
  author = {Ding, Qiaoqiao and Chen, Gaoyu and Zhang, Xiaoqun and Huang, Qiu and Gao, Hui Jiand Hao},
  year = {2020},
  journal = {Physics in Medicine \& Biology}
}

@inproceedings{dosovitskiyImageWorth16x162021,
  title={An Image is Worth 16x16 Words: Transformers for Image Recognition at Scale},
  author={Alexey Dosovitskiy and Lucas Beyer and Alexander Kolesnikov and Dirk Weissenborn and Xiaohua Zhai and Thomas Unterthiner and Mostafa Dehghani and Matthias Minderer and Georg Heigold and Sylvain Gelly and Jakob Uszkoreit and Neil Houlsby},
  booktitle={International Conference on Learning Representations},
  year={2021},
}

@inproceedings{dupontDataFunctaYour2022,
  title = {From Data to Functa: {{Your}} Data Point Is a Function and You Can Treat It like One},
  author = {Dupont, Emilien and Kim, Hyunjik and Eslami, S. M. Ali and Rezende, Danilo Jimenez and Rosenbaum, Dan},
  booktitle = ICML,
  year = {2023}
}

@article{feldkampPracticalConebeamAlgorithm1984,
  title = {Practical Cone-Beam Algorithm},
  author = {Feldkamp, L. A. and Davis, L. C. and Kress, J. W.},
  year = {1984},
  journal = {Journal of the Optical Society of America}
}

@inproceedings{fungJFBJacobianFreeBackpropagation2021,
  title = {{{JFB}}: {{Jacobian-Free Backpropagation}} for {{Implicit Networks}}},
  author = {Fung, Samy Wu and Heaton, Howard and Li, Qiuwei and McKenzie, Daniel and Osher, Stanley and Yin, Wotao},
  booktitle = {AAAI Conference on Artificial Intelligence},
  year = {2022}
}

@article{getreuerRudinOsherFatemiTotalVariation2012,
  title = {Rudin-{{Osher-Fatemi Total Variation Denoising}} Using {{Split Bregman}}},
  author = {Getreuer, Pascal},
  year = {2012},
  journal = {Image Processing On Line},
}

@article{giltonDeepEquilibriumArchitectures2021,
  title = {Deep {{Equilibrium Architectures}} for {{Inverse Problems}} in {{Imaging}}},
  author = {Gilton, Davis and Ongie, Gregory and Willett, Rebecca},
  year = {2021},
  journal = {IEEE Transactions on Computational Imaging},
}

@unpublished{hanDeepResidualLearning2016,
  title = {Deep {{Residual Learning}} for {{Compressed Sensing CT Reconstruction}} via {{Persistent Homology Analysis}}},
  author = {Han, Yo Seob and Yoo, Jaejun and Ye, Jong Chul},
  year = {2016},
  journal = {arXiv preprint arXiv:1611.06391},
}

@inproceedings{hoDenoisingDiffusionProbabilistic2020,
  title = {Denoising {{Diffusion Probabilistic Models}}},
  booktitle = {Advances in {{Neural Information Processing Systems}}},
  author = {Ho, Jonathan and Jain, Ajay and Abbeel, Pieter},
  year = {2020},
}

@unpublished{kingmaAdamMethodStochastic2017,
  title = {Adam: {{A Method}} for {{Stochastic Optimization}}},
  author = {Kingma, Diederik P. and Ba, Jimmy},
  year = {2014},
  journal = {arXiv preprint arXiv:1412.6980},
}

@article{leuschnerQuantitativeComparisonDeep2021,
  title = {Quantitative {{Comparison}} of {{Deep Learning-Based Image Reconstruction Methods}} for {{Low-Dose}} and {{Sparse-Angle CT Applications}}},
  author = {Leuschner, Johannes and Schmidt, Maximilian and Ganguly, Poulami Somanya and Andriiashen, Vladyslav and Coban, Sophia Bethany and Denker, Alexander and Bauer, Dominik and Hadjifaradji, Amir and Batenburg, Kees Joost and Maass, Peter and {van Eijnatten}, Maureen},
  year = {2021},
  journal = {Journal of Imaging},
}

@article{leuschnerLoDoPaBCTDatasetBenchmark2021,
  title = {The {{LoDoPaB-CT Dataset}}: {{A Benchmark Dataset}} for {{Low-Dose CT Reconstruction Methods}}},
  author = {Leuschner, Johannes and Schmidt, Maximilian and Baguer, Daniel Otero and Maa{\ss}, Peter},
  year = {2021},
  journal = {Scientific Data}
}

@unpublished{liNerfAccGeneralNeRF2022,
  title = {{{NerfAcc}}: {{A General NeRF Acceleration Toolbox}}},
  author = {Li, Ruilong and Tancik, Matthew and Kanazawa, Angjoo},
  year = {2022},
  journal = {arXiv preprint arXiv:2210.04847},
}

@inproceedings{liuOnlineDeepEquilibrium2022,
  title = {Online {{Deep Equilibrium Learning}} for {{Regularization}} by {{Denoising}}},
  booktitle = {Advances in Neural Information Processing Systems},
  author = {Liu, Jiaming and Xu, Xiaojian and Gan, Weijie and Shoushtari, Shirin and Kamilov, Ulugbek},
  year = {2022},
}

@inproceedings{lunzAdversarialRegularizersInverse2018,
  title = {Adversarial {{Regularizers}} in {{Inverse Problems}}},
  booktitle = {Advances in {{Neural Information Processing Systems}}},
  author = {Lunz, Sebastian and Öktem, Ozan and Schönlieb, Carola-Bibiane},
  year = {2018},
}

@misc{mallatWaveletTourSignal1999,
  title = {A Wavelet Tour of Signal Processing},
  author = {Mallat, Stephane},
  date = {1999},
  url = {https://www.di.ens.fr/~mallat/papiers/WaveletTourChap1-6.pdf},
  organization = {Academic press},
}

@inproceedings{mildenhallNeRFRepresentingScenes2020,
  title = {{{NeRF}}: {{Representing Scenes}} as {{Neural Radiance Fields}} for {{View Synthesis}}},
  booktitle = {European Conference on Computer Vision},
  author = {Mildenhall, Ben and Srinivasan, Pratul P. and Tancik, Matthew and Barron, Jonathan T. and Ramamoorthi, Ravi and Ng, Ren},
  year = {2020}
}

@article{mukherjeeLearnedReconstructionMethods2023,
  title = {Learned {{Reconstruction Methods With Convergence Guarantees}}: {{A}} Survey of Concepts and Applications},
  author = {Mukherjee, Subhadip and Hauptmann, Andreas and Öktem, Ozan and Pereyra, Marcelo and Schönlieb, Carola-Bibiane},
  year = {2023},
  journal = SPM
}

@article{mullerInstantNeuralGraphics2022,
  title = {Instant Neural Graphics Primitives with a Multiresolution Hash Encoding},
  author = {Müller, Thomas and Evans, Alex and Schied, Christoph and Keller, Alexander},
  year = {2022},
  journal = TOG
}

@unpublished{nguyenComparingPlugandPlayUnrolled2024,
  title = {{Comparing Plug-and-Play and Unrolled networks}},
  author = {Nguyen, Minh Hai and Weiss, Pierre},
  url = {https://hal.science/hal-04703008},
  year = {2024},
  hal_id = {hal-04703008}
}

@inproceedings{paszkePyTorchImperativeStyle2019,
  author = {Paszke, Adam and Gross, Sam and Massa, Francisco and Lerer, Adam and Bradbury, James and Chanan, Gregory and Killeen, Trevor and Lin, Zeming and Gimelshein, Natalia and Antiga, Luca and Desmaison, Alban and Kopf, Andreas and Yang, Edward and DeVito, Zachary and Raison, Martin and Tejani, Alykhan and Chilamkurthy, Sasank and Steiner, Benoit and Fang, Lu and Bai, Junjie and Chintala, Soumith},
  title       = {PyTorch: An Imperative Style, High-Performance Deep Learning Library},
  booktitle           = NIPS,
  year            = {2019}
}

@article{pesquetLearningMaximallyMonotone2021,
  title = {Learning {{Maximally Monotone Operators}} for {{Image Recovery}}},
  author = {Pesquet, Jean-Christophe and Repetti, Audrey and Terris, Matthieu and Wiaux, Yves},
  year = {2021},
  journal = JOIS
}

@inproceedings{ronnebergerUNetConvolutionalNetworks2015,
  title = {U-{{Net}}: {{Convolutional Networks}} for {{Biomedical Image Segmentation}}},
  booktitle = {Medical {{Image Computing}} and {{Computer-Assisted Intervention}}},
  author = {Ronneberger, Olaf and Fischer, Philipp and Brox, Thomas},
  year = {2015}
}

@inproceedings{rudzusikaInvertibleLearnedPrimalDual2021,
    title={Invertible Learned Primal-Dual},
    author={Jevgenija Rudzusika and Buda Bajic and Ozan {\"O}ktem and Carola-Bibiane Sch{\"o}nlieb and Christian Etmann},
    booktitle={NeurIPS 2021 Workshop on Deep Learning and Inverse Problems},
    year={2021}
}

@inproceedings{sanderMomentumResidualNeural2021,
  title = {Momentum {{Residual Neural Networks}}},
  booktitle = {{{International Conference}} on {{Machine Learning}}},
  author = {Sander, Michael E. and Ablin, Pierre and Blondel, Mathieu and Peyré, Gabriel},
  year = {2021}
}

@article{shenNeRPImplicitNeural2022,
  title = {{{NeRP}}: {{Implicit Neural Representation Learning With Prior Embedding}} for {{Sparsely Sampled Image Reconstruction}}},
  author = {Shen, Liyue and Pauly, John and Xing, Lei},
  year = {2022},
  journal = {IEEE Transactions on Neural Networks and Learning Systems},
}

@inproceedings{songSolvingInverseProblems2022,
  title={Solving Inverse Problems in Medical Imaging with Score-Based Generative Models},
  author={Yang Song and Liyue Shen and Lei Xing and Stefano Ermon},
  booktitle={International Conference on Learning Representations},
  year={2022},
}

@inproceedings{sunBlockCoordinateRegularization2019,
  title = {Block {{Coordinate Regularization}} by {{Denoising}}},
  booktitle = {Advances in {{Neural Information Processing Systems}}},
  author = {Sun, Yu and Liu, Jiaming and Kamilov, Ulugbek},
  year = {2019}
}

@article{sunOnlinePlugandPlayAlgorithm2019,
  title = {An {{Online Plug-and-Play Algorithm}} for {{Regularized Image Reconstruction}}},
  author = {Sun, Yu and Wohlberg, Brendt and Kamilov, Ulugbek S.},
  year = {2019},
  journal = TCI
}

@article{sunCoILCoordinateBasedInternal2021,
  title = {{{CoIL}}: {{Coordinate-Based Internal Learning}} for {{Tomographic Imaging}}},
  author = {Sun, Yu and Liu, Jiaming and Xie, Mingyang and Wohlberg, Brendt and Kamilov, Ulugbek S.},
  year = {2021},
  journal = TCI,
}

@inproceedings{tancikFourierFeaturesLet2020,
  title = {Fourier {{Features Let Networks Learn High Frequency Functions}} in {{Low Dimensional Domains}}},
  booktitle = NIPS,
  author = {Tancik, Matthew and Srinivasan, Pratul and Mildenhall, Ben and Fridovich-Keil, Sara and Raghavan, Nithin and Singhal, Utkarsh and Ramamoorthi, Ravi and Barron, Jonathan and Ng, Ren},
  year = {2020}
}

@unpublished{tangAcceleratingDeepUnrolling2022,
  title = {Accelerating {{Deep Unrolling Networks}} via {{Dimensionality Reduction}}},
  author = {Tang, Junqi and Mukherjee, Subhadip and Sch{\"o}nlieb, Carola-Bibiane},
  year = {2022},
  journal = {arXiv preprint arXiv:2208.14784},
}

@article{vanaarleFastFlexibleXray2016,
  title = {Fast and Flexible {{X-ray}} Tomography Using the {{ASTRA}} Toolbox},
  author = {Van Aarle, Wim and Palenstijn, Willem Jan and Cant, Jeroen and Janssens, Eline and Bleichrodt, Folkert and Dabravolski, Andrei and De Beenhouwer, Jan and Joost Batenburg, K. and Sijbers, Jan},
  year = {2016},
  journal = {Optics Express}
}

@article{vanaarleASTRAToolboxPlatform2015,
  title = {The {{ASTRA Toolbox}}: {{A}} Platform for Advanced Algorithm Development in Electron Tomography},
  author = {Van Aarle, Wim and Palenstijn, Willem Jan and De Beenhouwer, Jan and Altantzis, Thomas and Bals, Sara and Batenburg, K. Joost and Sijbers, Jan},
  year = {2015},
  journal = {Ultramicroscopy},
}

@inproceedings{venkatakrishnanPlugPlayPriorsModel2013,
  title = {Plug-and-{{Play}} Priors for Model Based Reconstruction},
  booktitle = {IEEE Global Conference on Signal and Information Processing},
  author = {Venkatakrishnan, Singanallur V. and Bouman, Charles A. and Wohlberg, Brendt},
  year = {2013}
}

@inproceedings{voPlugPlayLearnedProximal2025,
  title = {Plug-and-{{Play Learned Proximal Trajectory}} for {{3D Sparse-View X-Ray Computed Tomography}}},
  booktitle = ECCV,
  author = {Vo, Romain and Escoda, Julie and Vienne, Caroline and Decencière, Étienne},
  year = {2024}
}

@article{xieNeuralFieldsVisual2022,
  title = {Neural {{Fields}} in {{Visual Computing}} and {{Beyond}}},
  author = {Xie, Yiheng and Takikawa, Towaki and Saito, Shunsuke and Litany, Or and Yan, Shiqin and Khan, Numair and Tombari, Federico and Tompkin, James and {sitzmann}, Vincent and Sridhar, Srinath},
  year = {2022},
  journal = {Computer Graphics Forum},
}

@inproceedings{zhaNAFNeuralAttenuation2022,
  title = {{{NAF}}: {{Neural Attenuation Fields}} for~{{Sparse-View CBCT Reconstruction}}},
  booktitle = MICCAI,
  author = {Zha, Ruyi and Zhang, Yanhao and Li, Hongdong},
  year = {2022}
}

@inproceedings{zhangLearningDeepCNN2017,
  author={Zhang, Kai and Zuo, Wangmeng and Gu, Shuhang and Zhang, Lei},
  booktitle=CVPR, 
  title={Learning Deep CNN Denoiser Prior for Image Restoration}, 
  year={2017}
}

@article{zhangPlugPlayImageRestoration2021,
  author={Zhang, Kai and Li, Yawei and Zuo, Wangmeng and Zhang, Lei and Van Gool, Luc and Timofte, Radu},
  journal=PAMI, 
  title={Plug-and-Play Image Restoration With Deep Denoiser Prior}, 
  year={2022}
}

@inproceedings{zhuDenoisingDiffusionModels2023,
  title = {Denoising {{Diffusion Models}} for {{Plug-and-Play Image Restoration}}},
  booktitle = CVPR,
  author = {Zhu, Yuanzhi and Zhang, Kai and Liang, Jingyun and Cao, Jiezhang and Wen, Bihan and Timofte, Radu and Van Gool, Luc},
  year = {2023}
}

@article{hauptmannModelBasedLearningAccelerated2020,
  author={Hauptmann, Andreas and Lucka, Felix and Betcke, Marta and Huynh, Nam and Adler, Jonas and Cox, Ben and Beard, Paul and Ourselin, Sebastien and Arridge, Simon},
  journal = TMI, 
  title={Model-Based Learning for Accelerated, Limited-View 3-D Photoacoustic Tomography}, 
  year={2018},
}

@article{hauptmann_convergent_2024,
	title = {Convergent {Regularization} in {Inverse} {Problems} and {Linear} {Plug}-and-{Play} {Denoisers}},
	journal = {Foundations of Computational Mathematics},
	author = {Hauptmann, Andreas and Mukherjee, Subhadip and Schönlieb, Carola-Bibiane and Sherry, Ferdia},
	year = {2024},
}

@inproceedings{sriram_end--end_2020,
author="Sriram, Anuroop
and Zbontar, Jure
and Murrell, Tullie
and Defazio, Aaron
and Zitnick, C. Lawrence
and Yakubova, Nafissa
and Knoll, Florian
and Johnson, Patricia",
title="End-to-End Variational Networks for Accelerated MRI Reconstruction",
booktitle=MICCAI,
year={2020},
}

@article{pezzotti_adaptive_2020,
  author={Pezzotti, Nicola and Yousefi, Sahar and Elmahdy, Mohamed S. and Van Gemert, Jeroen Hendrikus Fransiscus and Schuelke, Christophe and Doneva, Mariya and Nielsen, Tim and Kastryulin, Sergey and Lelieveldt, Boudewijn P. F. and Van Osch, Matthias J. P. and De Weerdt, Elwin and Staring, Marius},
  journal={IEEE Access}, 
  title={An Adaptive Intelligence Algorithm for Undersampled Knee MRI Reconstruction}, 
  year={2020}
}

@article{schmid_decomposing_2000,
	title = {Decomposing a matrix into circulant and diagonal factors},
	journal = {Linear Algebra and its Applications},
	author = {Schmid, Michael and Steinwandt, Rainer and Müller-Quade, Jörn and Rötteler, Martin and Beth, Thomas},
	year = {2000},
}

@book{Vetterli_Kovačević_Goyal_2014, 
    place={Cambridge},
    title={Foundations of Signal Processing}, 
    publisher={Cambridge University Press}, 
    author={Vetterli, Martin and Kovačević, Jelena and Goyal, Vivek K}, 
    year={2014}
}

@article{souza_open_2018,
	title = {An open, multi-vendor, multi-field-strength brain {MR} dataset and analysis of publicly available skull stripping methods agreement},
	journal = {NeuroImage},
	author = {Souza, Roberto and Lucena, Oeslle and Garrafa, Julia and Gobbi, David and Saluzzi, Marina and Appenzeller, Simone and Rittner, Letícia and Frayne, Richard and Lotufo, Roberto},
	year = {2018},
}

@article{lee_deep_2018,
	title = {Deep {Residual} {Learning} for {Accelerated} {MRI} {Using} {Magnitude} and {Phase} {Networks}},
	volume = {65},
	journal = {IEEE Transactions on Biomedical Engineering},
	author = {Lee, Dongwook and Yoo, Jaejun and Tak, Sungho and Ye, Jong Chul},
	year = {2018},
}

@inproceedings{Zamir_2022_CVPR,
    author    = {Zamir, Syed Waqas and Arora, Aditya and Khan, Salman and Hayat, Munawar and Khan, Fahad Shahbaz and Yang, Ming-Hsuan},
    title     = {Restormer: Efficient Transformer for High-Resolution Image Restoration},
    booktitle = CVPR,
    year      = {2022},
}

@article{WANG2021102579,
title = {Deep learning for fast MR imaging: A review for learning reconstruction from incomplete k-space data},
journal = {Biomedical Signal Processing and Control},
year = {2021},
author = {Shanshan Wang and Taohui Xiao and Qiegen Liu and Hairong Zheng},
}

@unpublised{zbontar_fastmri_2019,
	title = {{fastMRI}: {An} {Open} {Dataset} and {Benchmarks} for {Accelerated} {MRI}},
	author = {Zbontar, Jure and Knoll, Florian and Sriram, Anuroop and Murrell, Tullie and Huang, Zhengnan and Muckley, Matthew J. and Defazio, Aaron and Stern, Ruben and Johnson, Patricia and Bruno, Mary and Parente, Marc and Geras, Krzysztof J. and Katsnelson, Joe and Chandarana, Hersh and Zhang, Zizhao and Drozdzal, Michal and Romero, Adriana and Rabbat, Michael and Vincent, Pascal and Yakubova, Nafissa and Pinkerton, James and Wang, Duo and Owens, Erich and Zitnick, C. Lawrence and Recht, Michael P. and Sodickson, Daniel K. and Lui, Yvonne W.},
	year = {2019},
    journal = {arXiv preprint arXiv:1811.08839},
}

@article{kiss_benchmarking_2025,
	title = {Benchmarking learned algorithms for computed tomography image reconstruction tasks},
	journal = {Applied Mathematics for Modern Challenges},
	author = {Kiss, Maximilian B. and Biguri, Ander and Shumaylov, Zakhar and Sherry, Ferdia and Batenburg, K. Joost and Schönlieb, Carola-Bibiane and Lucka, Felix},
	year = {2025},
}

@article{kellman_memory-efficient_2020,
	title = {Memory-{Efficient} {Learning} for {Large}-{Scale} {Computational} {Imaging}},
	journal = {IEEE Transactions on Computational Imaging},
	author = {Kellman, Michael and Zhang, Kevin and Markley, Eric and Tamir, Jon and Bostan, Emrah and Lustig, Michael and Waller, Laura},
	year = {2020},
}

@article{aggarwal_modl_2019,
	title = {{MoDL}: {Model} {Based} {Deep} {Learning} {Architecture} for {Inverse} {Problems}},
	journal = {IEEE Transactions on Medical Imaging},
	author = {Aggarwal, Hemant Kumar and Mani, Merry P. and Jacob, Mathews},
	year = {2019},
}

@unpublished{terris_reconstruct_2025,
	title = {Reconstruct {Anything} {Model}: a lightweight foundation model for computational imaging},
	author = {Terris, Matthieu and Hurault, Samuel and Song, Maxime and Tachella, Julian},
	year = {2025},
    journal = {arXiv preprint arXiv:2503.08915},
}

@article{zheng_efficient_2023,
	title = {Efficient {Identification} of {Butterfly} {Sparse} {Matrix} {Factorizations}},
	journal = {SIAM Journal on Mathematics of Data Science},
	author = {Zheng, Léon and Riccietti, Elisa and Gribonval, Rémi},
	year = {2023},
}

@book{elad_sparse_2010,
	title = {Sparse and {Redundant} {Representations}: {From} {Theory} to {Applications} in {Signal} and {Image} {Processing}},
	publisher = {Springer New York},
	author = {Elad, Michael},
	year = {2010},
}

@article{bolte_proximal_2014,
	title = {Proximal alternating linearized minimization for nonconvex and nonsmooth problems},
	journal = {Mathematical Programming},
	author = {Bolte, Jérôme and Sabach, Shoham and Teboulle, Marc},
	year = {2014},
}

@article{le_spurious_2023,
	title = {Spurious {Valleys}, {NP}-{Hardness}, and {Tractability} of {Sparse} {Matrix} {Factorization} with {Fixed} {Support}},
	journal = {SIAM Journal on Matrix Analysis and Applications},
	author = {Le, Quoc-Tung and Riccietti, Elisa and Gribonval, Remi},
	year = {2023},
}

@article{gnanasambandam_secrets_2024,
  author={Gnanasambandam, Abhiram and Sanghvi, Yash and Chan, Stanley H.},
  journal={IEEE Transactions on Computational Imaging}, 
  title={The Secrets of Non-Blind Poisson Deconvolution}, 
  year={2024},
}

@unpublished{tachella_deepinverse_2025,
	title = {{DeepInverse}: {A} {Python} package for solving imaging inverse problems with deep learning},
	author = {Tachella, Julián and Terris, Matthieu and Hurault, Samuel and Wang, Andrew and Chen, Dongdong and Nguyen, Minh-Hai and Song, Maxime and Davies, Thomas and Davy, Leo and Dong, Jonathan and Escande, Paul and Hertrich, Johannes and Hu, Zhiyuan and Liaudat, Tobías I. and Laurent, Nils and Levac, Brett and Massias, Mathurin and Moreau, Thomas and Modrzyk, Thibaut and Monroy, Brayan and Neumayer, Sebastian and Scanvic, Jérémy and Sarron, Florian and Sechaud, Victor and Schramm, Georg and Vo, Romain and Weiss, Pierre},
	year = {2025},
    journal = {arXiv preprint arXiv:2505.20160},
}

@inproceedings{hurault_relaxed_2023,
author = {Hurault, Samuel and Chambolle, Antonin and Leclaire, Arthur and Papadakis, Nicolas},
title = {A Relaxed Proximal Gradient Descent Algorithm for Convergent Plug-and-Play with Proximal Denoiser},
year = {2023},
booktitle = {International Conference on Scale Space and Variational Methods in Computer Vision},
}

@article{ramzi_ncpdnet_2022,
  author={Ramzi, Zaccharie and G R, Chaithya and Starck, Jean-Luc and Ciuciu, Philippe},
  journal={IEEE Transactions on Medical Imaging}, 
  title={NC-PDNet: A Density-Compensated Unrolled Network for 2D and 3D Non-Cartesian MRI Reconstruction}, 
  year={2022},
}

@software{tiny-cuda-nn,
	author = {M\"uller, Thomas},
	license = {BSD-3-Clause},
	month = {4},
	title = {{tiny-cuda-nn}},
	url = {https://github.com/NVlabs/tiny-cuda-nn},
	version = {2.0},
	year = {2021}
}
}

\clearpage
\setcounter{page}{1}
\maketitlesupplementary
\begin{appendices} \label{appendix}
\crefalias{section}{appendix}

\section{Details about training configurations} \label{appendix:training:details}

Every learned model in \cref{tab:walnut:quantitative,tab:calgary:quantitative} is trained following the same global configuration: we use Adam optimizer \cite{kingmaAdamMethodStochastic2017} with an initial learning rate of $10^{-4}$. We train for $10^5$ steps with a batch size of $4$ (artificially increased via gradient accumulation if necessary). We use cosine annealing which decays the learning rate to $10^{-8}$ at the end of training. 

\paragraph{Fitting the normal operator approximation} We fit the normal operator approximation $\mH(\vm, \bm{\lambda})$ on Gaussian random vectors. For each operator we train for $3000$ steps with a batch size of $4$ using Adam optimizer with a fixed learning rate of $2.10^{-2}$ for CBCT and $2.5 {\times} 10^{-2}$ for MC-MRI. 

\subsection{Walnut-CBCT}

The dataset \cite{dersarkissianConebeamXrayComputed2019} contains 42 CBCT volumes of different walnuts. Each acquisition contains $1200$ radiographies of size $972 \times 768$ pixels, acquired over a full $360^\circ$ circular trajectory. The reconstruction volumes have a size of $501^3$ voxels. The ground-truth volumes are obtained by running an accelerated gradient descent scheme using the full-view data \cite{dersarkissianConebeamXrayComputed2019}. The train/val split contains 34 volumes, while the test split contains 8 volumes.

\begin{figure}[h!] \centering %\setkeys{Gin}{width=0.30\columnwidth}

    \includegraphics[height=0.5\linewidth]{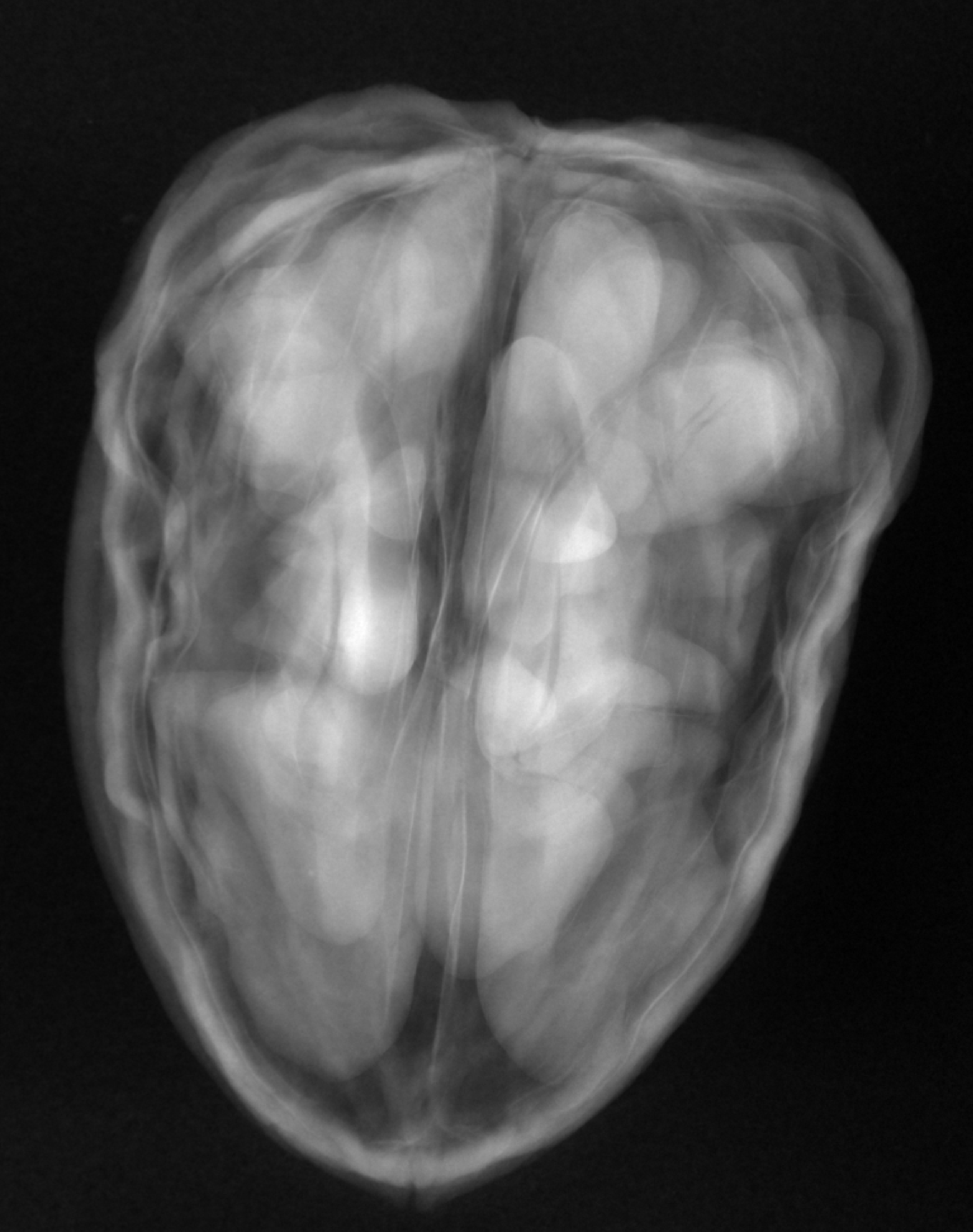}
    \includegraphics[height=0.5\linewidth]{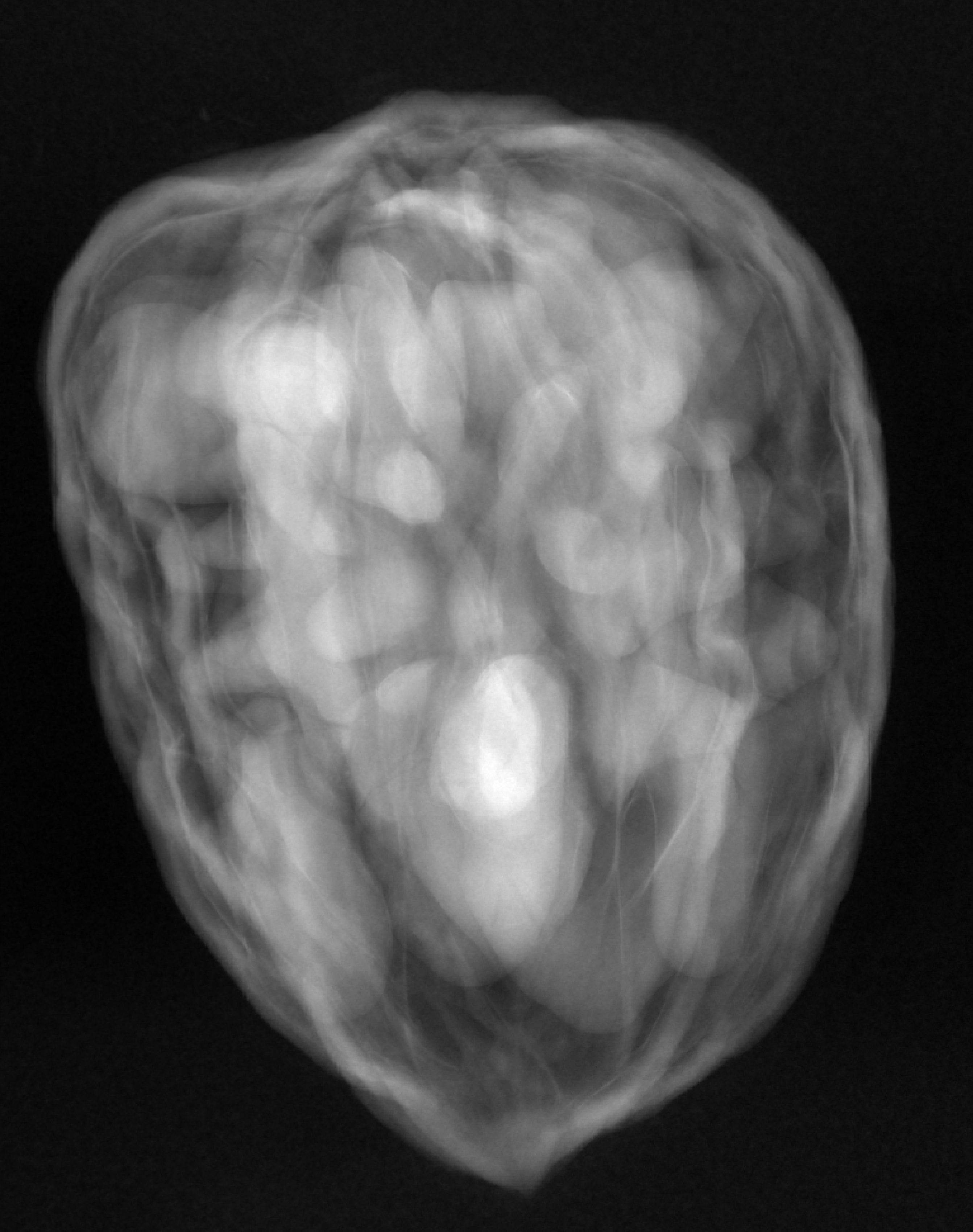}
    \caption{Illustrations of the \textbf{Walnut-CBCT} dataset. Examples of two radiographies.} \label{supp:fig:walnut:dataset}
 \end{figure}

\paragraph{DRUNet initialization} The 2D DRUNet \cite{zhangPlugPlayImageRestoration2021} are initialized with the \texttt{deepinverse} weights \cite{tachella_deepinverse_2025}. During subsequent training, \ie unrolling or post-processing, we fix the required noise value to $\sigma=0.03$. The 3D DRUNet is initialized using the 2D DRUNet weights by inflating the 2D kernels to 3D. All kernels are of shape $3\times3$, for an equivalent $3\times3\times3$ kernel, we copy the 2D weights to the central slice along the depth dimension and initialize the other slices with zeros. For upsampling and downsampling layers of shape $2\times2$, we duplicate the 2D weights along the depth dimension to form $2\times2\times2$ kernels, and normalize them to preserve the overall kernel norm.

\paragraph{Ours} Training \textbf{our} method, \ie 3D DRUNet unrolled for $K=3$ iterations with \textbf{domain partitioning} (cuboid patch size of $8\times384^2$) takes approximately 185 hours on a single NVIDIA H100 GPU with 80GB of video memory. Combining \textbf{domain partitioning} with \textbf{normal operator approximation} further reduces the training time to approximately 135 hours on the same hardware.

\paragraph{TV} We solve (\ref{eq:variational:problem}) with $g(x) = \| \nabla x \|_1$ using FISTA \cite{beckFastIterativeShrinkageThresholding2009} for $1000$ iterations with $\lambda = 0.2$, and step size $\eta = 1 / L$, where $L$ is the spectral norm of $\mA^\top\mA$ estimated via power iteration.

\paragraph{PnP-$\alpha$PGD} For 2D and 3D, we respectively use the post-processing DRUNet trained on 2D slices and 3D patches as described above as the prior. More precisely, we run an accelerated PGD scheme \cite{beckFastIterativeShrinkageThresholding2009} for $K=40$ iterations with step size $\eta = 1.0 / L$. For stability reasons, we relax the network evaluation with $\operatorname{D}_\alpha = (1-\alpha) \mathrm{Id} + \alpha \operatorname{D}_\phi$, where $\operatorname{D}_\phi$ is the corresponding DRUNet. Similar to \cite{voPlugPlayLearnedProximal2025}, we define $\alpha = \frac{\eta \lambda}{1 + \eta \lambda}$, where $\lambda$ is the weight of the implicit prior defined by the artifact removal network. We set $\lambda=10.$ for both 2D and 3D.

\paragraph{INR} Training an \texttt{instant-ngp} requires specific library dependencies. We use the specific \texttt{CUDA} implementation from \texttt{tiny-cuda-nn} \cite{tiny-cuda-nn} and the ray-casting code from \texttt{nerfacc} \cite{liNerfAccGeneralNeRF2022}. The INR uses a hash grid encoding \cite{mullerInstantNeuralGraphics2022} with 16 levels and a capacity of $T=2^{21}$, a feature dimension of 2, a base resolution of 16, and a finest resolution of 256. The MLP has 2 hidden layers with 64 hidden units each. We train the INR by stochastic coordinate descent with Adam optimizer. The learning rate is initialized to $10^{-4}$ and decayed to $10^{-8}$ following a cosine scheduling for $25$k steps. We cast $2048$ rays per step, compute the associated data-fidelity for these rays and backpropagate the gradients to update the MLP and hash grid parameters. 

\paragraph{DPIR[RAM]} We use the PnP Half-Quadratic Splitting (HQS) method from \cite{zhangPlugPlayImageRestoration2021} with a Reconstruct Anything Model (RAM) prior \cite{terris_reconstruct_2025}. We run it for $K=20$ iterations, with $\lambda=1/0.23$ and a decreasing noise schedule $\sigma_k$ starting from $\sigma_1=49/255.$ to $\sigma_K=2.10^{-3}$.

\subsection{Calgary-Campinas MC-MRI}

\begin{figure}[h!] \centering 
    \includegraphics[height=0.30\columnwidth]{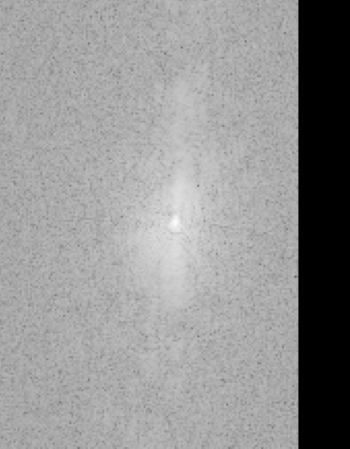}%
    \includegraphics[height=0.30\columnwidth]{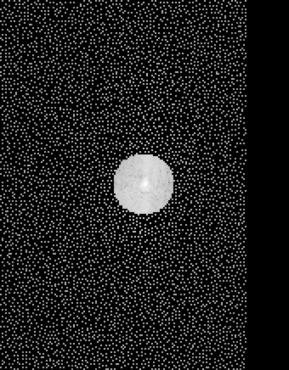}
    \vspace*{0.5em}
    \includegraphics[height=0.30\columnwidth]{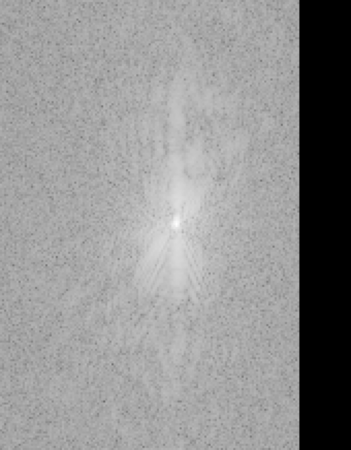}%
    \includegraphics[height=0.30\columnwidth]{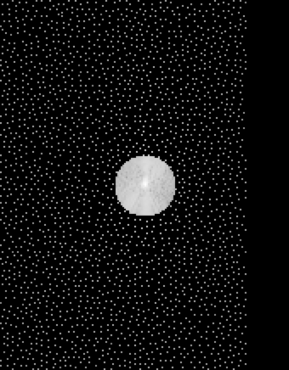}
    \caption{Illustrations of the \textbf{Calgary-Campinas} dataset. Examples of k-space measurements. (\textit{left}) Slice of fully-sampled k-space (1st coil), and subsequently undersampled k-space with acceleration rate of $5$. (\textit{right}) Slice of fully-sampled k-space (1st coil) and, corresponding undersampled k-space with acceleration rate of $10$.} \label{supp:fig:calgary:dataset}
 \end{figure}

\noindent The dataset \cite{souza_open_2018} contains 59 fully-sampled Cartesian Multi-Coil MRI acquisitions of different brains. Each acquisition contains $12$ coils with a k-space of size $256 \times 218 \times 170$. The k-space data is provided in a hybrid format, \ie x-ky-kz, where a 1D FFT has already been applied along the frequency-encoded direction. The acceleration is performed along the phase encoding and slice encoding directions (ky and kz). This reduces the 3D problem to a set of independent 2D problems along the frequency-encoded direction (x), which allows us to train standard unrolled network without \textbf{domain partitioning}. We use a set of 48 samples for the train/val split and 21 samples for the test split. The ground-truth images are obtained by applying \textit{root-sum-of-squares} (RSS) reconstruction on the fully-sampled k-space data. The sensitivity maps are estimated as in \cite{pezzotti_adaptive_2020} by applying inverse FFT on the fully-sampled centered region of k-space data.

\paragraph{DRUNet initialization} The \texttt{deepinverse} library does not provide pretrained weights for complex data so both the 2D DRUNet and 3D DRUNet are trained from scratch. 

\paragraph{Unrolled} As opposed to Walnut-CBCT, we can compare our method to a standard unrolled network without \textbf{domain partitioning} since the problem reduces to a set of independent 2D problems along the frequency-encoded direction (x). We use a sub-problem size of $8 \times 218 \times 170$ which maximizes VRAM usage during training. Training the 3D unrolled network for $K=5$ iterations takes approximately 60 hours on a single NVIDIA H100 GPU with 80GB of video memory.

\paragraph{Ours} Training \textbf{our} method, \ie 3D DRUNet unrolled for $K=5$ iterations with \textbf{domain partitioning} (cuboid patch size of $8\times 128^2$) takes approximately 17 hours and twice less memory that standard unrolling on a single NVIDIA H100 GPU with 80GB of video memory. Combining \textbf{domain partitioning} with \textbf{normal operator approximation} does not reduce training time as the original MC-MRI forward operator is already efficient to compute.

\paragraph{TV} We solve (\ref{eq:variational:problem}) with $g(x) = \| \nabla x \|_1$ using FISTA \cite{beckFastIterativeShrinkageThresholding2009} for $1000$ iterations with $\lambda = 7.4 \times 10^{-4}$, and step size $\eta = 1.$, as the MC-MRI forward already has a spectral norm of 1.

\paragraph{PnP-$\alpha$PGD} For 2D and 3D, we respectively use the post-processing DRUNet trained on 2D slices and 3D patches as described above as the prior. More precisely, we run an accelerated PGD scheme \cite{beckFastIterativeShrinkageThresholding2009} for $K=12$ iterations with step size $\eta = 1.0$. For stability reasons, we relax the network evaluation with $\operatorname{D}_\alpha = (1-\alpha) \mathrm{Id} + \alpha \operatorname{D}_\phi$, where $\operatorname{D}_\phi$ is the corresponding DRUNet. Similar to \cite{voPlugPlayLearnedProximal2025}, we define $\alpha = \frac{\eta \lambda}{1 + \eta \lambda}$, where $\lambda$ is the weight of the implicit prior defined by the artifact removal network. We set $\lambda=2.10^{-6}$ for both 2D and 3D.

\paragraph{INR} As \texttt{tiny-cuda-nn} \cite{tiny-cuda-nn} does not natively support complex computation, required for FFT computations, we implement a similar grid-based INR architecture using PyTorch. The encoding is a latent grid of size $80^3$ with $8$ features on each vertex. The MLP has 2 hidden layers with 64 hidden units each. We train the INR by gradient descent with Adam optimizer for $1000$ steps. The learning rate is fixed and set to $1e-2$. At each step we compute the data-fidelity on the full volume and backpropagate the gradients to update the MLP and grid parameters.

\paragraph{DPIR[RAM]} We use the PnP Half-Quadratic Splitting (HQS) method from \cite{zhangPlugPlayImageRestoration2021} with a Reconstruct Anything Model (RAM) prior which natively provides the possibility to process complex data \cite{terris_reconstruct_2025}. We run it for $K=20$ iterations, with $\lambda=1/0.23$ and a decreasing noise schedule $\sigma_k$ starting from $\sigma_1=5.10^{-2}$ to $\sigma_K=2.10^{-3}$.

\section{Details about test-time configurations} \label{appendix:test:details}

In this section we detail the sub-procedure of \cref{alg:test:patch:unrolling}, \ie the procedure to get $\tilde{\vx}$, namely we deploy a standard unrolling scheme by evaluating the prior on sequential patches. 

We break each step (\ref{eq:unrolled}) in $\operatorname{R}_\phi(\vy, \mA)$ as follows: \textbf{(i)} a first gradient descent step, computed on the full volume and full problem (\ref{eq:inverse:problem}) $\vx_{k}' = \vx_{k} - \eta \nabla_{\vx_{k}} d(\mA\vx_{k}, \vy)$, \textbf{(ii)} followed by a prior step on sequential patches, $\vx_{k+1, p} = \operatorname{D}_\phi(\mS_p \vx_{k}')$, where $\mS_p$ extracts the $p$-th patch from the full volume. Finally, we aggregate the processed patches to form the full volume $\vx_{k+1}$. We repeat this procedure for $K$ iterations to get $\tilde{\vx} = \vx_{K}$.

As opposed to training, this procedure is possible at test-time since we do not need to store activations for each processed patch, allowing us to process arbitrarily large volumes with limited memory. 

\paragraph{Patch-based strategy} At test-time, we observe that the trained networks are robust to the choice of patch size. For \textbf{MC-MRI}, we only deploy a standard unrolling procedure and use patches of size $8 \times 218 \times 170$ with 3D methods and $218 \times 170$ with 2D methods, \ie full spatial dimensions in H${\times}$W. We either use a stride of size $4$ along the depth dimension in 3D, or evaluate slice-by-slice in 2D. Then we aggregate to build a  prediction of size $256 \times 218 \times 170$. For \textbf{CBCT}, we use two different strategies. When computing the estimation of the ground-truth $\tilde{\vx}$ with the standard unrolling procedure, we use patches of size $8 \times 501^2$ with 3D methods and $501^2$ with 2D methods. We either use a stride of size $4$ along the depth dimension in 3D, or evaluate slice-by-slice in 2D. Then we aggregate to build a prediction of size $501^3$. In the second part, we deploy unrolling with domain partitioning and use patches of size $8 \times 384^2$ or $384^2$ for 3D and 2D methods respectively. 

\paragraph{Evaluation strategy} When evaluating the methods on the Walnut-CBCT dataset, we crop the ground-truth and predicted volumes to the central $300^3$ voxels to avoid boundary artifacts and focus on part of the volumes that contains well-defined material.

\section{Additional results} \label{appendix:additional:results}

\subsection{Influence of the patch size on the performances}

\begin{figure}[h!]
    \centering
    \includegraphics[width=0.9\linewidth]{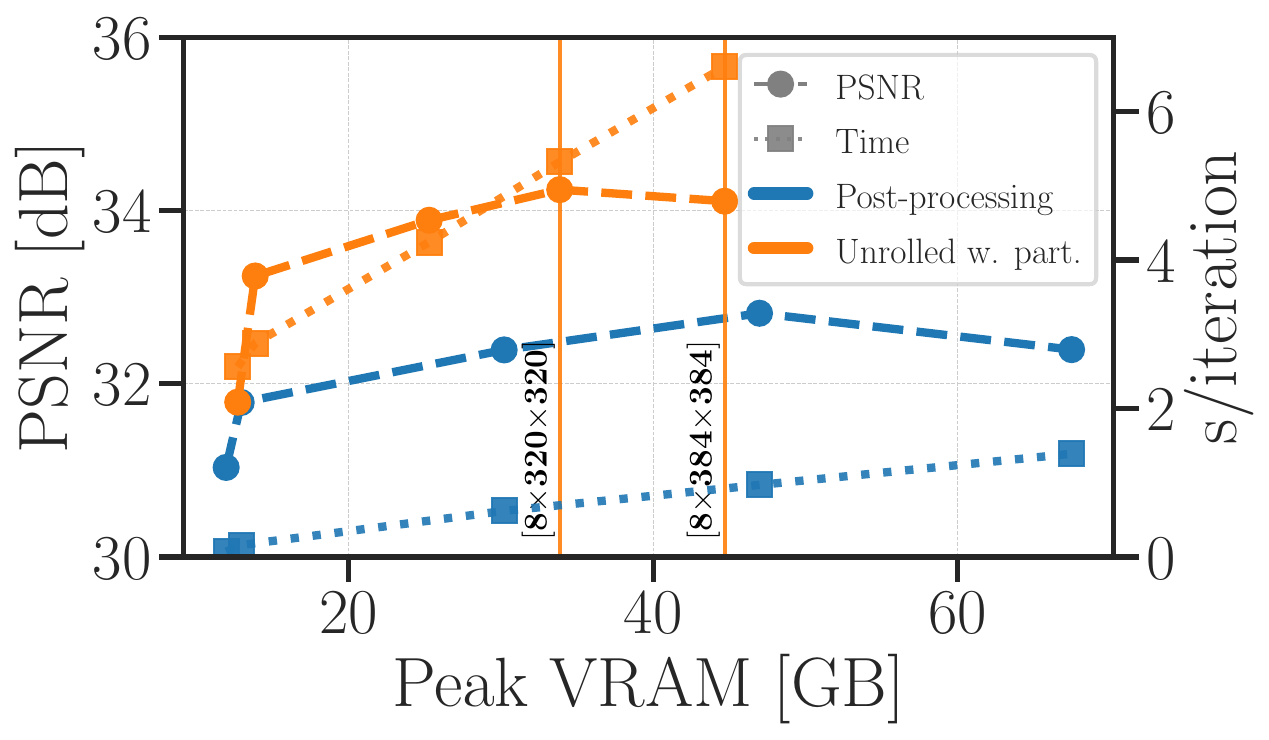}
    \caption{\textbf{Walnut-CBCT} -  Average PSNR and time complexity against peak memory consumption during training. We vary the VRAM budget by changing the patch size used during domain partitioning. Larger patches lead to better performance at the cost of higher memory consumption. We do not show the complexity of standard unrolling (without partitioning) as a single H100 GPU is not sufficient for training it.}
    \label{supp:fig:walnut:patch:size:influence}
\end{figure}

\begin{figure}[h!]
    \centering
    \includegraphics[width=0.9\linewidth]{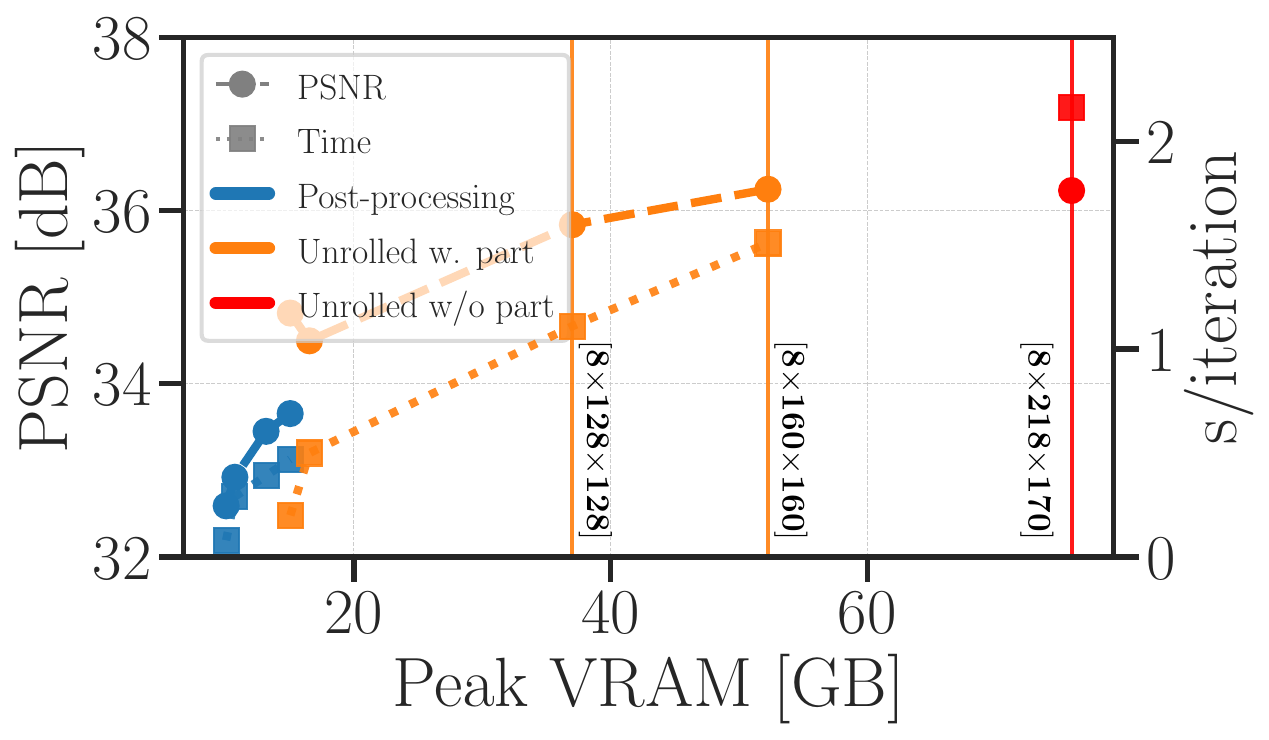}
    \caption{\textbf{Calgary-Campinas MC-MRI} - Average PSNR and time complexity against peak memory consumption during training. We vary the VRAM budget by changing the patch size used during domain partitioning. Larger patches lead to better performance at the cost of higher memory consumption. }
    \label{supp:fig:calgary:patch:size:influence}
\end{figure}

\noindent In \cref{supp:fig:walnut:patch:size:influence,supp:fig:calgary:patch:size:influence} we provide the evolution of average PSNR and time complexity against peak memory consumption during training for different patch sizes on Walnut-CBCT and Calgary-Campinas MC-MRI datasets respectively. 

\vspace{0.5em}

\noindent We observe that larger patches lead to better performance at the cost of higher memory consumption. We note that on both dataset, for the same memory budget, our method with domain partitioning outperforms the standard post-processing. On Walnut-CBCT, we use patch sizes in $[(384 \times 384), (8 \times 128^2), (8 \times 256^2), (8 \times 320^2), (8 \times 384^2)]$, while on Calgary-Campinas MC-MRI, we use patch sizes in $[(218 \times 170), (8 \times 64^2), (8 \times 128^2), (8 \times 160^2), (8 \times 218 \times 170)]$.

\vspace{0.5em}

\noindent Interestingly, we observe that on the MC-MRI experiment, performance starts to plateau for patch sizes larger than $8 \times 128^2$, suggesting that the standard unrolling with no partitioning does not bring significant benefits compared to our method with domain partitioning for this specific problem and network complexity.

\subsection{Normal operator approximations}

\begin{figure}[h!] \centering %\setkeys{Gin}{width=0.30\columnwidth}

    \stackinset{l}{2pt}{t}{2pt}{\large \whitetext{$\vx$}}{%
        \includegraphics[width=0.33\columnwidth]{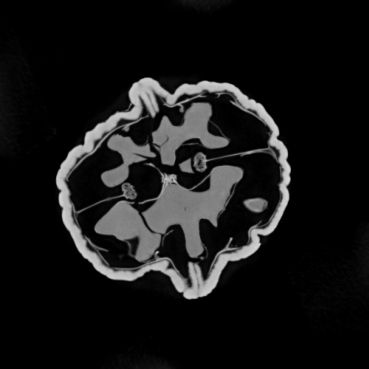}%
    }%
    \stackinset{l}{2pt}{t}{2pt}{\large \whitetext{$\mA^\top \mA \vx$}}{%
        \includegraphics[width=0.33\columnwidth]{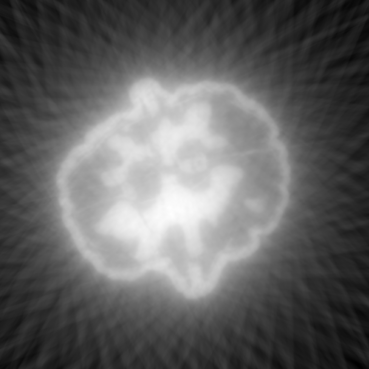}%
    }%
    \stackinset{l}{2pt}{t}{2pt}{\large \whitetext{$\mH(\vm, \bm{\lambda}) \vx$}}{%
        \includegraphics[width=0.33\columnwidth]{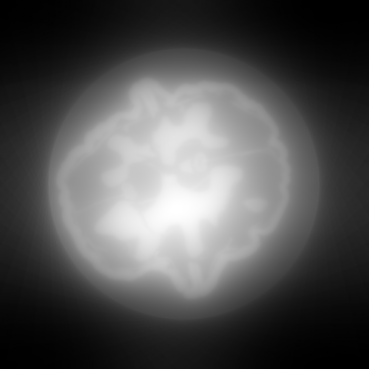}%
    }%

    \hspace*{0.00005em}
    \stackinset{l}{2pt}{t}{2pt}{\large \whitetext{$\bm{\lambda}$}}{%
        \includegraphics[width=0.33\columnwidth]{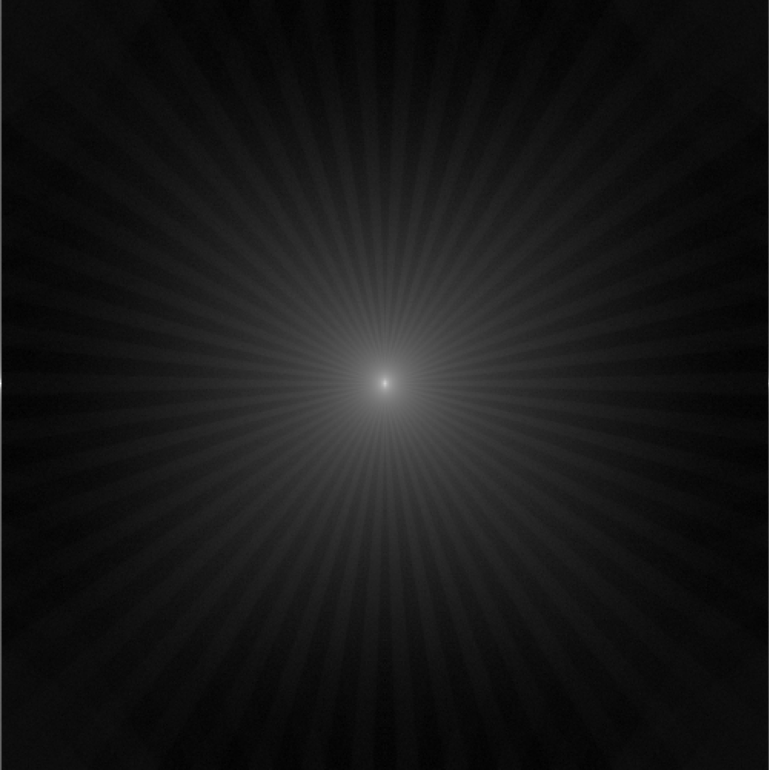}%
    }%
    \stackinset{l}{2pt}{t}{2pt}{\large \whitetext{$\vm$}}{%
        \includegraphics[width=0.33\columnwidth]{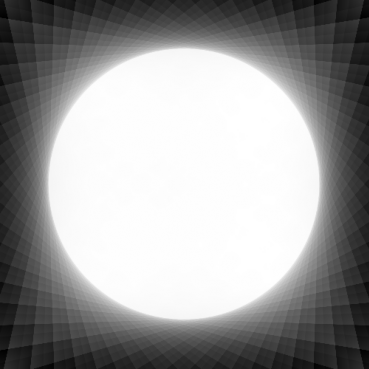}%
    }%
    \stackinset{l}{2pt}{t}{2pt}{\small \whitetext{$(\mA^\top \mA \vx - \mH \vx)^2$}}{%
        \includegraphics[width=0.33\columnwidth]{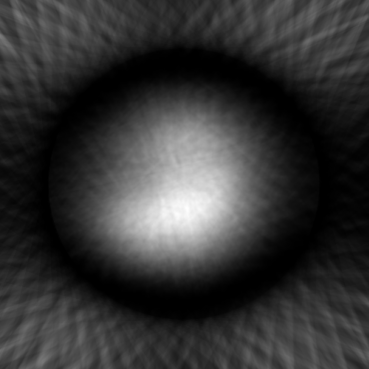}
    }%

    \caption{Illustrations of the normal operator approximation on \textbf{Walnut-CBCT}. (\textit{top row}) Original volume slice $\vx$, exact normal operator evaluation $\mA^\top \mA \vx$, and approximated normal operator $\mH(\vm, \bm{\lambda}) \vx$. (\textit{bottom row}) Learned filter $\bm{\lambda}$, learned mask $\vm$, and squared approximation error $(\mA^\top \mA \vx - \mH \vx)^2$.} \label{supp:fig:cbct:approx}
 \end{figure}

\paragraph{Walnut-CBCT} In \cref{supp:fig:cbct:approx}, we provide additional illustrations of the normal operator approximation on Walnut-CBCT. We show a slice of the original volume $\vx$, the exact normal operator evaluation $\mA^\top \mA \vx$, and the approximated normal operator $\mH(\vm, \bm{\lambda}) \vx$. We see that $\bm{\lambda}$, which corresponds to a filter in the Fourier domain, exhibits patterns predicted by the Fourier slice theorem, \ie it performs sampling along radial directions.

\begin{figure}[ht!] \centering %\setkeys{Gin}{width=0.30\columnwidth}

    \stackinset{l}{2pt}{t}{2pt}{\large \whitetext{$|\vx|$}}{%
        \includegraphics[width=0.33\columnwidth]{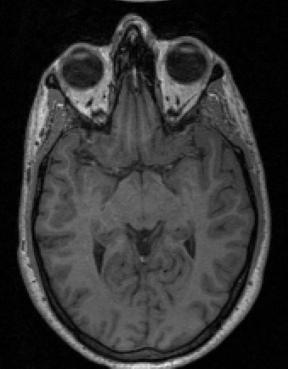}%
    }%
    \stackinset{l}{2pt}{t}{2pt}{\large \whitetext{$|\mA^H \mA \vx|$}}{%
        \includegraphics[width=0.33\columnwidth]{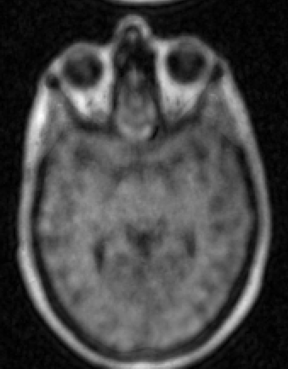}%
    }%
    \stackinset{l}{2pt}{t}{2pt}{\large \whitetext{$|\mH(\vm, \bm{\lambda}) \vx|$}}{%
        \includegraphics[width=0.33\columnwidth]{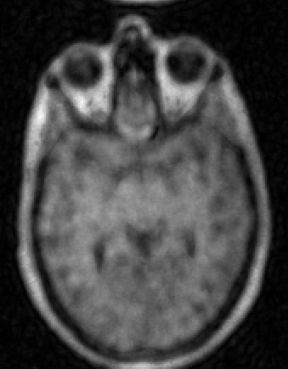}%
    }%

    \hspace*{0.00005em}
    \stackinset{l}{2pt}{t}{2pt}{\large \whitetext{$|\bm{\lambda}|$}}{%
        \includegraphics[width=0.33\columnwidth]{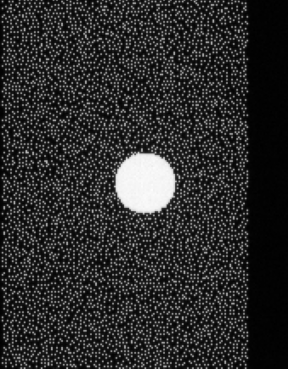}%
    }%
    \stackinset{l}{2pt}{t}{2pt}{\large \whitetext{$|\vm|$}}{%
        \includegraphics[width=0.33\columnwidth]{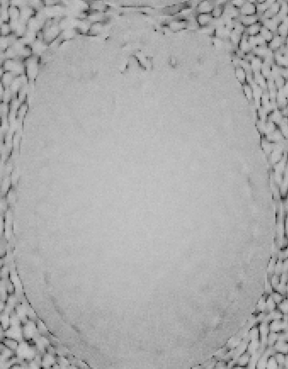}%
    }%
    \stackinset{l}{2pt}{t}{2pt}{\small \whitetext{$|\mA^H \mA \vx - \mH \vx|^2$}}{%
        \includegraphics[width=0.33\columnwidth]{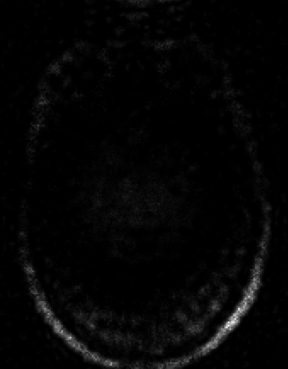}
    }%

    \caption{Illustrations of the normal operator approximation on \textbf{Calgary-Campinas}. (\textit{top row}) Original volume slice $|\vx|$, exact normal operator evaluation $|\mA^H \mA \vx|$, and approximated normal operator $|\mH(\vm, \bm{\lambda}) \vx|$. (\textit{bottom row}) Learned filter $|\bm{\lambda}|$, learned mask $|\vm|$, and squared approximation error $|\mA^H \mA \vx - \mH \vx|^2$.} \label{supp:fig:calgary:approx}
 \end{figure}

\paragraph{Calgary-Campinas MC-MRI} In \cref{supp:fig:calgary:approx}, we provide additional illustrations of the normal operator approximation on Calgary-Campinas MC-MRI. We show a slice of the original volume $|\vx|$, the exact normal operator evaluation $|\mA^H \mA \vx|$, and the approximated normal operator $|\mH(\vm, \bm{\lambda}) \vx|$. We see that $\bm{\lambda}$, which corresponds to a filter in the Fourier domain, performs the same Fourier masking operation as the original acceleration pattern (\cref{supp:fig:calgary:dataset}). The spatial modulation $\vm$ performs an operation similar to coil sensitivity maps.

\end{appendices}

\end{document}